\RequirePackage{fix-cm}
\documentclass[twocolumn]{svjour3}          
\smartqed  
\usepackage{amsmath}
\usepackage{booktabs}
\usepackage{amssymb}
\usepackage{mathtools}
\usepackage{xspace}
\usepackage{hyperref}
\usepackage{graphicx}
\usepackage{caption}
\usepackage{subcaption}
\usepackage{multirow}
\usepackage{color, colortbl}
\usepackage{fancyvrb}
\usepackage{blindtext}
\usepackage{appendix}
\usepackage{natbib}
\usepackage{bm}
\usepackage{algorithm}
\usepackage{etoolbox}
\usepackage{lipsum}
\usepackage{array}
\newcolumntype{M}{>{$}c<{$}}

\usepackage[noend]{algpseudocode}

\definecolor{Gray}{gray}{0.9}
\definecolor{pink}{rgb}{1.0, 0.13, 0.32}

\journalname{International Journal of Computer Vision}

\hypersetup{pdfstartview={FitH}}
\hypersetup{pdfborder={0 0 0}}
\hypersetup{pdfpagemode={UseNone}}
\hypersetup{colorlinks=true}
\hypersetup{citecolor=blue}
\hypersetup{linkcolor=blue}

\begin{document}
\begin{sloppypar}

\title{Assessing the Robustness of Visual Question Answering Models}


\author{Jia-Hong Huang, Modar Alfadly, Bernard Ghanem, Marcel Worring 
}


\institute{Jia-Hong Huang \at
              Universiteit van Amsterdam, Amsterdam, the Netherlands\\
              King Abdullah University of Science and Technology, Makkah, KSA\\
              \email{j.huang@uva.nl}           
           \and
           Modar Alfadly \at
              King Abdullah University of Science and Technology, Makkah, KSA\\
              \email{modar.alfadly@kaust.edu.sa} 
           \and
           Bernard Ghanem \at
              King Abdullah University of Science and Technology, Makkah, KSA\\
              \email{bernard.ghanem@kaust.edu.sa} 
           \and
           Marcel Worring \at
              Universiteit van Amsterdam, Amsterdam, the Netherlands\\
              \email{m.worring@uva.nl}
}

\date{Received: date / Accepted: date}

\maketitle

\begin{abstract}
Deep neural networks have been playing an essential role in the task of Visual Question Answering (VQA). Until recently, their accuracy has been the main focus of research. Now there is a trend toward assessing the robustness of these models against adversarial attacks by evaluating the accuracy of these models under increasing levels of noisiness in the inputs of VQA models. In VQA, the attack can target the image and/or the proposed query question, dubbed main question, and yet there is a lack of proper analysis of this aspect of VQA. In this work, we propose a new method that uses semantically related questions, dubbed basic questions, acting as noise to evaluate the robustness of VQA models. We hypothesize that as the similarity of a basic question to the main question decreases, the level of noise increases. To generate a reasonable noise level for a given main question, we rank a pool of basic questions based on their similarity with this main question. We cast this ranking problem as a $LASSO$ optimization problem. We also propose a novel robustness measure $R_{score}$ and two large-scale basic question datasets in order to standardize robustness analysis of VQA models. The experimental results demonstrate that the proposed evaluation method is able to effectively analyze the robustness of VQA models. To foster the VQA research, we will publish our proposed datasets.

\end{abstract}

\section{Introduction}
Visual Question Answering (VQA) is one of the most challenging computer vision tasks in which an algorithm is given a natural language question about an image and is tasked with producing a natural language answer for that question-image pair. Recently, various VQA models \cite{4,5,9,31,37,41,57,58,59,77,82,agrawal2018don,vedantam2019probabilistic,chen2020counterfactual,sheng2021human,kolling2022efficient} have been proposed to tackle this problem, and their main performance measure is accuracy. 

The community has started to realize that accuracy is not the only metric to evaluate model performance \cite{kafle2017visual,kafle2017analysis}. More specifically, these models should also be robust, {\em i.e.}, their output should not be affected much by some small \emph{noise} or \emph{perturbation} added to the input, such as replacing words by similar words, phrases, and sentences for input questions, or slightly modified pixel values. The idea of analyzing model robustness as well as training robust models is already a rapidly growing research topic for deep learning models applied to images only \cite{61,62,63}. To the best of our knowledge, an acceptable and standardized method to measure robustness in VQA models does not exist. 

\begin{figure}[t]
\begin{center}
   \includegraphics[width=0.98\linewidth]{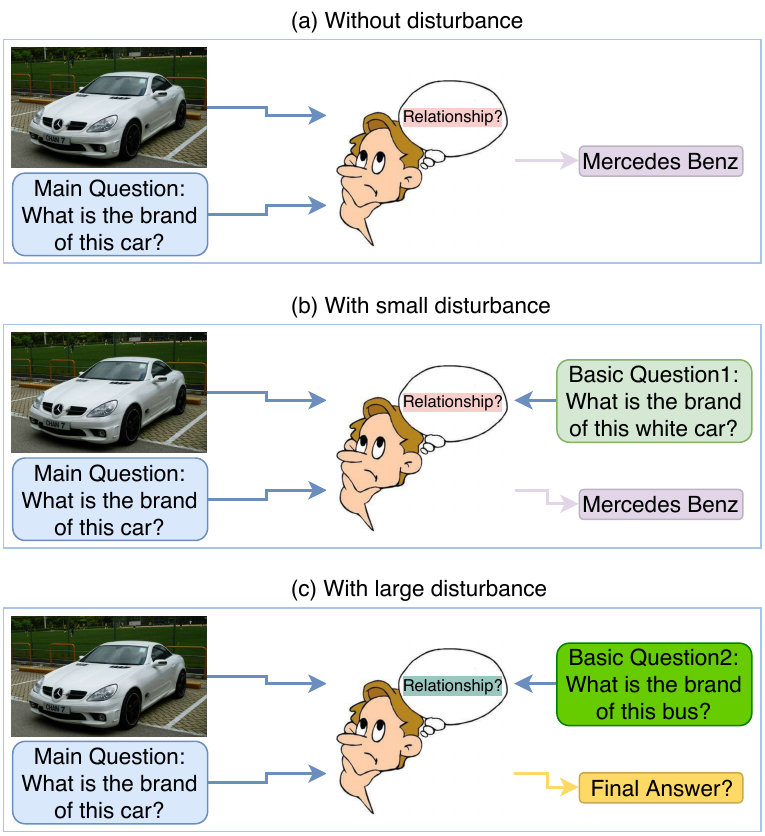}
\end{center}
\vspace{-12pt}
   \caption{Inspired by Deductive Reasoning in Human Thinking \cite{74}, this figure showcases the behavior of humans when subjected to multiple questions about a certain subject. In case (a) and (b), the person may have the same answer ``Mercedes Benz'' in mind. However, in case (c), s/he would start to think more about the relations among the given questions and candidate answers to form the final answer which may be different from the final answer in case (a) and (b). If the person is given more basic questions, s/he would start to think about all the possible relations of all the provided questions and possible answer candidates. These relationships will clearly be more complicated, especially when the additional basic questions have low similarity scores to the main question. In such cases, they will mislead the person. That is to say, those extra basic questions are large disturbances. Note that the relationships and the final answer in the case (a) and (b) can be the same, but different from the case (c). We use different colors to make the above clearer.}
\label{fig:figure1}
\end{figure}

\begin{figure}[t]
\begin{center}
   \includegraphics[width=0.98\linewidth]{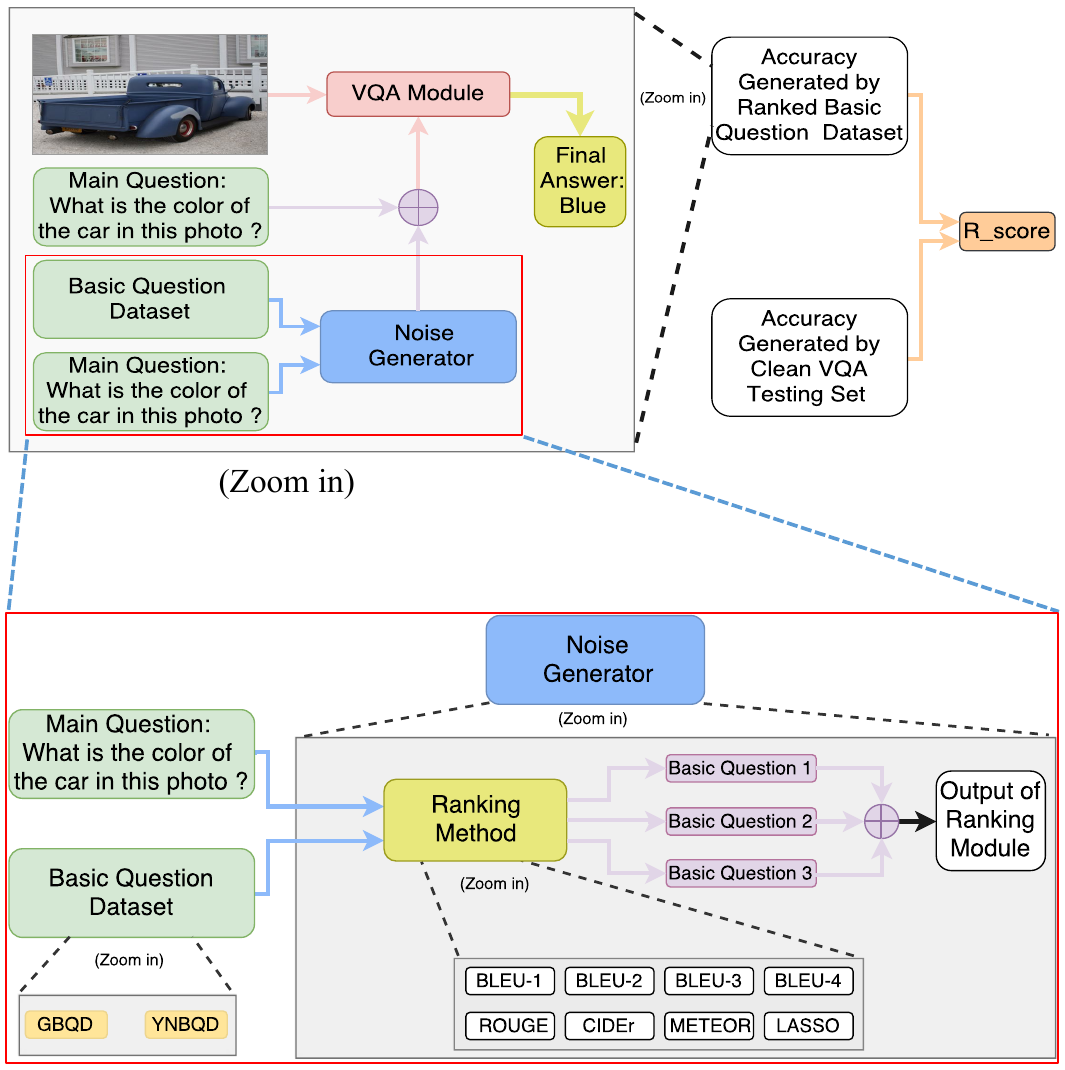}
\end{center}
\vspace{-12pt}
   \caption{Our proposed method for measuring the robustness of VQA models. The $R_{score}$ -- our proposed robustness measure -- is generated based the \textit{``Accuracy Generated by Ranked Basic Question Dataset''} and \textit{``Accuracy Generated by Clean VQA Testing Set''}. In the upper white box of the upper part of the figure, we have two main components, a VQA Module, and a Noise Generator. A detailed view of the Noise Generator is in the lower part of the figure. Then, we have two choices, GBQD and YNBQD, of Basic Question Dataset and eight different choices of question ranking methods. If a new Basic Question Dataset or ranking method is proposed in the future, we can add them into our proposed method. The output of Noise Generator is the concatenation of three ranked basic questions. ``$\oplus$'' denotes the direct concatenation of basic questions.
}
\label{fig:figure100}
\end{figure}

To find a robustness measure, we note the ultimate goal for VQA models is to perform as humans do for the same task. Now, if a human is presented with a question or the same question which is accompanied by some highly similar questions, s/he tends to give the same or a very similar answer in both cases. Evidence of this has been reported in psychology \cite{74}. In our work, we call an input question the main question and define a basic question as a question semantically similar to the given main question. When we add or replace some words or phrases by semantically similar entities to the main question, the VQA model should output the same or a very similar answer. This is illustrated in Figure \ref{fig:figure1}. We can consider these added entities as small perturbations or noise to the input. The model is robust if it produces the same answer. 
Because robustness analysis requires studying the accuracy of VQA models under different noise levels, we need to know how to quantify the level of noise for the given question. We hypothesize that a basic question with larger similarity score to the main question is considered to inject a smaller amount of noise if it is added to the main question and vice versa. Inspired by the above reasoning, we propose a novel method for measuring the robustness of VQA models. Figure \ref{fig:figure100} depicts the structure of our method. It contains two modules, a VQA model and a Noise Generator. The Noise Generator takes a plain text main question (MQ) and a plain text basic question dataset (BQD) as input. It starts by ranking the basic questions in BQD by their similarity to MQ using a text similarity ranking method. 
To measure the robustness of this VQA model, the accuracy with and without the generated noise for different noise levels is compared. We propose a robustness measure $R_{score}$ to quantify performance.

For the question ranking method, given a main question and a basic question, we can have different measures that quantify the similarity of those questions and produce a score. These different similarities lead to a different ranking. Commonly used text similarity metrics, such as BLEU (BiLingual Evaluation Understudy) \cite{49}, are based on computing the overlapping of two texts. However, these metrics cannot capture the semantic meaning of text very well. So, question rankings based on the commonly used text similarity metrics are not accurate.  

To improve the question ranking quality, we propose a new method formulated using $LASSO$ optimization and compare it to other rankings produced by the commonly used textual similarity measures. Then, we do perform this comparison to rank our proposed BQDs, General Basic Question Dataset (GBQD) and Yes/No Basic Question Dataset (YNBQD). Furthermore, we evaluate the robustness of six pretrained state-of-the-art VQA models \cite{4,41,57,59}. Finally, we conduct extensive experiments to compare our proposed $LASSO$ ranking method with the other metrics in BQD ranking. 

Note that since the basic question (BQ) rankings based on those commonly used textual similarity measures are not effective, the noise level is not controllable based on those measures. However, our proposed $LASSO$ basic question ranking method is effective. It is capable of quantifying and controlling the strength of the injected noise level. In this paper, our main contributions are summarized as follows:

\begin{itemize}
    \item[$\bullet$]  We introduce two large-scale basic questions datasets and make available two datasets for VQA robustness evaluation.
    \item[$\bullet$]  We propose a novel method to measure the robustness of VQA models and test it on six different state-of-the-art VQA models.
    \item[$\bullet$]  We propose a new $LASSO$-based text similarity ranking method and show that it outperforms seven popular similarity metrics. 
\end{itemize}

The rest of our paper is organized as follows. In section 2, we review several related works. In section 3, we discuss the details of our proposed method and demonstrate how to use it for measuring the robustness of VQA models. Furthermore, in sections 4 and 5, we present the various analyses on our proposed General Basic Question Dataset (GBQD) and Yes/No Basic Question Dataset (YNBQD) \cite{huang2019novel}. Finally, in section 6, we compare the performance of the state-of-the-art VQA models in terms of robustness and accuracy.

\noindent
\vspace{+0.1cm}\textbf{Relations to our previous work}

This paper is an improved work based on our previous conference paper accepted by the Thirty-Third AAAI Conference on Artificial Intelligence (AAAI-2019) as an oral paper \cite{huang2019novel}. Compared to this work, the improvements are summarized as follows. First, we propose a framework, referring to Figure \ref{fig:figure8}, and a threshold-based criterion, referring to Algorithm \ref{algorithm1}, to exploit BQs to analyze the most robust HieCoAtt VQA model \cite{41}. 
Second, we show that the step of question sentences preprocessing is necessary for our proposed $LASSO$ ranking method, and it guarantees that the proposed method works correctly. Third, 
we conduct an extended experiment on YNBQD. The current paper is a complete restructured and rewritten version.

\section{Related Work}
Over the past few years, VQA has emerged as an interesting and intelligent task which has drawn lots of attention from the research community with a variety of approaches \cite{2,5,6,9,16,26,34,39,41,42,54,64,agrawal2018don,vedantam2019probabilistic,chen2020counterfactual,sheng2021human,kolling2022efficient}. The above works involve different fields including natural language progressing (NLP), computer vision, and machine learning. In the following subsections, we discuss numerous bodies of work including the sentence evaluation metrics, models' accuracy and robustness, and datasets, related to our paper.

\noindent
\vspace{+0.1cm}\textbf{Sentence Evaluation Metrics}

Sentence evaluation metrics have been widely used in several areas such as video/image captioning \cite{yu2016video,huang2021deepopht,huang2022non,huang2021contextualized,huang2021deep,huang2021longer} and text summarization \cite{barzilay1999using}. In this work, we exploit the commonly used metrics to measure the similarity between BQ and MQ. BLEU (BiLingual Evaluation Understudy) \cite{49} is one of the most popular metrics in machine translation based on precision. Yet, its effectiveness is questioned by some works such as \cite{70,71}. METEOR \cite{69} is based on the harmonic mean of unigram precision and recall, and it can handle the stemming and synonym matching, which is designated to fix problems found with BLEU and produces a better correlation with translations by human experts. Regarding the difference between METEOR and BLEU, METEOR evaluates the correlation at the sentence and segment level whereas BLEU looks for correlations at the corpus level. ROUGE (Recall Oriented Understudy of Gisting Evaluation) \cite{68} is another popular recall-based metric in the text summarization community, and it tends to reward the longer sentences with the higher recall. CIDEr \cite{67}, a consensus-based metric, rewards a sentence for being similar to the majority of descriptions written by the human expert and this metric is mostly used in the image captioning community. It extends the existing metrics with \emph{tf-idf} weights of \emph{n}-grams between candidate sentence and reference sentence. Sometimes, unnecessary parts of the sentence are also weighted and this leads to ineffective scores. So, CIDEr is an inefficient metric for natural language sentence evaluation in some sense. In our experiments, we take all of the metrics above and our proposed $LASSO$ ranking approach to rank BQs and compare their BQ ranking performance.

\noindent
\vspace{+0.1cm}\textbf{Evaluating Image Captioning}

Some commonly used techniques, such as encoding and decoding, in the image captioning task \cite{1,27,28,48} are also used in the VQA task. In \cite{48}, the authors try to use a language model to combine a set of possible words detected in several regions of the input image, and then generate some description for the image. The authors of \cite{28} exploit a convolutional neural networks model to extract the high-level image features, and then give an LSTM unit these features as the first input. In \cite{1}, the authors propose an algorithm to generate a word at each time step by paying attention to local image regions related to the predicted word at the current time step. In \cite{27}, the authors propose a deep neural networks model to learn how to embed the language and visual information into a common multimodal space. To the best of our knowledge, the existing image captioning algorithms only can generate rough and short descriptions for a given image, and those descriptions tend to be grammatically similar to the ones in the training set. Also, although BLEU is commonly used to evaluate the result of the image captioning task, it isn't the most proper metric to evaluate the quality of the image captioning result because of its innate property.



\noindent
\vspace{+0.1cm}\textbf{Evaluating Visual Question Answering}

In VQA, we have two types of inputs with different modalities including the question sentence and image, so VQA is a multimodal task. In \cite{26,57,58,59,60}, the authors have tried to focus on modeling the interactions between two different embedding spaces. The authors of \cite{26,60} have shown that the bilinear interaction between two embedding spaces is very successful in deep learning for fine-grained classification and multimodal language modeling. In \cite{58}, the authors propose a method, Multimodal Compact Bilinear (MCB) pooling, to compute an outer product between visual and textual features. The authors of \cite{59} propose a tensor-based method, Multimodal Low-rank Bilinear (MLB) pooling, to parameterize the full bilinear interactions between image and question sentence embedding spaces. In \cite{57}, the authors propose another method to efficiently parameterize the bilinear interactions between textual and visual representations, and they also show that MCB and MLB are special cases of their proposed method. In \cite{32}, the authors exploit Recurrent Neural Networks (RNN) and Convolutional Neural Networks (CNN) to build a question generation algorithm, but it sometimes generates questions with invalid grammar.
The authors of \cite{9,33,64} exploit RNN to combine the word and image features for the VQA task. In \cite{30}, the authors have tried to exploit convolutions to group the neighboring features of word and image. Gated Recurrent Unit (GRU) \cite{10} is another variant of RNN, and the authors of \cite{6} use it to encode an input question. Additionally, they introduce a dynamic parameter layer in their CNN model, and the weights of the model are adaptively predicted by the embedded question features. The above VQA methods are all based on accuracy-based datasets. To the best of our knowledge, there is no existing VQA method evaluated by a robustness-based dataset since that kind of dataset does not exist.

\noindent
\vspace{+0.1cm}\textbf{Robustness of Neural Network Models}

Recently, the authors of \cite{61,62,63,kafle2017visual,kafle2017analysis,huang2019novel,huang2017vqabq,huang2017robustness,huang2017robustnessMS,hu2019silco,huck2018auto,liu2018synthesizing,yang2018novel,di2021dawn} have tried to discuss the robustness issue of deep learning models from the image, \cite{61,62} or text \cite{huang2019novel} point of view. In \cite{61,62}, the authors analyze the robustness of learning models by adding some noise or perturbations into images and observe how the predicted result will be affected. The authors of \cite{moosavi2018robustness} provide theoretical evidence on the existence of a strong relation between small curvature and large
robustness. Moreover, they propose an efficient regularizer that encourages small curvatures and also show that the regularizer leads to a significant boost in robustness of neural networks. To the best of our knowledge, most of the existing works play with adding noise to the image input. In this work, we play with noise added to the text input. We consider the BQs of a given MQ is a kind of noise of the given MQ. Then, we exploit the BQs to do the robustness analysis of VQA models.


\noindent
\vspace{+0.1cm}\textbf{Datasets for Visual Question Answering}



Recently, many accuracy-based VQA datasets have been proposed. To the best of our knowledge, DAQUAR (DAtaset for QUestion Answering on Real-world images) dataset \cite{2} is the first proposed dataset, which contains about 12.5 thousand manually annotated question-answer pairs on about 1449 indoor scenes \cite{84}. A question in the original DAQUAR dataset only has a single ground truth answer. The authors of \cite{64} collect additional answers for each question to extend the DAQUAR. After the introduction of DAQUAR, three other VQA datasets based on MS-COCO \cite{51} have been proposed, namely \cite{85,4,33}. The authors of \cite{85} have transformed existing annotations for the image caption generation task into question-answer pairs based on a syntactic parser \cite{8} and a set of hand-designed rules. In \cite{4}, the authors have proposed another popular dataset, called VQA. It contains around $614000$ questions concerning the visual content of $205000$ real-world images. Also, it has $150000$ questions based on $50000$ abstract scenes. Additionally, the authors of the VQA dataset provide $10$ answers for each image. The VQA test set answers are not released because of the VQA challenge workshop. Finally, the authors of \cite{33} have annotated about $158000$ images with $316000$ Chinese question-answer pairs with the corresponding English translations. In \cite{90}, the authors try to simplify the evaluation of the performance of VQA models by introducing Visual Madlibs, a multiple choice question answering (QA) by filling the blanks task. In the task, a VQA model has to choose one out of four provided answers based on a given image and the prompt. Formulating VQA task in this way wipes out ambiguities in answer candidates. A simple accuracy metric is used to measure the performance of different VQA models. However, VQA models require holistic reasoning based on the given images in this task. It remains challenging for machines, despite the simple evaluation. In \cite{2,77,87}, the automatic and simple performance evaluation metrics have been a part of building the VQA dataset. The authors of the Visual7W dataset \cite{34} have built question-answer pairs based on the Visual Genome dataset \cite{91}, and it contains around $330000$ natural language questions. In contrast to the other datasets such as VQA or DAQUAR, the Visual Genome dataset focuses on the so-called six Ws, namely \textit{what, where, when, who, why,} and \textit{how}, which can be answered with a text-based sentence. Additionally, Visual7W extends question and answer pairs with extra groundings of the correspondences, and it not only includes natural language answers but also answers requiring locating the object. Then, the Visual7W contains multiple-choice answers similar to Visual Madlibs \cite{90}. In \cite{92}, the authors have proposed Xplore-M-Ego, which is a dataset of images with natural language queries, a media retrieval system, and collective memories. Their work focuses on a user-centric, dynamic scenario, where the given answers are conditioned not only on questions, but also on the geographical position of the questioner. There is another task, called video question answering, which is related to VQA. It needs to understand long term relations in the video. In \cite{89}, the authors have proposed a task which needs to fill in blanks in captions associated with videos. The task requires inferring the past, describing the present and predicting the future in a diverse set of video descriptions data ranging from movies \cite{89,94,95} and cooking videos \cite{93} to web videos \cite{ji2019query}. However, the above datasets are accuracy-based and they cannot be used in the evaluation of the robustness of VQA models. In this work, we propose GBQD and YNBQD robustness-based datasets.

\section{Methodology}



In this section, we introduce our proposed method. We start with a discussion on how to embed questions and use different metrics to generate BQs. Then, we discuss how to analyze the robustness of six pretrained state-of-the-art VQA models by BQ. The overall method is illustrated in Figure \ref{fig:figure100}. It consists of two main components, dubbed the VQA module and Noise Generator respectively. The VQA module contains the model we want to do robustness analysis on, while the Noise Generator utilizes eight ranking methods, namely BLEU-1, BLEU-2, BLEU-3, BLEU-4, ROUGE, CIDEr, METEOR, and our proposed $LASSO$ ranking method, to generate noise for a given main question. According to our hypothesis mentioned in the \textit{Introduction}, a set of effectively ranked BQs based on some ranking method should have a decreasing accuracy.
Let us first introduce some basic notations for our method.

\noindent
\vspace{+0.1cm}\textbf{Question Encoding}

The first step in our method is the embedding of the question sentences. Let $w_{i}^{1},...,w_{i}^{N}$ be the words in question $q_{i}$, with $w_{i}^{t}$ denoting the $t$-th word for $q_{i}$ and $\mathbf{x}_{i}^t$ denoting the $t$-th word embedding for $q_{i}$. Word2Vec \cite{47}, GloVe \cite{11} and Skip-thoughts \cite{43} are popular text encoders \cite{huang2020query,huang2021gpt2mvs}. Since we define a BQ as a question semantically similar to the given MQ, we need an encoder that can better capture the semantic meaning of a sentence. Among these encoders, Skip-thoughts focuses on the semantic meaning of the whole sentence, capturing relations between words. So, we use  Skip-thoughts to embed the questions in this paper. The Skip-thoughts model exploits an RNN encoder with GRU \cite{10} activations, which maps an English sentence, {\em i.e.}, $q_{i}$, into a feature vector $\mathbf{v} \in \mathbf{R}^{4800}$. We encode all the training and validation questions of the VQA dataset \cite{4} into the columns of $\mathbf{A}$, a matrix of Skip-thoughts embedded basic question candidates. We use $\mathbf{b}$ to denote a Skip-thoughts encoded main question.

The question encoder at each time step generates a hidden state $\mathbf{h}_{i}^{t}$. It can be considered as the representation of the sequence \{$w_{i}^{1},..., w_{i}^{t}$\}. So, the hidden state $\mathbf{h}_{i}^{N}$ represents the whole sequence $\{ w_{i}^{1},...,w_{i}^{t},...,w_{i}^{N}\}$, {\em i.e.}, a question sentence in our case. For convenience, here we drop the index $i$ and iterate the following sequential equations to encode a question:
\begin{equation}
    ~~~~~~~~~~~~~~~~~\mathbf{r}^{t}~=~\sigma (\mathbf{U}_{r}\mathbf{h}^{t-1}+\mathbf{W}_{r}\mathbf{x}^{t})
\end{equation}
\begin{equation}
    ~~~~~~~~~~~~~~~~~\mathbf{z}^{t}~=~\sigma(\mathbf{U}_{z}\mathbf{h}^{t-1}+\mathbf{W}_{z}\mathbf{x}^{t})
\end{equation} 
\begin{equation}
    ~~~~~~~~~~~~\bar{\mathbf{h}}^{t}~=~\mathrm{tanh}(\mathbf{U}(\mathbf{r}^{t}\odot \mathbf{h}^{t-1})+\mathbf{Wx}^{t})
\end{equation} 
\begin{equation}
    ~~~~~~~~~~~~~~\mathbf{h}^{t}~=~\mathbf{z}^{t}\odot  \bar{\mathbf{h}}^{t}+(1-\mathbf{z}^{t})\odot \mathbf{h}^{t-1}, 
\end{equation}

\noindent
where $\mathbf{U}_{r}$, $\mathbf{U}_{z}$, $\mathbf{W}_{r}$, $\mathbf{W}_{z}$, $\mathbf{U}$ and $\mathbf{W}$ are the matrices of weight parameters. $\bar{\mathbf{h}}^{t}$ is the state update at time step $t$, $\mathbf{r}^{t}$ is the reset gate, $\odot$ denotes an element-wise product, $\mathbf{z}^{t}$ is the update gate, and ${\mathbf{h}}^{t}=0$ as $t=0$. These two gates take values between zero and one. Finally, $\sigma$ denotes the activation function.

\noindent
\vspace{+0.1cm}\textbf{Level-controllable Noise Generator}

Based on the assumption mentioned in our \textit{Introduction}, level-controllable noise, {\em i.e.}, BQ, generation will involve similarity-based ranking. As we have mentioned in the \textit{Introduction}, the existing textual similarity measures, such as BLEU, CIDEr, METEOR, and ROUGE, cannot effectively capture the semantic similarity. In this work, we propose a new optimization-based ranking method to address this issue. 
The problem of generating BQs that are similar to an MQ can be cast as a $LASSO$ optimization problem. By embedding all the main questions and the basic question candidates using Skip-thoughts, $LASSO$ modelling helps us to determine a sparse number of basic questions suited to represent the given main question. The $LASSO$ model is expressed by the following optimization:
\vspace{-0.15cm}
\begin{equation}
    ~~~~~~~~~~~~~~~~~\min_{\mathbf{x}}~\frac{1}{2}\left \| \mathbf{A}\mathbf{x}-\mathbf{b} \right \|_{2}^{2}+\lambda \left \| \mathbf{x} \right \|_{1}, 
\label{eq:lasso}
\end{equation}
where $\lambda$ is a tradeoff parameter which controls the quality of BQs.

To develop our basic question dataset (BQD), we combine the unique questions in the training and validation datasets of the most popular VQA dataset \cite{4} and we use the testing dataset as our main question candidates. Also, we need to do ``question sentences preprocessing'', in particular, making sure that none of the main questions is contained in our basic question dataset, as otherwise, $LASSO$ modelling cannot give a useful ranking. Otherwise and because we are encouraging sparsity, the ranking will neglect all other questions and give them a similarity score of zero. 


\noindent
\vspace{+0.1cm}\textbf{BQ Generation by $\mathbf{LASSO}$-based Ranking Method}

In this subsection, we describe how to use the $LASSO$-based ranking method to generate the basic questions of a given main question (refer to Figure \ref{fig:figure100}). Now, we are ready to deal with our $LASSO$ optimization problem to get the sparse solution $\mathbf{x}$. One can consider the elements of $\mathbf{x}$ to be the similarity score between the main question $\mathbf{b}$ and the corresponding BQ in  $\mathbf{A}$. The first embedded BQ candidate is the first column of $\mathbf{A}$ and the corresponding similarity score is the first element of $\mathbf{x}$ and so on. Furthermore, we collect the top-$k$ BQs of each given MQ based on the ranking of scores in $\mathbf{x}$. Intuitively, if a BQ has a higher similarity score to a given query question, it implies that this BQ is more similar to the given MQ and vice versa. Additionally, because most of the VQA models have the highest accuracy performance in answering yes/no questions, we argue that yes/no questions are the simplest questions for VQA models in the sense of accuracy. Hence, we also create a Yes/No Basic Question dataset based on the aforementioned basic question generation approach.


\noindent
\vspace{+0.1cm}\textbf{Details of the Proposed Basic Question Dataset}

The size of the basic questions dataset has a great impact on the noise generation method. Intuitively, the more questions you have, the more chance it has to contain similar questions to any given main question. In our work, based on the ${LASSO}$-based ranking method, we propose two large-scale basic question datasets, General Basic Question Dataset and Yes/No Basic Question Dataset. Note that, in our dataset collections, we set $k=21$ because after top-$21$ the similarity scores of BQs are negligible. As such, we get the ranked BQs of $244,302$ testing question candidates. 
The proposed General and Yes/No BQ datasets, with the format $\{\text{Image},~MQ,~21~(BQ + \text{corresponding~ similarity~score})\}$, contain $81,434$ images from the testing images of MS COCO dataset \cite{51} and $244,302$ main questions from the testing questions of VQA dataset (open-ended task) \cite{4}. Furthermore, our General and Yes/No basic questions are extracted from the validation and training questions of VQA dataset (open-ended task) and the corresponding similarity scores of General and Yes/No BQ are generated by our $LASSO$ ranking approach. That is to say, in our GBQD and YNBQD, there are $5,130,342$ (General BQ + corresponding similarity score) tuplets and $5,130,342$ (Yes/No BQ + corresponding similarity score) tuplets. 



\noindent\vspace{+0.1cm}\textbf{Robustness Analysis by General and Yes/No Basic Questions}

To measure the robustness of any VQA model, we measure how its accuracy changes when its input is corrupted with noise. The noise can be completely random, structured and/or semantically related to the final task. Since the input in VQA is an MQ-image pair, the noise can be injected into both. The noise to the question should have some contextual semantics for the measure to be informative, instead of introducing misspellings or changing or dropping random words. Here we propose a novel robustness measure for VQA by introducing semantically relevant noise to the questions where we are able to control the level of noise.


The authors of VQA dataset \cite{4} provide the open-ended and multiple-choice tasks for evaluation. For the multiple-choice task, an answer should be selected from 18 answer candidates. However, the answer of the open-ended task can be any phrase or word. For both tasks, the answers are evaluated by accuracy, which is considered to reflect human consensus. We measure accuracy as defined in \cite{4} and given by the following:

\begin{equation}
    \label{eq:acc-vqa}
    \text{Accuracy}_{VQA}=\frac{1}{N}\sum_{i=1}^{N}\min\left \{ \frac{\sum_{t\in T_{i}}\mathbb{I}[a_{i}=t]}{3},1 \right \},
\end{equation}
where $\mathbb{I}[\cdot]$ is an indicator function, $N$ is the total number of examples, $a_{i}$ is the predicted answer, and $T_{i}$ is an answer set of the $i^{th}$ image-question pair. That is, a predicted answer is considered correct  if there are  at least 3 annotators who agree with it and the score depends on the total number of agreements when the predicted answer is incorrect.

\begin{equation}
    \label{eq:absolute-difference}
    ~~~~~~~~~~~~~~~~~Acc_{di} = \left|Acc_{vqa}-Acc_{bqd}\right|,
\end{equation}
where $Acc_{vqa}$ and $Acc_{bqd}$ are based on Equation (\ref{eq:acc-vqa}).

\begin{figure}
\begin{center}
   \includegraphics[width=1.0\linewidth]{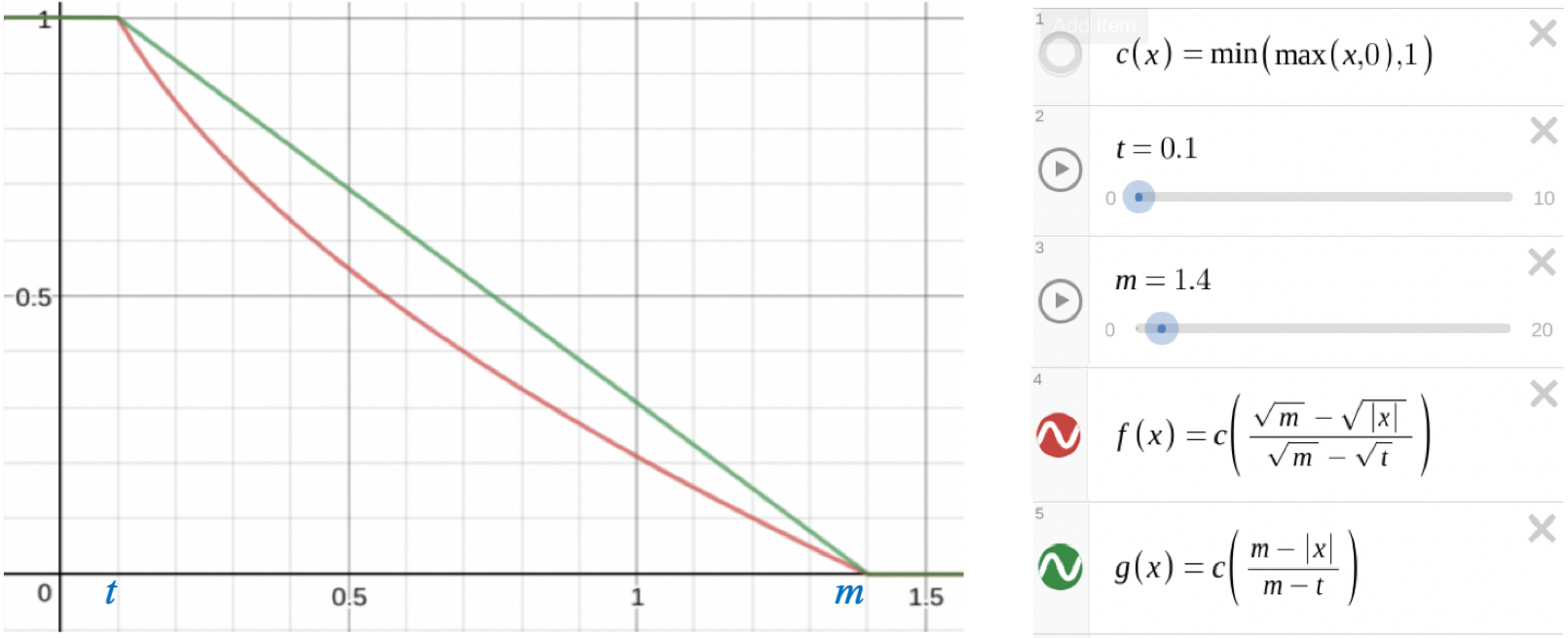}
\end{center}
\vspace{-12pt}
   \caption{The visualization of $R_{score}$ function denoted in red. For convenience, we only plot the right-hand part of the $f(x)$ and $g(x)$. The parameters $t$ and $m$ are the tolerance and maximum robustness limit, respectively. For more explanation, please refer to the \textit{Robustness Analysis by General and Yes/No Basic Questions}.}
\label{fig:figure71}
\end{figure}

To analyze the robustness of a VQA model, we first measure the accuracy of the model on the clean VQA dataset \cite{4} and we call it $Acc_{vqa}$. Then, we append the top ranked $k$ BQs to each of the MQs in the clean dataset and recompute the accuracy of the model on this noisy input and we call it $Acc_{bqd}$. Finally, we compute the absolute difference, $Acc_{di}$, based on Equation (\ref{eq:absolute-difference}) and we report the robustness score $R_{score}$. The parameters $t$ and $m$ in Equation (\ref{eq:robustness-cleaner}) are the tolerance and maximum robustness limit, respectively. We want the score to be sensitive if the difference is small, but not before $t$, and less sensitive if it is large, but not after $m$. So, the robustness score $R_{score}$ is designed to decrease smoothly between $1$ and $0$ as $Acc_{di}$ moves from $t$ to $m$ and remains constant outside this range. The rate of change of this transition is exponentially decreasing from exponential to sublinear in the range $[t, m]$. 

\begin{equation}
    \label{eq:robustness-cleaner}
    ~~~~~~~~~~~~R_{score} = clamp_{0}^{1}\left(\frac{\sqrt{m}-\sqrt{{Acc_{di}}}}{\sqrt{m}-\sqrt{t}}\right)
\end{equation}

\begin{equation}
    \label{eq:robustness-clamp}
    ~~~~~~~~~~~~~~~~clamp_{a}^{b}(x) = \max\left(a,{\min\left(b,x\right)}\right),
\end{equation}
where $0 \leq t < m \leq 100$. To make the above clearer, we visualize the $R_{score}$ function in Figure \ref{fig:figure71}.

\section{Experiments}
In this section, we explain our implementation details and the experiments we conducted to validate and analyze our proposed method.


\begin{table*}
\centering 
\begin{subtable}{0.75\linewidth}
\resizebox{\columnwidth}{110.0pt}{
\scalebox{1}{
\begin{tabular}{|c|c|c|}
\hline
BQ ID & \multicolumn{1}{l|}{Similarity Score} & BQ \\ \hline
\begin{tabular}[c]{@{}c@{}}01\\
02\\
03\end{tabular}  & \begin{tabular}[c]{@{}c@{}}0.295\\
0.240\\
0.142\end{tabular} & \begin{tabular}[c]{@{}c@{}}How old is the truck? \\    
How old is this car? \\   
How old is the vehicle?\end{tabular} \\
\hline
\begin{tabular}[c]{@{}c@{}}04\\
05\\ 
06\end{tabular} & \begin{tabular}[c]{@{}c@{}}0.120\\
0.093\\ 
0.063\end{tabular} & \begin{tabular}[c]{@{}c@{}}What number is the car?\\ 
What color is the car?\\ 
How old is the bedroom?\end{tabular}\\
\hline
\begin{tabular}[c]{@{}c@{}}07\\
08\\ 
09\end{tabular} & \begin{tabular}[c]{@{}c@{}}0.063\\ 
0.037\\ 
0.033\end{tabular} & \begin{tabular}[c]{@{}c@{}}What year is the car?\\
Where is the old car?\\
How old is the seat?\end{tabular}\\ 
\hline
\begin{tabular}[c]{@{}c@{}}10\\
11\\ 
12\end{tabular}  & \begin{tabular}[c]{@{}c@{}}0.032\\ 
0.028\\ 
0.028\end{tabular} & \begin{tabular}[c]{@{}c@{}}How old is the cart?\\ 
What make is the blue car?\\ 
How old is the golden retriever?\end{tabular}\\ 
\hline
\begin{tabular}[c]{@{}c@{}}13\\
14\\ 
15\end{tabular}  & \begin{tabular}[c]{@{}c@{}}0.024\\ 
0.022\\ 
0.020\end{tabular} & \begin{tabular}[c]{@{}c@{}}What is beneath the car?\\ 
Is the car behind him a police car?\\ 
How old is the pilot?\end{tabular}\\ 
\hline
\begin{tabular}[c]{@{}c@{}}16\\
17\\ 
18\end{tabular} & \begin{tabular}[c]{@{}c@{}}0.017\\ 0.016\\ 
0.016\end{tabular} & \begin{tabular}[c]{@{}c@{}}How old are you?\\ 
How old is the laptop?\\ 
How old is the television?\end{tabular}\\ 
\hline
\begin{tabular}[c]{@{}c@{}}19\\
20\\ 
21\end{tabular} & \begin{tabular}[c]{@{}c@{}}0.015\\ 0.015\\ 
0.015\end{tabular} & \begin{tabular}[c]{@{}c@{}}What make is the main car?\\ 
What type and model is the car?\\ 
What is lifting the car?\end{tabular}\\ 
\hline

\end{tabular}}}
\caption{``MQ: How old is the car?'' and image ``(a)'' in Figure \ref{fig:figure10}.}
\label{table:table51}

\vspace{0.2cm}

\resizebox{\columnwidth}{130.0pt}{
\scalebox{1}{
\begin{tabular}{|c|c|c|}
\hline
BQ ID  & \multicolumn{1}{l|}{Similarity Score} & BQ \\ \hline
\begin{tabular}[c]{@{}c@{}}01\\
02\\
03\end{tabular}  & \begin{tabular}[c]{@{}c@{}}0.281\\
0.108\\
0.055\end{tabular} & \begin{tabular}[c]{@{}c@{}}Where is the cat sitting on? \\    
What is this cat sitting on? \\   
What is cat sitting on?\end{tabular} \\
\hline
\begin{tabular}[c]{@{}c@{}}04\\
05\\ 
06\end{tabular} & \begin{tabular}[c]{@{}c@{}}0.053\\
0.050\\ 
0.047\end{tabular} & \begin{tabular}[c]{@{}c@{}}What is the cat on the left sitting on?\\ 
What is the giraffe sitting on?\\ 
What is the cat sitting in the car?\end{tabular}\\
\hline
\begin{tabular}[c]{@{}c@{}}07\\
08\\ 
09\end{tabular} & \begin{tabular}[c]{@{}c@{}}0.046\\ 
0.042\\ 
0.041\end{tabular} & \begin{tabular}[c]{@{}c@{}}That is the black cat sitting on?\\
What is the front cat sitting on?\\
What is the cat perched on?\end{tabular}\\ 
\hline
\begin{tabular}[c]{@{}c@{}}10\\
11\\ 
12\end{tabular}  & \begin{tabular}[c]{@{}c@{}}0.041\\ 
0.037\\ 
0.035\end{tabular} & \begin{tabular}[c]{@{}c@{}}What's the cat sitting on? \\ 
What is the cat leaning on? \\ 
What object is the cat sitting on?\end{tabular}\\ 
\hline
\begin{tabular}[c]{@{}c@{}}13\\
14\\ 
15\end{tabular}  & \begin{tabular}[c]{@{}c@{}}0.029\\ 
0.023\\ 
0.022\end{tabular} & \begin{tabular}[c]{@{}c@{}}What is the doll sitting on?\\ 
How is the cat standing?\\ 
What is the cat setting on?\end{tabular}\\ 
\hline
\begin{tabular}[c]{@{}c@{}}16\\
17\\ 
18\end{tabular} & \begin{tabular}[c]{@{}c@{}}0.022\\ 0.021\\ 
0.021\end{tabular} & \begin{tabular}[c]{@{}c@{}}What is the cat walking on?\\ 
What is the iPhone sitting on?\\ 
What is the cat napping on?\end{tabular}\\ 
\hline
\begin{tabular}[c]{@{}c@{}}19\\
20\\ 
21\end{tabular} & \begin{tabular}[c]{@{}c@{}}0.020\\ 0.018\\ 
0.018\end{tabular} & \begin{tabular}[c]{@{}c@{}}What is the dog sitting at?\\ 
What is the birds sitting on?\\ 
What is the sitting on?\end{tabular}\\ 
\hline

\end{tabular}}}
\caption{``MQ: What is the cat sitting on?'' and image ``(b)'' in Figure \ref{fig:figure10}.}
\label{table:table52}

\end{subtable}
\caption{``MQ: How old is the car?'' and image ``(a)'' corresponds to Figure \ref{fig:figure10}-(a). ``MQ: What is the cat sitting on?'' and image ``(b)'' corresponds to Figure \ref{fig:figure10}-(b).}
\label{table:table5}
\vspace{-0.5cm}
\end{table*}


\begin{table*}
\renewcommand\arraystretch{1.28}
\setlength\tabcolsep{11pt}
    \centering
\begin{subtable}[t]{0.3\linewidth}
\centering
\scalebox{0.53}{
    \begin{tabular}{ c | c c c c | c} 
     Task Type &    & \multicolumn{3}{c}{Open-Ended} &  \\ [0.5ex]
     \hline
     Method &    & \multicolumn{3}{c}{MUTAN without Attention} &  \\ [0.5ex]
     \hline
     Test Set&  \multicolumn{4}{c}{dev} & diff \\ [0.5ex]
     \hline
     Partition & Other & Num & Y/N & All & All \\ [0.5ex] 
     \hline
     First-dev & 37.78 & 34.93 & 68.20 & \textbf{49.96} & \textbf{10.20}  \\ 
     
     Second-dev & 37.29 & 35.03 & 65.62 & \textbf{48.67} & \textbf{11.49} \\
     
     Third-dev & 34.81 & 34.39 & 62.85 & \textbf{46.27} & \textbf{13.89}  \\
     
     Fourth-dev & 34.25 & 34.29 & 63.60 & \textbf{46.30} & \textbf{13.86}  \\
     
     Fifth-dev & 33.89 & 34.66 & 64.19 & \textbf{46.41} & \textbf{13.75} \\
     
     Sixth-dev & 33.15 & 34.68 & 64.59 & \textbf{46.22} & \textbf{13.94}  \\
     
     Seventh-dev & 32.80 & 33.99 & 63.59 & \textbf{45.57} & \textbf{14.59}  \\
     \hline
     First-std & 38.24 & 34.54 & 67.55 & \textbf{49.93} & \textbf{10.52} \\
     \hline
     
     Original-dev  & 47.16 & 37.32 & 81.45 & \textbf{60.16} & -\\
     Original-std  & 47.57 & 36.75 & 81.56 & \textbf{60.45} & - \\

     \hline
    \end{tabular}}
    \centering
    \captionsetup{justification=centering}
    \caption{MUTAN without Attention model evaluation results.}

\scalebox{0.53}{
    \begin{tabular}{ c | c c c c | c} 
     Task Type &    & \multicolumn{3}{c}{Open-Ended} &  \\ [0.5ex]
     \hline
     Method &    & \multicolumn{3}{c}{HieCoAtt (Alt,VGG19)} &  \\ [0.5ex]
     \hline
     Test Set&  \multicolumn{4}{c}{dev} & diff \\ [0.5ex]
     \hline
     Partition & Other & Num & Y/N & All & All \\ [0.5ex] 
     \hline
     First-dev & 44.44 & 37.53 & 71.11 & \textbf{54.63} & \textbf{5.85}  \\ 
     
     Second-dev & 42.62 & 36.68 & 68.67 & \textbf{52.67} & \textbf{7.81} \\
     
     Third-dev & 41.60 & 35.59 & 66.28 & \textbf{51.08} & \textbf{9.4}  \\
     
     Fourth-dev & 41.09 & 35.71 & 67.49 & \textbf{51.34} & \textbf{9.14}  \\
     
     Fifth-dev & 39.83 & 35.55 & 65.72 & \textbf{49.99} & \textbf{10.49} \\
     
     Sixth-dev & 39.60 & 35.99 & 66.56 & \textbf{50.27} & \textbf{10.21}  \\
     
     Seventh-dev & 38.33 & 35.47 & 64.89 & \textbf{48.92} & \textbf{11.56}  \\
     \hline
     First-std & 44.77 & 36.08 & 70.67 & \textbf{54.54} & \textbf{5.78} \\
     \hline

     Original-dev  & 49.14 & 38.35 & 79.63 & \textbf{60.48} & -\\
     Original-std  & 49.15 & 36.52 & 79.45 & \textbf{60.32} & - \\

     \hline
    \end{tabular}}
    \centering
    \captionsetup{justification=centering}
    \caption{HieCoAtt (Alt,VGG19) model evaluation results.}
\end{subtable}%
    \hfil
\begin{subtable}[t]{0.3\linewidth}

\centering
\scalebox{0.53}{
    \begin{tabular}{ c | c c c c | c} 
     Task Type &    & \multicolumn{3}{c}{Open-Ended} &  \\ [0.5ex]
     \hline
     Method &    & \multicolumn{3}{c}{MLB with Attention} &  \\ [0.5ex]
     \hline
     Test Set&  \multicolumn{4}{c}{dev} & diff \\ [0.5ex]
     \hline
     Partition & Other & Num & Y/N & All & All \\ [0.5ex] 
     \hline
     First-dev & 49.31 & 34.62 & 72.21 & \textbf{57.12} & \textbf{8.67}  \\ 
     
     Second-dev & 48.53 & 34.84 & 70.30 & \textbf{55.98} & \textbf{9.81} \\
     
     Third-dev & 48.01 & 33.95 & 69.15 & \textbf{55.16} & \textbf{10.63}  \\
     
     Fourth-dev & 47.20 & 34.02 & 69.31 & \textbf{54.84} & \textbf{10.95}  \\
     
     Fifth-dev & 45.85 & 34.07 & 68.95 & \textbf{54.05} & \textbf{11.74} \\
     
     Sixth-dev & 44.61 & 34.30 & 68.59 & \textbf{53.34} & \textbf{12.45}  \\
     
     Seventh-dev & 44.71 & 33.84 & 67.76 & \textbf{52.99} & \textbf{12.80}  \\
     \hline
     First-std & 49.07 & 34.13 & 71.96 & \textbf{56.95} & \textbf{8.73} \\
     \hline
     
     Original-dev  & 57.01 & 37.51 & 83.54 & \textbf{65.79} & -\\
     Original-std  & 56.60 & 36.63 & 83.68 & \textbf{65.68} & - \\

     \hline
    \end{tabular}}
    \centering
    \captionsetup{justification=centering}
    \caption{MLB with Attention model evaluation results.}
    
\scalebox{0.53}{
    \begin{tabular}{ c | c c c c | c} 
     Task Type &    & \multicolumn{3}{c}{Open-Ended} &  \\ [0.5ex]
     \hline
     Method &    & \multicolumn{3}{c}{MUTAN with Attention} &  \\ [0.5ex]
     \hline
     Test Set&  \multicolumn{4}{c}{dev} & diff \\ [0.5ex]
     \hline
     Partition & Other & Num & Y/N & All & All \\ [0.5ex] 
     \hline
     First-dev & 51.51 & 35.62 & 68.72 & \textbf{56.85} & \textbf{9.13}  \\ 
     
     Second-dev & 49.86 & 34.43 & 66.18 & \textbf{54.88} & \textbf{11.10} \\
     
     Third-dev & 49.15 & 34.50 & 64.85 & \textbf{54.00} & \textbf{11.98}  \\
     
     Fourth-dev & 47.96 & 34.26 & 64.72 & \textbf{53.35} & \textbf{12.63}  \\
     
     Fifth-dev & 47.20 & 33.93 & 64.53 & \textbf{52.88} & \textbf{13.10} \\
     
     Sixth-dev & 46.48 & 33.90 & 64.37 & \textbf{52.46} & \textbf{13.52}  \\
     
     Seventh-dev & 46.88 & 33.13 & 64.10 & \textbf{52.46} & \textbf{13.52}  \\
     \hline
     First-std & 51.34 & 35.22 & 68.32 & \textbf{56.66} & \textbf{9.11} \\
     \hline
     
     Original-dev  & 56.73 & 38.35 & 84.11 & \textbf{65.98} & -\\
     Original-std  & 56.29 & 37.47 & 84.04 & \textbf{65.77} & - \\

     \hline
    \end{tabular}}
    \centering
    \captionsetup{justification=centering}
    \caption{MUTAN with Attention model evaluation results.}
\end{subtable}%
    \hfil
\begin{subtable}[t]{0.3\linewidth}
        
\centering
\scalebox{0.53}{
    \begin{tabular}{ c | c c c c | c} 
     Task Type &    & \multicolumn{3}{c}{Open-Ended} &  \\ [0.5ex]
     \hline
     Method &    & \multicolumn{3}{c}{HieCoAtt (Alt,Resnet200)} &  \\ [0.5ex]
     \hline
     Test Set&  \multicolumn{4}{c}{dev} & diff \\ [0.5ex]
     \hline
     Partition & Other & Num & Y/N & All & All \\ [0.5ex] 
     \hline
     First-dev & 46.51 & 36.33 & 70.41 & \textbf{55.22} & \textbf{6.59}  \\ 
     
     Second-dev & 45.19 & 36.78 & 67.27 & \textbf{53.34} & \textbf{8.47} \\
     
     Third-dev & 43.87 & 36.28 & 65.29 & \textbf{51.84} & \textbf{9.97}  \\
     
     Fourth-dev & 43.41 & 36.25 & 65.94 & \textbf{51.88} & \textbf{9.93}  \\
     
     Fifth-dev & 42.02 & 35.89 & 66.09 & \textbf{51.23} & \textbf{10.58} \\
     
     Sixth-dev & 42.03 & 36.40 & 65.66 & \textbf{51.12} & \textbf{10.69}  \\
     
     Seventh-dev & 40.68 & 36.08 & 63.49 & \textbf{49.54} & \textbf{12.27}  \\
     \hline
     First-std & 46.77 & 35.22 & 69.66 & \textbf{55.00} & \textbf{7.06}\\
     \hline
     
     Original-dev  & 51.77 & 38.65 & 79.70 & \textbf{61.81} & -\\
     Original-std  & 51.95 & 38.22 & 79.95 & \textbf{62.06} & - \\

     \hline
    \end{tabular}}
    \centering
    \captionsetup{justification=centering}
    \caption{HieCoAtt (Alt,Resnet200) model evaluation results.}

\scalebox{0.53}{
    \begin{tabular}{ c | c c c c | c} 
     Task Type &    & \multicolumn{3}{c}{Open-Ended} &  \\ [0.5ex]
     \hline
     Method &    & \multicolumn{3}{c}{LSTM Q+I} &  \\ [0.5ex]
     \hline
     Test Set&  \multicolumn{4}{c}{dev} & diff \\ [0.5ex]
     \hline
     Partition & Other & Num & Y/N & All & All \\ [0.5ex] 
     \hline
     First-dev & 29.24 & 33.77 & 65.14 & \textbf{44.47} & \textbf{13.55}  \\ 
     
     Second-dev & 28.02 & 32.73 & 62.68 & \textbf{42.75} & \textbf{15.27} \\
     
     Third-dev & 26.32 & 33.10 & 60.22 & \textbf{40.97} & \textbf{17.05}  \\
     
     Fourth-dev & 25.27 & 31.70 & 61.56 & \textbf{40.86} & \textbf{17.16}  \\
     
     Fifth-dev & 24.73 & 32.63 & 61.55 & \textbf{40.70} & \textbf{17.32} \\
     
     Sixth-dev & 23.90 & 32.14 & 61.42 & \textbf{40.19} & \textbf{17.83}  \\
     
     Seventh-dev & 22.74 & 31.36 & 60.60 & \textbf{39.21} & \textbf{18.81}  \\
     \hline
     First-std & 29.68 & 33.76 & 65.09 & \textbf{44.70} & \textbf{13.48} \\
     \hline
     
     Original-dev  & 43.40 & 36.46 & 80.87 & \textbf{58.02} & -\\
     Original-std  & 43.90 & 36.67 & 80.38 & \textbf{58.18} & - \\

     \hline
    \end{tabular}}
    \centering
    \captionsetup{justification=centering}
    \caption{LSTM Q+I model evaluation results.}
\end{subtable}
\caption{The table shows the six state-of-the-art pretrained VQA models evaluation results on the GBQD and VQA dataset. ``-'' indicates the results are not available, ``-std'' represents the accuracy of VQA model evaluated on the complete testing set of GBQD and VQA dataset and ``-dev'' indicates the accuracy of VQA model evaluated on the partial testing set of GBQD and VQA dataset. In addition, $diff = Original_{dev_{All}} - X_{dev_{All}}$, where $X$ is equal to the ``First'', ``Second'', etc.}
\label{table:table6}
\end{table*}

\begin{table*}
\renewcommand\arraystretch{1.28}
\setlength\tabcolsep{11pt}
    \centering
\begin{subtable}[t]{0.3\linewidth}
\centering
\scalebox{0.53}{
    \begin{tabular}{ c | c c c c | c} 
     Task Type &    & \multicolumn{3}{c}{Open-Ended} &  \\ [0.5ex]
     \hline
     Method &    & \multicolumn{3}{c}{MUTAN without Attention} &  \\ [0.5ex]
     \hline
     Test Set&  \multicolumn{4}{c}{dev} & diff \\ [0.5ex]
     \hline
     Partition & Other & Num & Y/N & All & All \\ [0.5ex] 
     \hline
     First-dev & 33.98 & 33.50 & 73.22 & \textbf{49.96} & \textbf{10.13}  \\ 
     
     Second-dev & 32.44 & 34.47 & 72.22 & \textbf{48.67} & \textbf{11.18} \\
     
     Third-dev & 32.65 & 33.60 & 71.76 & \textbf{46.27} & \textbf{11.36}  \\
     
     Fourth-dev & 32.77 & 33.79 & 71.14 & \textbf{46.30} & \textbf{11.53}  \\
     
     Fifth-dev & 32.46 & 33.51 & 70.90 & \textbf{46.41} & \textbf{11.81} \\
     
     Sixth-dev & 33.02 & 33.18 & 69.88 & \textbf{46.22} & \textbf{12.00}  \\
     
     Seventh-dev & 32.73 & 33.28 & 69.74 & \textbf{45.57} & \textbf{12.18}  \\
     \hline
     First-std & 34.06 & 33.24 & 72.99 & \textbf{49.93} & \textbf{10.43} \\
     \hline
     
     Original-dev  & 47.16 & 37.32 & 81.45 & \textbf{60.16} & -\\
     Original-std  & 47.57 & 36.75 & 81.56 & \textbf{60.45} & - \\

     \hline
    \end{tabular}}
    \centering
    \captionsetup{justification=centering}
    \caption{MUTAN without Attention model evaluation results.}

\scalebox{0.53}{
    \begin{tabular}{ c | c c c c | c} 
     Task Type &    & \multicolumn{3}{c}{Open-Ended} &  \\ [0.5ex]
     \hline
     Method &    & \multicolumn{3}{c}{HieCoAtt (Alt,VGG19)} &  \\ [0.5ex]
     \hline
     Test Set&  \multicolumn{4}{c}{dev} & diff \\ [0.5ex]
     \hline
     Partition & Other & Num & Y/N & All & All \\ [0.5ex] 
     \hline
     First-dev & 40.80 & 30.34 & 76.92 & \textbf{54.49} & \textbf{5.99}  \\ 
     
     Second-dev & 39.63 & 30.67 & 76.49 & \textbf{53.78} & \textbf{6.70} \\
     
     Third-dev & 39.33 & 31.12 & 75.48 & \textbf{53.28} & \textbf{7.20}  \\
     
     Fourth-dev & 39.31 & 29.78 & 75.12 & \textbf{52.97} & \textbf{7.51}  \\
     
     Fifth-dev & 39.38 & 29.87 & 74.96 & \textbf{52.95} & \textbf{7.53} \\
     
     Sixth-dev & 39.13 & 30.74 & 73.95 & \textbf{52.51} & \textbf{7.97}  \\
     
     Seventh-dev & 38.90 & 31.14 & 73.80 & \textbf{52.39} & \textbf{8.09}  \\
     \hline
     First-std & 40.88 & 28.82 & 76.67 & \textbf{54.37} & \textbf{5.95} \\
     \hline

     Original-dev  & 49.14 & 38.35 & 79.63 & \textbf{60.48} & -\\
     Original-std  & 49.15 & 36.52 & 79.45 & \textbf{60.32} & - \\

     \hline
    \end{tabular}}
    \centering
    \captionsetup{justification=centering}
    \caption{HieCoAtt (Alt,VGG19) model evaluation results.}
\end{subtable}%
    \hfil
\begin{subtable}[t]{0.3\linewidth}

\centering
\scalebox{0.53}{
    \begin{tabular}{ c | c c c c | c} 
     Task Type &    & \multicolumn{3}{c}{Open-Ended} &  \\ [0.5ex]
     \hline
     Method &    & \multicolumn{3}{c}{MLB with Attention} &  \\ [0.5ex]
     \hline
     Test Set&  \multicolumn{4}{c}{dev} & diff \\ [0.5ex]
     \hline
     Partition & Other & Num & Y/N & All & All \\ [0.5ex] 
     \hline
     First-dev & 46.57 & 32.09 & 76.60 & \textbf{57.33} & \textbf{8.46}  \\ 
     
     Second-dev & 45.83 & 32.43 & 75.29 & \textbf{56.47} & \textbf{9.32} \\
     
     Third-dev & 45.17 & 32.52 & 74.87 & \textbf{55.99} & \textbf{9.80}  \\
     
     Fourth-dev & 45.11 & 32.31 & 73.73 & \textbf{55.47} & \textbf{10.32}  \\
     
     Fifth-dev & 44.35 & 31.95 & 72.93 & \textbf{54.74} & \textbf{11.05} \\
     
     Sixth-dev & 43.75 & 31.21 & 72.03 & \textbf{54.00} & \textbf{11.79}  \\
     
     Seventh-dev & 43.88 & 32.59 & 71.99 & \textbf{54.19} & \textbf{11.60}  \\
     \hline
     First-std & 46.11 & 31.46 & 76.84 & \textbf{57.25} & \textbf{8.43} \\
     \hline
     
     Original-dev  & 57.01 & 37.51 & 83.54 & \textbf{65.79} & -\\
     Original-std  & 56.60 & 36.63 & 83.68 & \textbf{65.68} & - \\

     \hline
    \end{tabular}}
    \centering
    \captionsetup{justification=centering}
    \caption{MLB with Attention model evaluation results.}
    
\scalebox{0.53}{
    \begin{tabular}{ c | c c c c | c} 
     Task Type &    & \multicolumn{3}{c}{Open-Ended} &  \\ [0.5ex]
     \hline
     Method &    & \multicolumn{3}{c}{MUTAN with Attention} &  \\ [0.5ex]
     \hline
     Test Set&  \multicolumn{4}{c}{dev} & diff \\ [0.5ex]
     \hline
     Partition & Other & Num & Y/N & All & All \\ [0.5ex] 
     \hline
     First-dev & 43.96 & 28.90 & 71.89 & \textbf{53.79} & \textbf{12.19}  \\ 
     
     Second-dev & 42.66 & 28.08 & 70.05 & \textbf{52.32} & \textbf{13.66} \\
     
     Third-dev & 41.62 & 29.12 & 69.58 & \textbf{51.74} & \textbf{14.24}  \\
     
     Fourth-dev & 41.53 & 29.30 & 67.96 & \textbf{51.06} & \textbf{14.92}  \\
     
     Fifth-dev & 40.46 & 27.66 & 68.03 & \textbf{50.39} & \textbf{15.59} \\
     
     Sixth-dev & 40.03 & 28.44 & 66.98 & \textbf{49.84} & \textbf{16.14}  \\
     
     Seventh-dev & 39.11 & 28.41 & 67.44 & \textbf{49.58} & \textbf{16.40}  \\
     \hline
     First-std & 43.55 & 28.70 & 71.76 & \textbf{53.63} & \textbf{12.14} \\
     \hline
     
     Original-dev  & 56.73 & 38.35 & 84.11 & \textbf{65.98} & -\\
     Original-std  & 56.29 & 37.47 & 84.04 & \textbf{65.77} & - \\

     \hline
    \end{tabular}}
    \centering
    \captionsetup{justification=centering}
    \caption{MUTAN with Attention model evaluation results.}
\end{subtable}%
    \hfil
\begin{subtable}[t]{0.3\linewidth}
        
\centering
\scalebox{0.53}{
    \begin{tabular}{ c | c c c c | c} 
     Task Type &    & \multicolumn{3}{c}{Open-Ended} &  \\ [0.5ex]
     \hline
     Method &    & \multicolumn{3}{c}{HieCoAtt (Alt,Resnet200)} &  \\ [0.5ex]
     \hline
     Test Set&  \multicolumn{4}{c}{dev} & diff \\ [0.5ex]
     \hline
     Partition & Other & Num & Y/N & All & All \\ [0.5ex] 
     \hline
     First-dev & 44.42 & 36.39 & 76.94 & \textbf{56.90} & \textbf{4.91}  \\ 
     
     Second-dev & 43.37 & 34.99 & 76.10 & \textbf{55.90} & \textbf{5.91} \\
     
     Third-dev & 42.22 & 33.97 & 75.80 & \textbf{55.11} & \textbf{6.70}  \\
     
     Fourth-dev & 42.52 & 34.21 & 75.33 & \textbf{55.09} & \textbf{6.72}  \\
     
     Fifth-dev & 42.81 & 34.69 & 75.21 & \textbf{55.23} & \textbf{6.58} \\
     
     Sixth-dev & 42.27 & 35.16 & 74.50 & \textbf{54.73} & \textbf{7.08}  \\
     
     Seventh-dev & 41.95 & 35.14 & 73.64 & \textbf{54.22} & \textbf{7.59}  \\
     \hline
     First-std & 44.93 & 35.59 & 76.82 & \textbf{57.10} & \textbf{4.96}\\
     \hline

     Original-dev  & 51.77 & 38.65 & 79.70 & \textbf{61.81} & -\\
     Original-std  & 51.95 & 38.22 & 79.95 & \textbf{62.06} & - \\

     \hline
    \end{tabular}}
    \centering
    \captionsetup{justification=centering}
    \caption{HieCoAtt (Alt,Resnet200) model evaluation results.}

\scalebox{0.53}{
    \begin{tabular}{ c | c c c c | c} 
     Task Type &    & \multicolumn{3}{c}{Open-Ended} &  \\ [0.5ex]
     \hline
     Method &    & \multicolumn{3}{c}{LSTM Q+I} &  \\ [0.5ex]
     \hline
     Test Set&  \multicolumn{4}{c}{dev} & diff \\ [0.5ex]
     \hline
     Partition & Other & Num & Y/N & All & All \\ [0.5ex] 
     \hline
     First-dev & 20.49 & 25.98 & 68.79 & \textbf{40.91} & \textbf{17.11}  \\ 
     
     Second-dev & 19.81 & 25.40 & 68.51 & \textbf{40.40} & \textbf{17.62} \\
     
     Third-dev & 18.58 & 24.95 & 68.53 & \textbf{39.77} & \textbf{18.25}  \\
     
     Fourth-dev & 18.50 & 24.82 & 67.83 & \textbf{39.43} & \textbf{18.59}  \\
     
     Fifth-dev & 17.68 & 24.68 & 67.99 & \textbf{39.09} & \textbf{18.93} \\
     
     Sixth-dev & 17.29 & 24.03 & 67.76 & \textbf{38.73} & \textbf{19.29}  \\
     
     Seventh-dev & 16.93 & 24.63 & 67.45 & \textbf{38.50} & \textbf{19.52}  \\
     \hline
     First-std & 20.84 & 26.14 & 68.88 & \textbf{41.19} & \textbf{16.99} \\
     \hline
     
     Original-dev  & 43.40 & 36.46 & 80.87 & \textbf{58.02} & -\\
     Original-std  & 43.90 & 36.67 & 80.38 & \textbf{58.18} & - \\

     \hline
    \end{tabular}}
    \centering
    \captionsetup{justification=centering}
    \caption{LSTM Q+I model evaluation results.}
\end{subtable}
\caption{The table shows the six state-of-the-art pretrained VQA models evaluation results on the YNBQD and VQA dataset. ``-'' indicates the results are not available, ``-std'' represents the accuracy of VQA model evaluated on the complete testing set of YNBQD and VQA dataset and ``-dev'' indicates the accuracy of VQA model evaluated on the partial testing set of YNBQD and VQA dataset. In addition, $diff = Original_{dev_{All}} - X_{dev_{All}}$, where $X$ is equal to the ``First'', ``Second'', etc.}
\label{table:table7}
\end{table*}

\begin{figure*}[t]
  \begin{subfigure}[b]{0.515\textwidth}
    \includegraphics[width=0.95\linewidth]{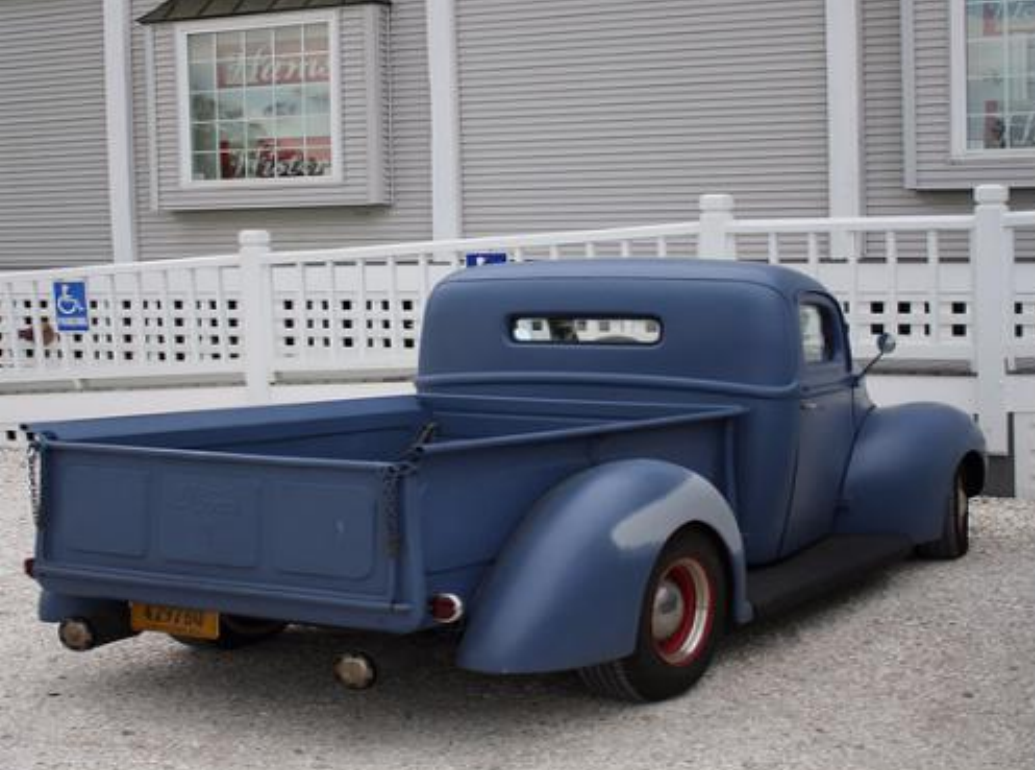}
    \caption{}
  \end{subfigure}
  \hfill
  \begin{subfigure}[b]{0.51\textwidth}
    \includegraphics[width=0.95\linewidth]{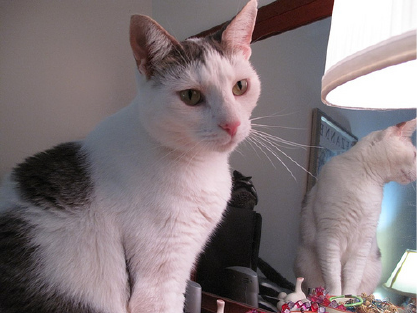}
    \caption{}
  \end{subfigure}
  \caption{Image ``(a)'' corresponds to Table \ref{table:table5}-(a). Image ``(b)'' corresponds to Table \ref{table:table5}-(b).}
\label{fig:figure10}
\end{figure*}

\vspace{3pt}\noindent\textbf{Dataset.~}

We conduct the experiments on GBQD, YNBQD and VQA dataset \cite{4}. The VQA dataset is based on the MS COCO dataset \cite{51}, and it includes $248,349$ training, $121,512$ validation and $244,302$ testing questions. Each question in VQA dataset is associated with 10 answers annotated by different people from AMT (Amazon Mechanical Turk). About $90\%$ of answers only have a single word and $98\%$ of answers have no more than three words. Regarding the GBQD and YNBQD, please refer to the \textit{Details of the Proposed Basic Question Dataset} section. To better understand them, we show some dataset examples in Table \ref{table:table5}.

\vspace{3pt}\noindent\textbf{Setup.~}

We encode all the training and validation questions of the VQA dataset into the columns of $\mathbf{A} \in \mathbf{R}^{4800 \times 186027}$, and the given main question to $\mathbf{b} \in \mathbf{R}^{4800}$ using the Skip-thought Vector \cite{43}. Regarding our $LASSO$ modeling and because the quality of BQ is mainly affected by the parameter $\lambda$, we choose $\lambda = 10^{-6}$, for the better quality, to generate our General and Yes/No BQ Datasets. Because the similarity scores are negligible after top 21 ranked BQs, we only collect the top 21 ranked General and Yes/No BQs and put them into our GBQD and YNBQD. Because most of the pretrained state-of-the-art VQA models are trained under the condition that the maximum number of input words is 26, we divide the 21 top ranked BQs, {\em i.e.}, $21=3*7$, into 7 consecutive partitions to do the robustness analysis, referring to Table \ref{table:table6} for GBQD and Table \ref{table:table7} for YNBQD. Note that, under the above setting, the total number of words for each MQ with 3 BQs is equal to or less than 26 words.

\vspace{3pt}\noindent\textbf{BQ Generation by Popular Text Evaluation Metrics.~}

In this subsection, we discuss the non-$LASSO$-based ranking methods to generate the basic questions of a given main question. We compare the performance of $LASSO$-based ranking method with non-$LASSO$-based ranking methods including seven popular sentence evaluation metrics \cite{49,67,68,69}, namely BLEU-1, BLEU-2, BLEU-3, BLEU-4, ROUGE, CIDEr and METEOR that are also used to measure the similarity score between MQ and BQs. Similar to the setup for building the General Basic Question Dataset (GBQD), we build a general basic question dataset for each metric.

\begin{table}
\centering
\scalebox{1.0}{
\begin{tabular}{|l|l|l|l|l|l|l|}
\hline
Model & \begin{tabular}[c]{@{}l@{}}LQI \end{tabular} &  \begin{tabular}[c]{@{}l@{}}HAV \end{tabular} & 
\begin{tabular}[c]{@{}l@{}}HAR \end{tabular} & 
\begin{tabular}[c]{@{}l@{}}MU \end{tabular} &
\begin{tabular}[c]{@{}l@{}}MUA \end{tabular} &
\begin{tabular}[c]{@{}l@{}}MLB \end{tabular}\\
\hline
$R_{score1}$ & 0.19 & \textbf{0.48} & 0.45 & 0.30 & 0.34 & 0.36\\ 
\hline
$R_{score2}$ & 0.08 & 0.48 & \textbf{0.53} & 0.30 & 0.23 & 0.37\\ 
\hline
\end{tabular}}
\caption{This table shows the robustness scores, $R_{score}$, of six state-of-the-art VQA models based on GBQD ($R_{score1}$), YNBQD ($R_{score2}$) and VQA \cite{4} dataset. LQI denotes LSTM Q+I, HAV denotes HieCoAtt (Alt,VGG19), HAR denotes HieCoAtt (Alt,Resnet200), MU denotes MUTAN without Attention, MUA denotes MUTAN with Attention and MLB denotes MLB with Attention. The $R_{score}$ parameters are $(t,~m) = (0.05,~20)$.}
\label{table:table1}
\vspace{-0.3cm}
\end{table}

\vspace{3pt}\noindent\textbf{Results and Analysis.~}

We describe our experimental results and robustness analysis next. 

\begin{figure*}[!tbp]
  \begin{subfigure}[b]{0.5035\textwidth}
    \includegraphics[width=0.98\linewidth]{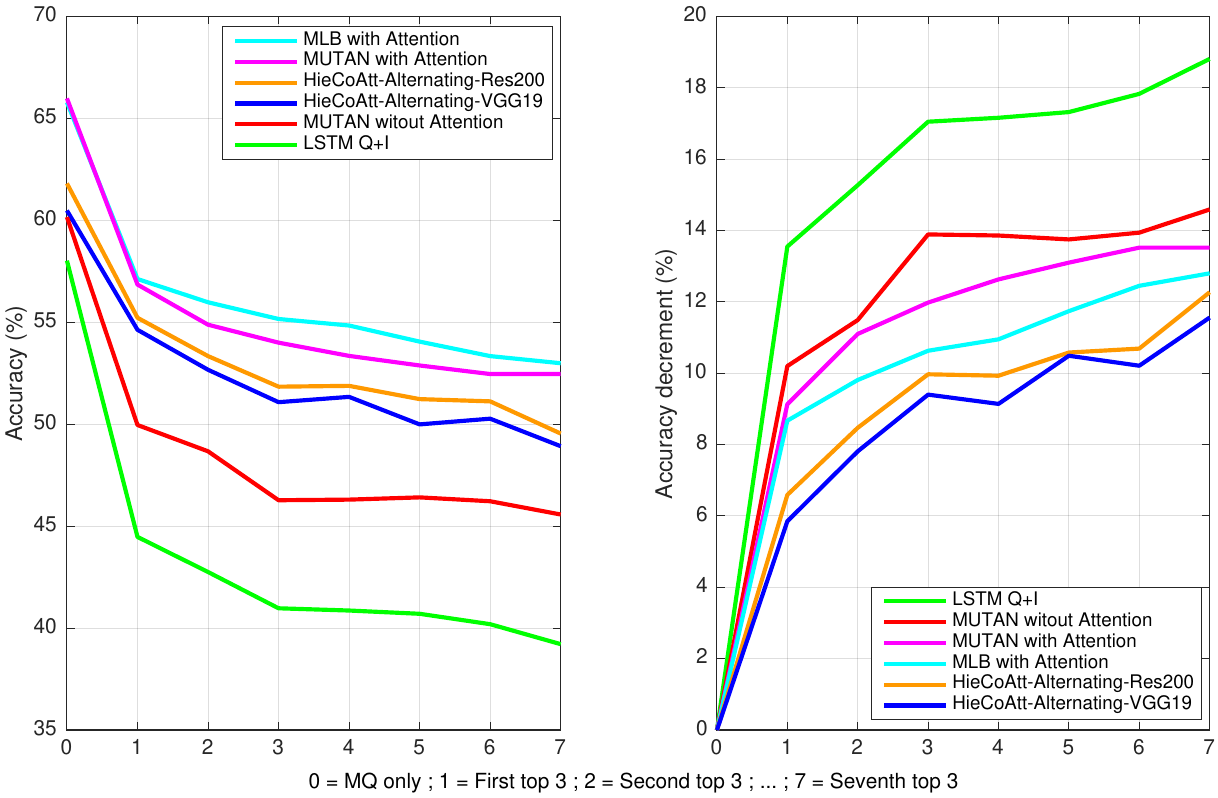}
    \caption{-1~~~~~~~~~~~~~~~~~~~~~~~~~~~~~~~~~~~(a)~-2}
  \end{subfigure}
  \hfill
  \begin{subfigure}[b]{0.5035\textwidth}
    \includegraphics[width=0.98\linewidth]{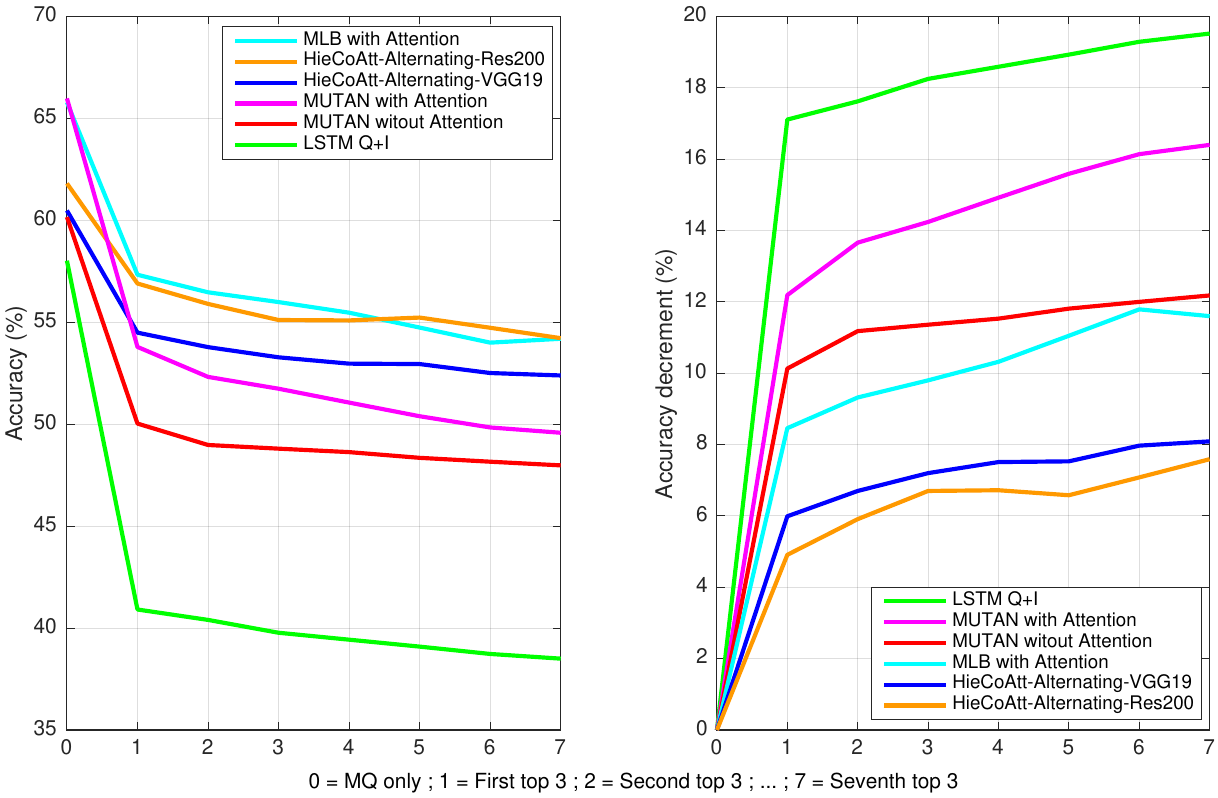}
    \caption{-1~~~~~~~~~~~~~~~~~~~~~~~~~~~~~~~~~~~(b)~-2}
  \end{subfigure}
  \caption{The figure shows the ``accuracy'' and ``accuracy decrement'' of the six state-of-the-art pretrained VQA models evaluated on GBQD, YNBQD and VQA \cite{4} datasets. These results are based on our proposed $LASSO$ BQ ranking method. Note that we divide the top 21 ranked GBQs into 7 partitions where each partition contains 3 ranked GBQs; this is in reference to (a)-1 and (a)-2. We also divide the top 21 ranked YNBQs into 7 partitions and each partition contains 3 ranked YNBQs; this is in reference to (b)-1 and (b)-2. BQs are acting as noise, so the partitions represent the noises ranked from the least noisy to the noisiest. That is, in this figure the first partition is the least noisy partition and so on. Because the plots are monotonously decreasing in  accuracy, or, equivalently, monotonously increasing in accuracy decrement, the ranking is effective. In this figure, ``First top 3'' represents the first partition, ``Second top 3'' represents the second partition and so on.}
\label{fig:figure4}
\vspace{-0.2cm}
\end{figure*}

\begin{figure*}[!tbp]
  \begin{subfigure}[b]{0.515\textwidth}
    \includegraphics[width=0.98\linewidth]{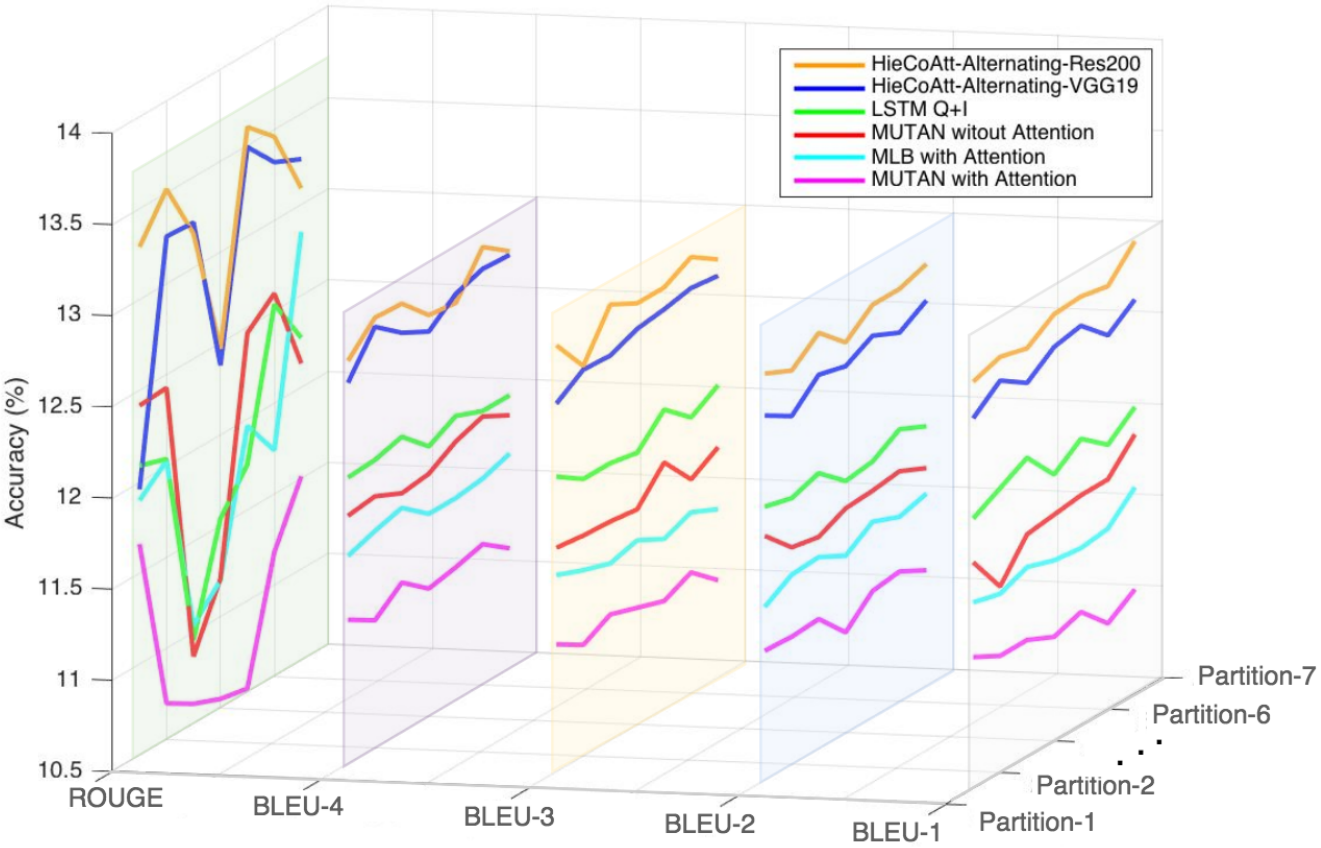}
    \caption{ROUGE, BLEU-4, BLEU-3, BLEU-2 and BLEU-1}
  \end{subfigure}
  \hfill
  \begin{subfigure}[b]{0.51\textwidth}
    \includegraphics[width=0.94\linewidth]{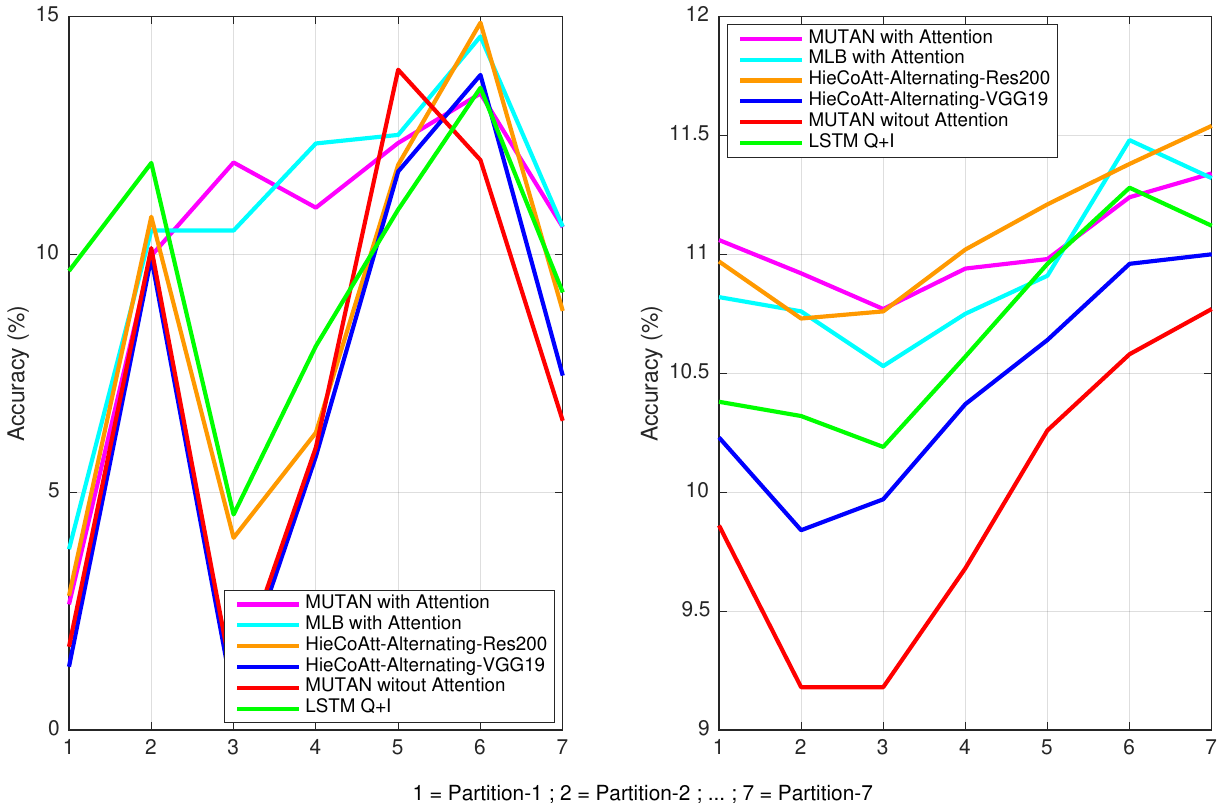}
    \caption{CIDEr~~~~~~~~~~~~~~~~~~~~~~~~~~~~~~~(c) METEOR}
  \end{subfigure}
  \caption{This figure shows the accuracy of six state-of-the-art pretrained VQA models evaluated on the GBQD and VQA dataset by different BQ ranking methods, BLEU-1, BLEU-2, BLEU-3, BLEU-4, ROUGE, CIDEr and METEOR. In (a), the grey shade denotes BLEU-1, blue shade denotes BLEU-2, orange shade denotes BLEU-3, purple shade denotes BLEU-4 and green shade denotes ROUGE. In this figure, the definition of partitions are same as Figure \ref{fig:figure4}. The original accuracy of the six VQA models can be referred to Table \ref{table:table6}-(a), Table \ref{table:table6}-(b), etc. To make the figure clear, we plot the results of CIDEr and METEOR in (b) and (c), respectively. Based on this figure and Figure \ref{fig:figure4} in our paper, our $LASSO$ ranking method performance is better than those seven ranking methods.}
\label{fig:figure5}
\end{figure*}

\vspace{3pt}\noindent\textbf{(i)} \textbf{Are the rankings of BQs effective?} We take the top 21 ranked BQs and divide them into 7 consecutive partitions and each partition contains 3 top ranked BQs. Figure \ref{fig:figure4}-(a)-1 shows that the accuracy decreases from the first partition to the seventh partition. Also, according to  Figure \ref{fig:figure4}-(a)-2, the accuracy decrement increases from the first partition to the seventh. The above two trends imply the similarity of BQs to the given MQ decreases from the first partition to the seventh partition ({\em i.e.}, the noise level increases). Specifically, the level of noise increases from the first partition to the seventh  because our assumption is that a BQ with  smaller similarity score to the given MQ indicates that this BQ introduces  more noise to the given MQ and vice versa. Note that when we replace the GBQD by YNBQD and do the same experiment (refer to Figure \ref{fig:figure4}-(b)-1 and Figure \ref{fig:figure4}-(b)-2), the trends are similar to those in GBQD. Based on Figure \ref{fig:figure4}, we conclude that the rankings by $LASSO$ ranking method are effective. However, based on Figure \ref{fig:figure5}, we discover that the accuracy of these 7 similarity metrics, $\{(BLEU_1..._4,~ROUGE,~CIDEr,~METEOR)\}$, are less monotonous and much more random from the first partition to the seventh partition. In other words, the level of noise is changing randomly from the first partition to the seventh partition. In fact, the accuracy in these results is very low compared to the original accuracy, referring to Table \ref{table:table6}. This means that the added BQs based on the 7 similarity metrics represent much more noise than the ones ranked by our $LASSO$ ranking method. Obviously, this will significantly  harm the accuracy of the state-of-the-art VQA models. According to the above, we see that the rankings by these 7 sentence similarity metrics are not effective in this context. 

\vspace{3pt}\noindent\textbf{(ii)} \textbf{Which VQA model is the most robust?} We divide current VQA models into two categories, attention-based and non-attention-based. Referring to Table \ref{table:table1}, HAV, HAR, MUA and MLB are attention-based models whereas LQI and MU are not. Generally speaking and according to  Table \ref{table:table1}, the attention-based VQA models are more robust than non-attention-based ones. However, when we consider  MU and MUA in  Table \ref{table:table1} ($R_{score2}$), the non-attention-based model (MU) is more robust than the attention-based model (MUA). Note that the difference between MU and MUA is only the attention mechanism. Yet, in Table \ref{table:table1} ($R_{score1}$), MUA is more robust than MU. It implies that the variety of BQ candidates affects the robustness of attention-based VQA models in some cases. Finally, based on Table \ref{table:table1}, we conclude that HieCoAtt \cite{41} is the most robust VQA model. Since the HieCoAtt model with co-attention mechanism which repeatedly exploits the text and image information to guide each other, it makes VQA models more robust \cite{41,huang2019novel}. Based on our experimental result, we know that HieCoAtt is the most robust VQA model, and this motivates us to conduct the extended experiments for this model.

\begin{figure}
\begin{center}
   \includegraphics[width=0.98\linewidth]{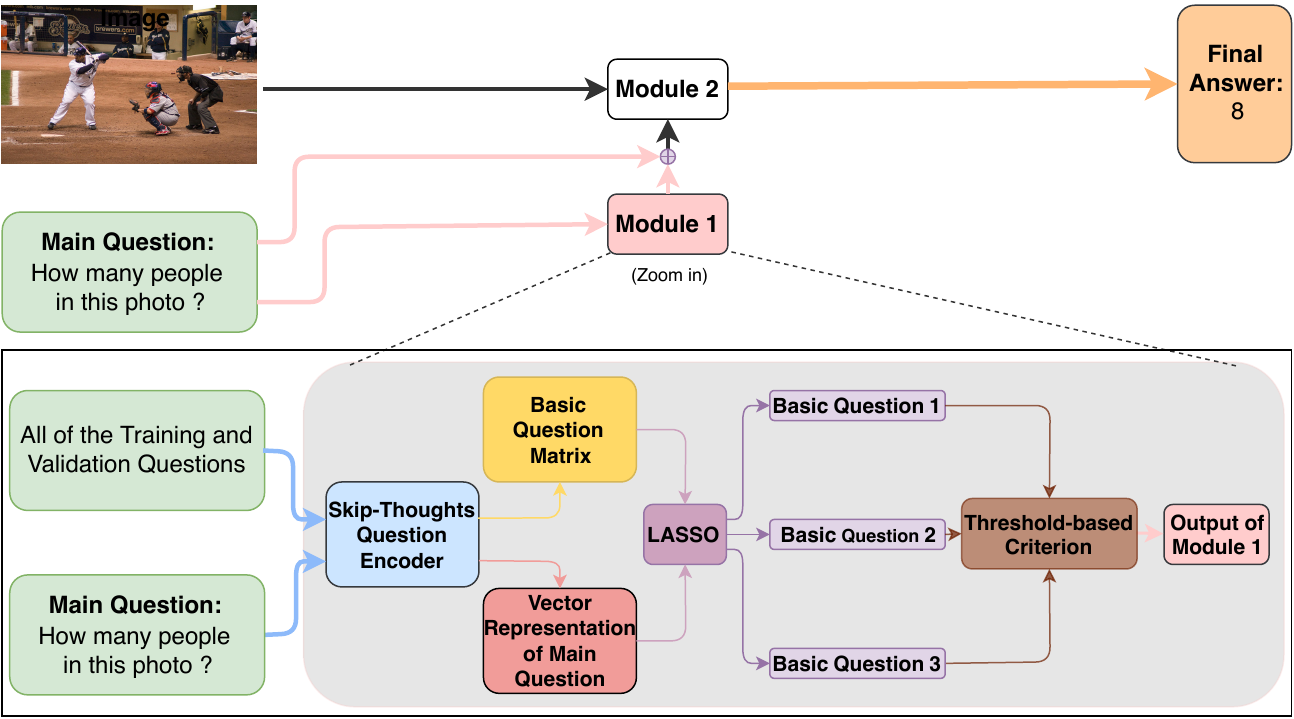}
   \caption{Visual Question Answering by Basic Questions (VQABQ) pipeline. Note that in Module 1 all of the training and validation questions are only encoded by Skip-Thought Question Encoder once for generating the Basic Question Matrix. That is, the next input of Skip-Thought Question Encoder is only a new main question. Module 2 is a VQA model which we want to test, and it is the HieCoAtt VQA model in our case. Regarding the input question of the HieCoAtt model, it is the direct concatenation of a given main question with the corresponding selected basic questions based on the Threshold-based Criterion. ``$\oplus$'' denotes the direct concatenation of basic questions.}
\label{fig:figure8}
\end{center}
\end{figure}

\begin{table}[t]
\centering
\scalebox{1.0}{
\begin{tabular}{|l|l|l|l|}
\hline
\multicolumn{1}{|c|}{} & score1 & score2/score1 & score3/score2 \\ \hline
avg & ~0.33  & ~~~~~~0.61 & ~~~~~~0.73 \\ 
\hline
std & ~0.20  & ~~~~~~0.27 & ~~~~~~0.21 \\
\hline
\end{tabular}}
\caption{In this table, ``avg'' denotes average and ``std'' denotes standard deviation.}
\label{table:table3}
\end{table}

\begin{table}[t]
\centering
\scalebox{1.0}{
\begin{tabular}{|l|l|l|l|l|}
\hline
\multicolumn{5}{|c|}{Opend-Ended Case (Total: 244302 questions)}\\ 
\hline
& \begin{tabular}[c]{@{}l@{}}~~~0 BQ \\
(96.84\%)\end{tabular} & \begin{tabular}[c]{@{}l@{}}~~1 BQ \\ (3.07\%)\end{tabular} & \begin{tabular}[c]{@{}l@{}}~~2 BQ \\ (0.09\%)\end{tabular} & \begin{tabular}[c]{@{}l@{}}~~3 BQ \\ (0.00\%)\end{tabular} \\
\hline
\# Q & ~236570  & ~~~7512  & ~~~211  & ~~~~~9 \\ 
\hline
\end{tabular}}
\caption{The table shows how many BQs are appended. ``$X$ BQ'' means $X$ BQs are appended by MQ, where $X = 0, 1, 2, 3$, and ``\# Q'' denote number of questions.}
\label{table:table4}
\vspace{-0.10cm}
\end{table}

\vspace{3pt}\noindent\textbf{(iii)} \textbf{Can basic questions directly help the accuracy of the HieCoAtt model?} According to Table \ref{table:table1}, we know that HieCoAtt is the most robust VQA model. Also, it was the previous state-of-the-art VQA model in the sense of accuracy \cite{41}. The above reasons motivate us to conduct the extended experiment and analysis of this model. We claim that if the quality of BQs is good enough, then using direct concatenation of MQ and BQs helps the accuracy of the HieCoAtt VQA model. To justify the claim, we propose a framework, Visual Question Answering by Basic Questions (VQABQ), to exploit selected BQs to analyze the HieCoAtt VQA model, referring to Figure \ref{fig:figure8}. We select BQs with a good quality based on a threshold-based criterion, referring to Algorithm \ref{algorithm1}. In our proposed BQD, each MQ has 21 corresponding BQs with scores and these scores are all between $[0-1]$ with the following order: 
\begin{equation}
    ~~~~~~~~~~~~~~~score1\geq score2\geq...\geq score21,
\end{equation}
where we further define three thresholds, $s1$, $s2$ and $s3$, for the selection process. For convenience, we only take the top 3 ranked BQs to do the selection. Then, we compute the averages ($avg$) and standard deviations ($std$) for $score1$, $score2/score1$, and $score3/score2$, respectively (refer to Table \ref{table:table3}). We use $avg \pm std$ to be the initial estimation of the above three thresholds. We discover that when $s1 = 0.60$, $s2 = 0.58$, and $s3 = 0.41$, we will get the BQs which best help the accuracy of the HieCoAtt VQA model in case of the MQ-BQs direct concatenation method.


According to Table \ref{table:table4}, about $96.84\%$ testing questions (MQs) cannot find the proper BQs to improve the accuracy of the HieCoAtt model by the MQ-BQs direct concatenation method. Although we only have around $3.16\%$ MQs benefit from the BQs, our method still makes the performance of the HieCoAtt model competitive, accuracy increasing from $60.32\%$ to $60.34\%$, referring to Table \ref{table:table100}. In other words, the number of questions answered correctly by our proposed method is around 49 questions more than the original HieCoAtt VQA model \cite{41}. It implies that if we have a good enough basic question dataset, then it helps us increase more accuracy. Accordingly, based on our experimental results, we believe that BQs with good enough quality help the accuracy of the HieCoAtt VQA model by using the direct concatenation method.


\begin{table}
\centering
\begin{tabular}{|l|l|l}
\multicolumn{3}{l}{~~~~~~~~~~~~~~~~~~~~~~~HieCoAtt (Alt,VGG19)} \\ 
\hline
\hline
\multicolumn{2}{l|}{~~~~~(s1, s2, s3)} & (test-dev-acc, Other, Num, Y/N) \\ \hline
\multicolumn{2}{l|}{(0.60, 0.58, 0.41)} & ~~~(60.49, 49.12, 38.43, 79.65) \\ \hline
\hline
\multicolumn{2}{l|}{~~~~~(s1, s2, s3)} & (test-std-acc, Other, Num, Y/N) \\ \hline
\multicolumn{2}{l|}{(0.60, 0.58, 0.41)} & ~~~(60.34, 49.16, 36.50, 79.49) \\ \hline
\hline
\end{tabular}
\vspace{-0.1cm}
\caption{Evaluation results of HieCoAtt (Alt,VGG19) model improved by Algorithm \ref{algorithm1}. Note that the original accuracy of HieCoAtt (Alt,VGG19) VQA model for ``test-dev-acc'' is $60.48$ and for ``test-std-acc'' is $60.32$.}
\label{table:table100}
\end{table}

\begin{algorithm}[t]
\caption{~~MQ-BQs Concatenation Algorithm}\label{algorithm1}
\begin{algorithmic}[1]
\State \textbf{Note that s1, s2, s3 are thresholds we can choose.}
\Procedure{MQ-BQs concatenation}{}

\If {$score1 > s1$} 
\State appending the given MQ and BQ1 with the largest score
\EndIf

\If {$score2/score1 > s2$} 
\State appending the given MQ, BQ1, and BQ2 with the second large score
\EndIf

\If {$score3/score2 > s3$} 
\State appending the given MQ, BQ1, BQ2, and BQ3 with the third large score
\EndIf

\EndProcedure
\end{algorithmic}
\end{algorithm}

\begin{figure}
\begin{center}
   \includegraphics[width=0.98\linewidth]{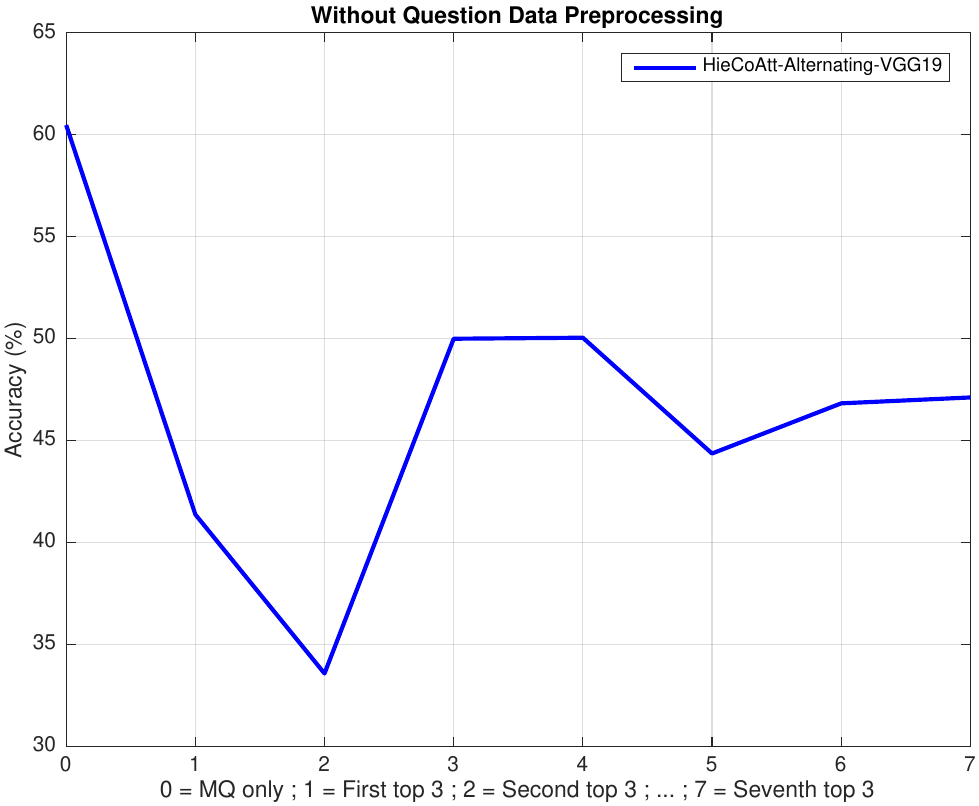}
\caption{This figure demonstrates what is the ranking result of jumping randomly. For convenience, we only take the most robust VQA model, HieCoAtt, to demonstrate the random jump. If we do not have question sentences preprocessing, then the proposed $LASSO$ ranking method is ineffective. That is, if we have done the question sentences preprocessing, the trend in this figure should be similar to Figure \ref{fig:figure4}-(a)-1 and Figure \ref{fig:figure4}-(b)-1. In this figure, ``MQ only'' represents the original query question and ``First top 3'' represents the first partition, ``Second top 3'' represents the second partition and so on. For the detailed numbers, please refer to Table \ref{table:table22}.}
\label{fig:figure9}
\end{center}
\end{figure}

\noindent
\vspace{3pt}\noindent\textbf{(iv)} \textbf{Is question sentences preprocessing necessary?} We claim that question sentences preprocessing is necessary for our proposed $LASSO$ ranking method. For convenience, we exploit the same HieCoAtt model to show the claim. 
In the previous \textit{Methodology} section, we do the question sentences preprocessing before the sentences embedding. If we do not have the step of question sentences preprocessing, the $LASSO$ ranking method will generate some random ranking result. For convenience, we take the same HieCoAtt VQA model to demonstrate what the random ranking is. As shown in Figure \ref{fig:figure9}, the ranking result is jumping randomly because of not doing the question sentences preprocessing. If the proposed method works correctly, the trend of Figure \ref{fig:figure9} should be monotone like trends in Figure \ref{fig:figure4}. 

\vspace{3pt}\noindent\textbf{(v)} \textbf{What are the pros and cons of each metric?} To compare with our proposed $LASSO$ basic question ranking method, we also conduct the basic question ranking experiments using the seven aforementioned text similarity metrics
on the same basic question candidate dataset. Although the ranking performance of these metrics is less than satisfactory, various works \cite{1,19,27,28,48} still use them for sentence evaluation because of their simple implementation.  As for our $LASSO$ ranking method, the ranking performance is quite effective, despite its simplicity. Note that, in practice, we will directly use our proposed datasets to test the robustness of VQA models without running the $LASSO$ ranking method again, so the computational complexity of $LASSO$ ranking method is not an issue in this case.


\begin{table}[t]
    \small
    \centering
    \scalebox{0.94}{
    \begin{tabular}{ c | c c c c | c} 
     Task Type &    & \multicolumn{3}{c}{Open-Ended} &  \\ [0.5ex]
     \hline
     Method &    & \multicolumn{3}{c}{HieCoAtt (Alt,VGG19)} &  \\ [0.5ex]
     \hline
     Test Set&  \multicolumn{4}{c}{dev} & diff \\ [0.5ex]
     \hline
     Partition & Other & Num & Y/N & All & All \\ [0.5ex] 
     \hline
     First-dev & 33.83 & 37.19 & 51.34 & \textbf{41.38} & \textbf{20.43}  \\ 
     
     Second-dev & 15.46 & 31.42 & 55.38 & \textbf{33.58} & \textbf{28.23} \\
     
     Third-dev & 35.33 & 36.53 & 70.76 & \textbf{50.00} & \textbf{11.81}  \\
     
     Fourth-dev & 36.05 & 36.46 & 70.05 & \textbf{50.05} & \textbf{11.76} \\
     
     Fifth-dev & 29.89 & 30.02 & 65.14 & \textbf{44.37} & \textbf{17.44} \\
     
     Sixth-dev & 35.81 & 34.48 & 63.02 & \textbf{46.83} & \textbf{14.98}  \\
     
     Seventh-dev & 39.12 & 34.45 & 59.84 & \textbf{47.12} & \textbf{14.69}  \\
     \hline
     
     Original-dev  & 51.77 & 38.65 & 79.70 & \textbf{61.81} & -\\
     Original-std  & 51.95 & 38.22 & 79.95 & \textbf{62.06} & - \\ 
     
     \hline
    \end{tabular}}
    \caption{The HieCoAtt (Alt,VGG19) model evaluation results on BQD and VQA dataset \cite{4} without question sentences preprocessing. ``-'' indicates the results are not available, ``-std'' means that the VQA model is evaluated by the complete testing set of BQD and VQA dataset, and ``-dev'' means that the VQA model is evaluated by the partial testing set of BQD and VQA dataset. In addition, $diff = Original_{dev_{All}} - X_{dev_{All}}$, where $X$ is equal to ``First'', ``Second'',..., ``Seventh''.}
\label{table:table22}
\end{table}

\begin{figure*}[!tbp]
  \begin{subfigure}[b]{0.50\textwidth}
    \includegraphics[width=0.86\linewidth]{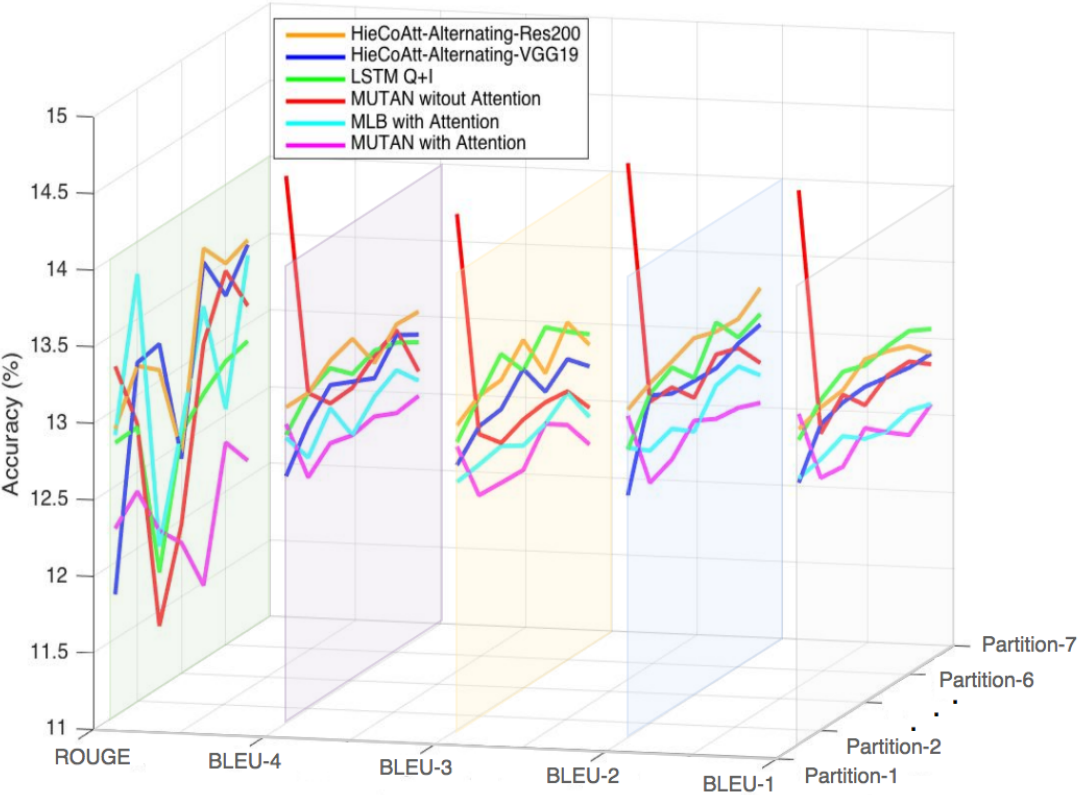}
    \caption{ROUGE, BLEU-4, BLEU-3, BLEU-2 and BLEU-1}
  \end{subfigure}
  \hfill
  \begin{subfigure}[b]{0.52\textwidth}
    \includegraphics[width=0.95\linewidth]{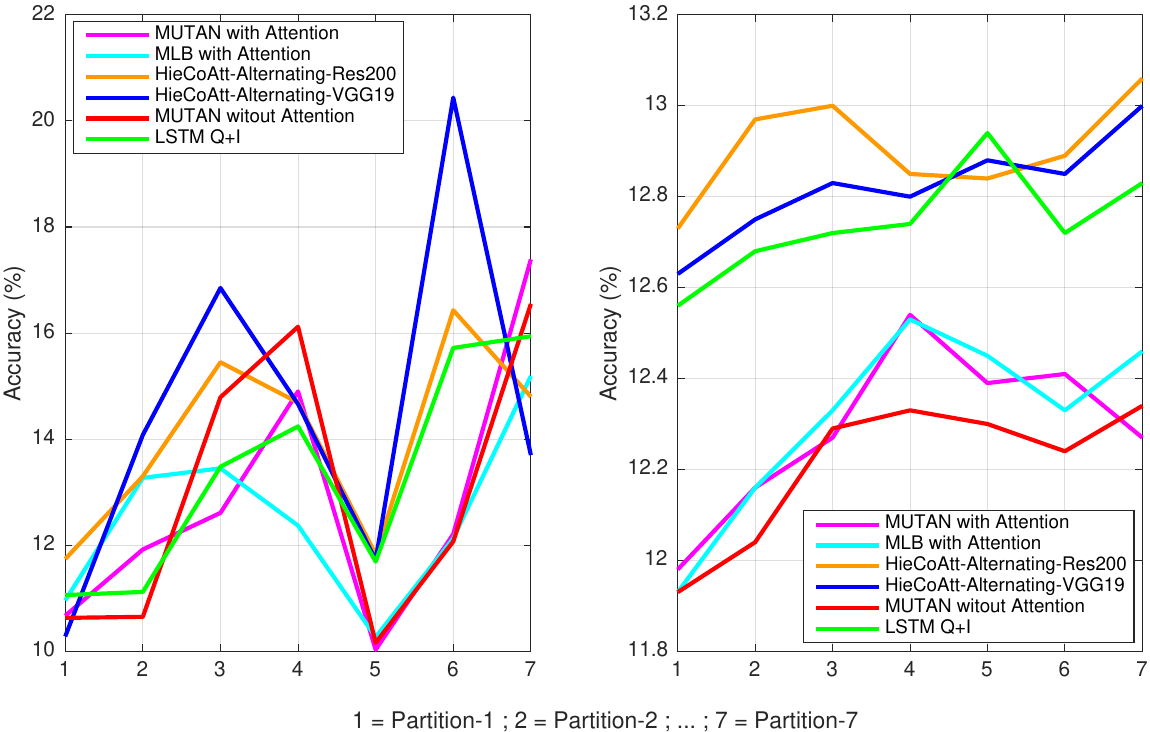}
    \caption{CIDEr~~~~~~~~~~~~~~~~~~~~~~~~~~~~~~~(c) METEOR}
  \end{subfigure}
  \caption{This figure shows the accuracy of six state-of-the-art pretrained VQA models evaluated on the YNBQD and VQA \cite{4} dataset by different BQ ranking methods, BLEU-1, BLEU-2, BLEU-3, BLEU-4, ROUGE, CIDEr and METEOR. The result in this figure is consistent with the result in Figure \ref{fig:figure5}. Note that in Figure \ref{fig:figure7}-(a), the grey shade denotes BLEU-1, blue shade denotes BLEU-2, orange shade denotes BLEU-3, purple shade denotes BLEU-4 and green shade denotes ROUGE. For more detailed explanation, please refer to the \textit{Extended experiments on YNBQD dataset} subsection.}
\label{fig:figure7}
\end{figure*}

\vspace{3pt}\noindent\textbf{(vi)} \textbf{Is the ranking in semantic meaning effective?} In the $LASSO$ BQ ranking method, the semantic meaning of a question cannot be ranked very accurately but it still works quite well. This is primarily due to the state-of-the-art question encoder, Skip-thoughts \cite{43}. It cannot completely capture the semantic meaning of the question and embed it into vector format. We believe that if more semantic encoders are developed in the future, the $LASSO$ ranking method can readily make use of them to produce more semantically driven ranking. Although the semantic meaning ranking by $LASSO$ ranking method is not very accurate, it is still acceptable. We provide some BQ ranking results using our $LASSO$ ranking method in Figure \ref{fig:figure10} and Table \ref{table:table5}.


\vspace{3pt}\noindent\textbf{(vii)} \textbf{What affects the quality of BQs?} In our model, $\lambda$ is one of the most important factors that can affect the quality of BQs. Through our experiments, we find that $\lambda \in [10^{-6}, 10^{-5}]$ yields satisfactory ranking performance. One can refer to Figure \ref{fig:figure4} to get an understanding of the satisfactory ranking performance. We provide some ranking examples based on $LASSO$ ranking method in Table \ref{table:table5} to show the quality of BQs when $\lambda = 10^{-6}$. 


\vspace{3pt}\noindent\textbf{(viii)} \textbf{Extended experiments on YNBQD dataset.} Although we have done the basic question ranking experiments by the seven different text similarity metrics, $BLEU_1..._4,~ROUGE,~CIDEr,$ and $METEOR$, on GBQD, we haven't done such ranking experiments by those metrics on YNBQD. So, we explain the experimental details in the following. We conduct the extended experiments on our proposed YNBQD dataset by the above seven different text similarity metrics. 
In Figure \ref{fig:figure7}, the definition of partitions are the same as Figure \ref{fig:figure4}. The original accuracy of the six VQA models is given in Table \ref{table:table7}-(a), Table \ref{table:table7}-(b), etc. For convenience, we plot the results of CIDEr and METEOR in Figure \ref{fig:figure7}-(b) and Figure \ref{fig:figure7}-(c), respectively. Based on Figure \ref{fig:figure7}, Figure \ref{fig:figure5}, and Figure \ref{fig:figure4}, we conclude that the proposed $LASSO$ ranking method performance is better than those seven ranking methods on both YNBQD and GBQD datasets. For the detail numbers of all the experiment, please refer to Table \ref{table:table6}, \ref{table:table7},..., \ref{table:table21}.

\vspace{-0.5cm}
\section{Discussion}
In this section, we present state-of-the-art VQA models among our six tested VQA models, \cite{4,41,57,59}, in different senses.


\noindent
\vspace{+0.1cm}\textbf{In the sense of robustness.} 

According to Table \ref{table:table1}, we observe that ``HieCoAtt (Alt,VGG19)'' model has the highest $R_{score1}$, 0.48. Furthermore, ``HieCoAtt (Alt,Resnet200)'' has the highest $R_{score2}$, 0.53. Therefore, for GBQD, ``HieCoAtt (Alt,VGG19)'' model is the state-of-the-art VQA model among our six tested VQA models in the sense of robustness. However, for YNBQD, ``HieCoAtt (Alt,Resnet200)'' model is the state-of-the-art VQA model among our six tested VQA models in the sense of robustness. Additionally, ``LSTM Q+I'' model has the lowest $R_{score1}$ and $R_{score2}$. Generally speaking, we can say that the attention-based VQA model is more robust than the non-attention-based one. 

\noindent
\vspace{+0.1cm}\textbf{In the sense of accuracy.} 

According to Table \ref{table:table6}, we discover that ``MUTAN with Attention'' model in Table \ref{table:table6}-(d) has the highest accuracy, 65.77, and ``LSTM Q+I'' has the lowest accuracy, 58.18. Therefore, ``MUTAN with Attention'' model is the state-of-the-art VQA model among our six tested VQA models in the sense of accuracy. Also, these results imply that the attention-based VQA model has higher accuracy than the non-attention-based one.

\vspace{-0.5cm}
\section{Conclusion}
In this work, we propose a novel method comprised of a number of components namely, large-scale General Basic Question Dataset, Yes/No Basic Question Dataset and robustness measure ($R_{score}$) for measuring the robustness of VQA models. 

Our method contains two main modules, VQA module and Noise Generator. The former one is able to rank the given BQs and the latter one is able to take the query, basic questions and an image as input and then output the natural language answer of the given query question about the image. The goal of the proposed method is to serve as a benchmark to help the community in building more accurate \emph{and} robust VQA models. 

Moreover, based on our proposed General and Yes/No Basic Question Datasets and $R_{score}$, we show that our $LASSO$ BQ ranking method has the better ranking performance among most of the popular text evaluation metrics. Finally, we have some new methods to evaluate the robustness of VQA models, so how to build a robust \emph{and} accurate VQA model will be interesting future work.



\appendices
\section{Appendices}
Detailed experimental results are presented in Table \ref{table:table8}, \ref{table:table9},..., \ref{table:table21}. 


\begin{table*}
\renewcommand\arraystretch{1.28}
\setlength\tabcolsep{11pt}
    \centering
\begin{subtable}[t]{0.3\linewidth}
\centering
\scalebox{0.53}{
    \begin{tabular}{ c | c c c c | c} 
     Task Type &    & \multicolumn{3}{c}{Open-Ended (BLEU-1)} &  \\ [0.5ex]
     \hline
     Method &    & \multicolumn{3}{c}{MUTAN without Attention} &  \\ [0.5ex]
     \hline
     Test Set&  \multicolumn{4}{c}{dev} & diff \\ [0.5ex]
     \hline
     Partition & Other & Num & Y/N & All & All \\ [0.5ex] 
     \hline
     First-dev & 2.73 & 2.92 & 24.63 & 11.74 & 48.42  \\ 
     
     Second-dev & 2.57 & 2.96 & 24.28 & 11.52 & 48.64 \\
     
     Third-dev & 2.79 & 2.89 & 24.53 & 11.72 & 48.44  \\
     
     Fourth-dev & 2.68 & 2.94 & 24.67 & 11.74 & 48.42  \\
     
     Fifth-dev & 2.69 & 2.87 & 24.73 & 11.76 & 48.40 \\
     
     Sixth-dev & 2.80 & 2.79 & 24.62 & 11.76 & 48.40  \\
     
     Seventh-dev & 2.77 & 2.99 & 25.01 & 11.92 & 48.24  \\
     \hline
     First-std & 2.52 & 2.70 & 24.66 & 11.66 & 48.79 \\
     \hline
     
     Original-dev  & 47.16 & 37.32 & 81.45 & 60.16 & -\\
     Original-std  & 47.57 & 36.75 & 81.56 & 60.45 & - \\

     \hline
    \end{tabular}}
    \centering
    \captionsetup{justification=centering}
    \caption{MUTAN without Attention model evaluation results.}

\scalebox{0.53}{
    \begin{tabular}{ c | c c c c | c} 
     Task Type &    & \multicolumn{3}{c}{Open-Ended (BLEU-1)} &  \\ [0.5ex]
     \hline
     Method &    & \multicolumn{3}{c}{HieCoAtt (Alt,VGG19)} &  \\ [0.5ex]
     \hline
     Test Set&  \multicolumn{4}{c}{dev} & diff \\ [0.5ex]
     \hline
     Partition & Other & Num & Y/N & All & All \\ [0.5ex] 
     \hline
     First-dev & 2.94 & 2.93 & 26.31 & 12.53 & 47.95  \\ 
     
     Second-dev & 2.95 & 3.05 & 26.56 & 12.65 & 47.83 \\
     
     Third-dev & 3.08 & 2.95 & 26.19 & 12.55 & 47.93  \\
     
     Fourth-dev & 3.17 & 2.90 & 26.36 & 12.66 & 47.82  \\
     
     Fifth-dev & 3.17 & 3.05 & 26.39 & 12.69 & 47.79 \\
     
     Sixth-dev & 3.21 & 3.11 & 25.99 & 12.55 & 47.93  \\
     
     Seventh-dev & 3.12 & 3.14 & 26.37 & 12.66 & 47.82  \\
     \hline
     First-std & 2.76 & 2.73 & 26.10 & 12.38 & 47.94 \\
     \hline

     Original-dev  & 49.14 & 38.35 & 79.63 & 60.48 & -\\
     Original-std  & 49.15 & 36.52 & 79.45 & 60.32 & - \\

     \hline
    \end{tabular}}
    \centering
    \captionsetup{justification=centering}
    \caption{HieCoAtt (Alt,VGG19) model evaluation results.}
\end{subtable}%
    \hfil
\begin{subtable}[t]{0.3\linewidth}

\centering
\scalebox{0.53}{
    \begin{tabular}{ c | c c c c | c} 
     Task Type &    & \multicolumn{3}{c}{Open-Ended (BLEU-1)} &  \\ [0.5ex]
     \hline
     Method &    & \multicolumn{3}{c}{MLB with Attention} &  \\ [0.5ex]
     \hline
     Test Set&  \multicolumn{4}{c}{dev} & diff \\ [0.5ex]
     \hline
     Partition & Other & Num & Y/N & All & All \\ [0.5ex] 
     \hline
     First-dev & 2.65 & 2.17 & 24.38 & 11.52 & 54.27  \\ 
     
     Second-dev & 2.75 & 2.21 & 24.17 & 11.48 & 54.31 \\
     
     Third-dev & 2.89 & 2.03 & 24.19 & 11.54 & 54.25  \\
     
     Fourth-dev & 2.90 & 2.26 & 24.01 & 11.49 & 54.30  \\
     
     Fifth-dev & 2.81 & 2.17 & 24.09 & 11.47 & 54.32 \\
     
     Sixth-dev & 2.80 & 2.16 & 24.15 & 11.49 & 54.30  \\
     
     Seventh-dev & 2.91 & 2.29 & 24.32 & 11.63 & 54.16  \\
     \hline
     First-std & 2.68 & 2.03 & 23.88 & 11.35 & 54.33 \\
     \hline
     
     Original-dev  & 57.01 & 37.51 & 83.54 & 65.79 & -\\
     Original-std  & 56.60 & 36.63 & 83.68 & 65.68 & - \\

     \hline
    \end{tabular}}
    \centering
    \captionsetup{justification=centering}
    \caption{MLB with Attention model evaluation results.}
    
\scalebox{0.53}{
    \begin{tabular}{ c | c c c c | c} 
     Task Type &    & \multicolumn{3}{c}{Open-Ended (BLEU-1)} &  \\ [0.5ex]
     \hline
     Method &    & \multicolumn{3}{c}{MUTAN with Attention} &  \\ [0.5ex]
     \hline
     Test Set&  \multicolumn{4}{c}{dev} & diff \\ [0.5ex]
     \hline
     Partition & Other & Num & Y/N & All & All \\ [0.5ex] 
     \hline
     First-dev & 2.06 & 2.05 & 24.37 & 11.22 & 54.76  \\ 
     
     Second-dev & 2.13 & 2.28 & 24.05 & 11.14 & 54.84 \\
     
     Third-dev & 2.13 & 2.07 & 24.10 & 11.14 & 54.84  \\
     
     Fourth-dev & 2.11 & 2.26 & 23.09 & 11.07 & 54.91  \\
     
     Fifth-dev & 2.15 & 2.35 & 23.97 & 11.12 & 54.86 \\
     
     Sixth-dev & 2.06 & 2.24 & 23.73 & 10.97 & 55.01  \\
     
     Seventh-dev & 2.06 & 2.17 & 23.99 & 11.07 & 54.91  \\
     \hline
     First-std & 2.04 & 2.17 & 24.00 & 11.10 & 54.67 \\
     \hline
     
     Original-dev  & 56.73 & 38.35 & 84.11 & 65.98 & -\\
     Original-std  & 56.29 & 37.47 & 84.04 & 65.77 & - \\

     \hline
    \end{tabular}}
    \centering
    \captionsetup{justification=centering}
    \caption{MUTAN with Attention model evaluation results.}
\end{subtable}%
    \hfil
\begin{subtable}[t]{0.3\linewidth}
        
\centering
\scalebox{0.53}{
    \begin{tabular}{ c | c c c c | c} 
     Task Type &    & \multicolumn{3}{c}{Open-Ended (BLEU-1)} &  \\ [0.5ex]
     \hline
     Method &    & \multicolumn{3}{c}{HieCoAtt (Alt,Resnet200)} &  \\ [0.5ex]
     \hline
     Test Set&  \multicolumn{4}{c}{dev} & diff \\ [0.5ex]
     \hline
     Partition & Other & Num & Y/N & All & All \\ [0.5ex] 
     \hline
     First-dev & 3.37 & 3.09 & 26.26 & 12.73 & 49.08  \\ 
     
     Second-dev & 3.39 & 3.12 & 26.33 & 12.78 & 49.03 \\
     
     Third-dev & 3.51 & 2.98 & 26.15 & 12.74 & 49.07  \\
     
     Fourth-dev & 3.48 & 3.09 & 26.40 & 12.84 & 48.97  \\
     
     Fifth-dev & 3.56 & 2.85 & 26.37 & 12.85 & 48.96 \\
     
     Sixth-dev & 3.52 & 3.02 & 26.30 & 12.82 & 48.99  \\
     
     Seventh-dev & 3.60 & 3.22 & 26.57 & 12.98 & 48.83  \\
     \hline
     First-std & 3.22 & 2.77 & 25.95 & 12.54 & 49.52\\
     \hline

     Original-dev  & 51.77 & 38.65 & 79.70 & 61.81 & -\\
     Original-std  & 51.95 & 38.22 & 79.95 & 62.06 & - \\

     \hline
    \end{tabular}}
    \centering
    \captionsetup{justification=centering}
    \caption{HieCoAtt (Alt,Resnet200) model evaluation results.}

\scalebox{0.53}{
    \begin{tabular}{ c | c c c c | c} 
     Task Type &    & \multicolumn{3}{c}{Open-Ended (BLEU-1)} &  \\ [0.5ex]
     \hline
     Method &    & \multicolumn{3}{c}{LSTM Q+I} &  \\ [0.5ex]
     \hline
     Test Set&  \multicolumn{4}{c}{dev} & diff \\ [0.5ex]
     \hline
     Partition & Other & Num & Y/N & All & All \\ [0.5ex] 
     \hline
     First-dev & 2.08 & 3.09 & 25.95 & 11.98 & 46.04  \\ 
     
     Second-dev & 1.98 & 3.35 & 26.18 & 12.06 & 45.96 \\
     
     Third-dev & 2.04 & 3.25 & 26.32 & 12.14 & 45.88  \\
     
     Fourth-dev & 2.01 & 3.26 & 25.94 & 11.96 & 46.06  \\
     
     Fifth-dev & 2.03 & 3.31 & 26.15 & 12.07 & 45.95 \\
     
     Sixth-dev & 2.16 & 3.41 & 25.68 & 11.95 & 46.07  \\
     
     Seventh-dev & 2.10 & 3.31 & 26.08 & 12.07 & 45.95  \\
     \hline
     First-std & 2.03 & 3.31 & 25.86 & 11.98 & 46.20 \\
     \hline
     
     Original-dev  & 43.40 & 36.46 & 80.87 & 58.02 & -\\
     Original-std  & 43.90 & 36.67 & 80.38 & 58.18 & - \\

     \hline
    \end{tabular}}
    \centering
    \captionsetup{justification=centering}
    \caption{LSTM Q+I model evaluation results.}
\end{subtable}
\caption{The table shows the six state-of-the-art pretrained VQA models evaluation results on the GBQD and VQA dataset. ``-'' indicates the results are not available, ``-std'' represents the accuracy of VQA model evaluated on the complete testing set of GBQD and VQA dataset and ``-dev'' indicates the accuracy of VQA model evaluated on the partial testing set of GBQD and VQA dataset. In addition, $diff = Original_{dev_{All}} - X_{dev_{All}}$, where $X$ is equal to the ``First'', ``Second'', etc.}
\label{table:table8}
\end{table*}

\begin{table*}
\renewcommand\arraystretch{1.28}
\setlength\tabcolsep{11pt}
    \centering
\begin{subtable}[t]{0.3\linewidth}
\centering
\scalebox{0.53}{
    \begin{tabular}{ c | c c c c | c} 
     Task Type &    & \multicolumn{3}{c}{Open-Ended (BLEU-2)} &  \\ [0.5ex]
     \hline
     Method &    & \multicolumn{3}{c}{MUTAN without Attention} &  \\ [0.5ex]
     \hline
     Test Set&  \multicolumn{4}{c}{dev} & diff \\ [0.5ex]
     \hline
     Partition & Other & Num & Y/N & All & All \\ [0.5ex] 
     \hline
     First-dev & 2.55 & 2.93 & 25.09 & 11.84 & 48.32  \\ 
     
     Second-dev & 2.57 & 2.94 & 24.69 & 11.69 & 48.47 \\
     
     Third-dev & 2.66 & 2.84 & 24.54 & 11.66 & 48.50  \\
     
     Fourth-dev & 2.70 & 2.91 & 24.65 & 11.73 & 48.43  \\
     
     Fifth-dev & 2.68 & 2.80 & 24.73 & 11.74 & 48.42 \\
     
     Sixth-dev & 2.64 & 3.09 & 24.74 & 11.76 & 48.40  \\
     
     Seventh-dev & 2.59 & 2.95 & 24.66 & 11.69 & 48.47  \\
     \hline
     First-std & 2.33 & 2.63 & 24.71 & 11.59 & 48.86 \\
     \hline
     
     Original-dev  & 47.16 & 37.32 & 81.45 & 60.16 & -\\
     Original-std  & 47.57 & 36.75 & 81.56 & 60.45 & - \\

     \hline
    \end{tabular}}
    \centering
    \captionsetup{justification=centering}
    \caption{MUTAN without Attention model evaluation results.}

\scalebox{0.53}{
    \begin{tabular}{ c | c c c c | c} 
     Task Type &    & \multicolumn{3}{c}{Open-Ended (BLEU-2)} &  \\ [0.5ex]
     \hline
     Method &    & \multicolumn{3}{c}{HieCoAtt (Alt,VGG19)} &  \\ [0.5ex]
     \hline
     Test Set&  \multicolumn{4}{c}{dev} & diff \\ [0.5ex]
     \hline
     Partition & Other & Num & Y/N & All & All \\ [0.5ex] 
     \hline
     First-dev & 2.87 & 2.98 & 26.30 & 12.50 & 47.98  \\ 
     
     Second-dev & 2.87 & 2.85 & 26.12 & 12.41 & 48.07 \\
     
     Third-dev & 2.92 & 2.97 & 26.37 & 12.55 & 47.93  \\
     
     Fourth-dev & 3.04 & 2.96 & 26.14 & 12.51 & 47.97  \\
     
     Fifth-dev & 3.00 & 3.20 & 26.32 & 12.59 & 47.89 \\
     
     Sixth-dev & 3.07 & 3.02 & 26.10 & 12.52 & 47.96  \\
     
     Seventh-dev & 2.99 & 3.17 & 26.40 & 12.61 & 47.87  \\
     \hline
     First-std & 2.79 & 2.81 & 26.14 & 12.41 & 47.91 \\
     \hline

     Original-dev  & 49.14 & 38.35 & 79.63 & 60.48 & -\\
     Original-std  & 49.15 & 36.52 & 79.45 & 60.32 & - \\

     \hline
    \end{tabular}}
    \centering
    \captionsetup{justification=centering}
    \caption{HieCoAtt (Alt,VGG19) model evaluation results.}
\end{subtable}%
    \hfil
\begin{subtable}[t]{0.3\linewidth}

\centering
\scalebox{0.53}{
    \begin{tabular}{ c | c c c c | c} 
     Task Type &    & \multicolumn{3}{c}{Open-Ended (BLEU-2)} &  \\ [0.5ex]
     \hline
     Method &    & \multicolumn{3}{c}{MLB with Attention} &  \\ [0.5ex]
     \hline
     Test Set&  \multicolumn{4}{c}{dev} & diff \\ [0.5ex]
     \hline
     Partition & Other & Num & Y/N & All & All \\ [0.5ex] 
     \hline
     First-dev & 2.68 & 2.27 & 24.15 & 11.45 & 54.34  \\ 
     
     Second-dev & 2.82 & 2.28 & 24.22 & 11.54 & 54.25 \\
     
     Third-dev & 2.84 & 2.17 & 24.24 & 11.55 & 54.24  \\
     
     Fourth-dev & 2.82 & 2.15 & 24.08 & 11.47 & 54.32  \\
     
     Fifth-dev & 2.91 & 2.18 & 24.21 & 11.57 & 54.22 \\
     
     Sixth-dev & 2.83 & 2.32 & 24.12 & 11.51 & 54.28  \\
     
     Seventh-dev & 2.81 & 2.42 & 24.20 & 11.55 & 54.13  \\
     \hline
     First-std & 2.59 & 2.11 & 24.31 & 11.49 & 54.19 \\
     \hline
     
     Original-dev  & 57.01 & 37.51 & 83.54 & 65.79 & -\\
     Original-std  & 56.60 & 36.63 & 83.68 & 65.68 & - \\

     \hline
    \end{tabular}}
    \centering
    \captionsetup{justification=centering}
    \caption{MLB with Attention model evaluation results.}
    
\scalebox{0.53}{
    \begin{tabular}{ c | c c c c | c} 
     Task Type &    & \multicolumn{3}{c}{Open-Ended (BLEU-2)} &  \\ [0.5ex]
     \hline
     Method &    & \multicolumn{3}{c}{MUTAN with Attention} &  \\ [0.5ex]
     \hline
     Test Set&  \multicolumn{4}{c}{dev} & diff \\ [0.5ex]
     \hline
     Partition & Other & Num & Y/N & All & All \\ [0.5ex] 
     \hline
     First-dev & 2.03 & 2.14 & 24.38 & 11.21 & 54.77  \\ 
     
     Second-dev & 2.15 & 2.19 & 24.20 & 11.20 & 54.78 \\
     
     Third-dev & 2.07 & 2.31 & 24.29 & 11.21 & 54.77  \\
     
     Fourth-dev & 2.09 & 2.19 & 23.89 & 11.05 & 54.93  \\
     
     Fifth-dev & 2.14 & 2.30 & 24.15 & 11.19 & 54.79 \\
     
     Sixth-dev & 2.17 & 2.22 & 24.17 & 11.21 & 54.77  \\
     
     Seventh-dev & 1.95 & 2.38 & 24.20 & 11.13 & 54.85  \\
     \hline
     First-std & 1.92 & 2.16 & 24.41 & 11.21 & 54.56 \\
     \hline
     
     Original-dev  & 56.73 & 38.35 & 84.11 & 65.98 & -\\
     Original-std  & 56.29 & 37.47 & 84.04 & 65.77 & - \\

     \hline
    \end{tabular}}
    \centering
    \captionsetup{justification=centering}
    \caption{MUTAN with Attention model evaluation results.}
\end{subtable}%
    \hfil
\begin{subtable}[t]{0.3\linewidth}
        
\centering
\scalebox{0.53}{
    \begin{tabular}{ c | c c c c | c} 
     Task Type &    & \multicolumn{3}{c}{Open-Ended (BLEU-2)} &  \\ [0.5ex]
     \hline
     Method &    & \multicolumn{3}{c}{HieCoAtt (Alt,Resnet200)} &  \\ [0.5ex]
     \hline
     Test Set&  \multicolumn{4}{c}{dev} & diff \\ [0.5ex]
     \hline
     Partition & Other & Num & Y/N & All & All \\ [0.5ex] 
     \hline
     First-dev & 3.26 & 3.06 & 26.39 & 12.73 & 49.08  \\ 
     
     Second-dev & 3.22 & 3.19 & 26.22 & 12.66 & 49.15 \\
     
     Third-dev & 3.36 & 2.94 & 26.41 & 12.78 & 49.03  \\
     
     Fourth-dev & 3.43 & 3.02 & 25.97 & 12.64 & 49.17  \\
     
     Fifth-dev & 3.43 & 2.95 & 26.29 & 12.76 & 49.05 \\
     
     Sixth-dev & 3.42 & 2.88 & 26.31 & 12.76 & 49.05  \\
     
     Seventh-dev & 3.32 & 3.11 & 26.51 & 12.81 & 49.00  \\
     \hline
     First-std & 3.05 & 2.85 & 26.18 & 12.56 & 49.50\\
     \hline

     Original-dev  & 51.77 & 38.65 & 79.70 & 61.81 & -\\
     Original-std  & 51.95 & 38.22 & 79.95 & 62.06 & - \\

     \hline
    \end{tabular}}
    \centering
    \captionsetup{justification=centering}
    \caption{HieCoAtt (Alt,Resnet200) model evaluation results.}

\scalebox{0.53}{
    \begin{tabular}{ c | c c c c | c} 
     Task Type &    & \multicolumn{3}{c}{Open-Ended (BLEU-2)} &  \\ [0.5ex]
     \hline
     Method &    & \multicolumn{3}{c}{LSTM Q+I} &  \\ [0.5ex]
     \hline
     Test Set&  \multicolumn{4}{c}{dev} & diff \\ [0.5ex]
     \hline
     Partition & Other & Num & Y/N & All & All \\ [0.5ex] 
     \hline
     First-dev & 1.91 & 3.27 & 26.13 & 12.00 & 46.02  \\ 
     
     Second-dev & 1.89 & 3.25 & 26.06 & 11.96 & 46.06 \\
     
     Third-dev & 1.85 & 3.24 & 26.23 & 12.01 & 46.01  \\
     
     Fourth-dev & 1.93 & 3.34 & 25.80 & 11.88 & 46.14  \\
     
     Fifth-dev & 1.93 & 3.37 & 25.85 & 11.90 & 46.12 \\
     
     Sixth-dev & 1.95 & 3.41 & 26.04 & 11.99 & 46.03  \\
     
     Seventh-dev & 1.86 & 3.28 & 26.00 & 11.92 & 46.10  \\
     \hline
     First-std & 1.98 & 2.80 & 26.39 & 12.13 & 46.05 \\
     \hline
     
     Original-dev  & 43.40 & 36.46 & 80.87 & 58.02 & -\\
     Original-std  & 43.90 & 36.67 & 80.38 & 58.18 & - \\

     \hline
    \end{tabular}}
    \centering
    \captionsetup{justification=centering}
    \caption{LSTM Q+I model evaluation results.}
\end{subtable}
\caption{The table shows the six state-of-the-art pretrained VQA models evaluation results on the GBQD and VQA dataset. ``-'' indicates the results are not available, ``-std'' represents the accuracy of VQA model evaluated on the complete testing set of GBQD and VQA dataset and ``-dev'' indicates the accuracy of VQA model evaluated on the partial testing set of GBQD and VQA dataset. In addition, $diff = Original_{dev_{All}} - X_{dev_{All}}$, where $X$ is equal to the ``First'', ``Second'', etc.}
\label{table:table9}
\end{table*}

\begin{table*}
\renewcommand\arraystretch{1.28}
\setlength\tabcolsep{11pt}
    \centering
\begin{subtable}[t]{0.3\linewidth}
\centering
\scalebox{0.53}{
    \begin{tabular}{ c | c c c c | c} 
     Task Type &    & \multicolumn{3}{c}{Open-Ended (BLEU-3)} &  \\ [0.5ex]
     \hline
     Method &    & \multicolumn{3}{c}{MUTAN without Attention} &  \\ [0.5ex]
     \hline
     Test Set&  \multicolumn{4}{c}{dev} & diff \\ [0.5ex]
     \hline
     Partition & Other & Num & Y/N & All & All \\ [0.5ex] 
     \hline
     First-dev & 2.63 & 2.72 & 24.77 & 11.73 & 48.43  \\ 
     
     Second-dev & 2.66 & 2.67 & 24.72 & 11.71 & 48.45 \\
     
     Third-dev & 2.71 & 2.53 & 24.66 & 11.70 & 48.46  \\
     
     Fourth-dev & 2.66 & 2.81 & 24.59 & 11.68 & 48.48  \\
     
     Fifth-dev & 2.72 & 2.64 & 25.00 & 11.85 & 48.31 \\
     
     Sixth-dev & 2.58 & 2.64 & 24.72 & 11.67 & 48.49  \\
     
     Seventh-dev & 2.73 & 2.56 & 24.78 & 11.76 & 48.40  \\
     \hline
     First-std & 2.60 & 3.04 & 24.33 & 11.60 & 48.85 \\
     \hline
     
     Original-dev  & 47.16 & 37.32 & 81.45 & 60.16 & -\\
     Original-std  & 47.57 & 36.75 & 81.56 & 60.45 & - \\

     \hline
    \end{tabular}}
    \centering
    \captionsetup{justification=centering}
    \caption{MUTAN without Attention model evaluation results.}

\scalebox{0.53}{
    \begin{tabular}{ c | c c c c | c} 
     Task Type &    & \multicolumn{3}{c}{Open-Ended (BLEU-3)} &  \\ [0.5ex]
     \hline
     Method &    & \multicolumn{3}{c}{HieCoAtt (Alt,VGG19)} &  \\ [0.5ex]
     \hline
     Test Set&  \multicolumn{4}{c}{dev} & diff \\ [0.5ex]
     \hline
     Partition & Other & Num & Y/N & All & All \\ [0.5ex] 
     \hline
     First-dev & 2.85 & 2.81 & 26.42 & 12.52 & 47.96  \\ 
     
     Second-dev & 2.96 & 2.90 & 26.52 & 12.62 & 47.86 \\
     
     Third-dev & 2.98 & 2.91 & 26.47 & 12.61 & 47.87  \\
     
     Fourth-dev & 3.03 & 3.05 & 26.52 & 12.67 & 47.81  \\
     
     Fifth-dev & 3.02 & 3.19 & 26.55 & 12.69 & 47.79 \\
     
     Sixth-dev & 3.17 & 3.27 & 26.41 & 12.72 & 47.76  \\
     
     Seventh-dev & 3.21 & 3.03 & 26.36 & 12.70 & 47.78  \\
     \hline
     First-std & 2.80 & 2.91 & 25.99 & 12.37 & 47.95 \\
     \hline

     Original-dev  & 49.14 & 38.35 & 79.63 & 60.48 & -\\
     Original-std  & 49.15 & 36.52 & 79.45 & 60.32 & - \\

     \hline
    \end{tabular}}
    \centering
    \captionsetup{justification=centering}
    \caption{HieCoAtt (Alt,VGG19) model evaluation results.}
\end{subtable}%
    \hfil
\begin{subtable}[t]{0.3\linewidth}

\centering
\scalebox{0.53}{
    \begin{tabular}{ c | c c c c | c} 
     Task Type &    & \multicolumn{3}{c}{Open-Ended (BLEU-3)} &  \\ [0.5ex]
     \hline
     Method &    & \multicolumn{3}{c}{MLB with Attention} &  \\ [0.5ex]
     \hline
     Test Set&  \multicolumn{4}{c}{dev} & diff \\ [0.5ex]
     \hline
     Partition & Other & Num & Y/N & All & All \\ [0.5ex] 
     \hline
     First-dev & 2.77 & 2.00 & 24.43 & 11.58 & 54.21  \\ 
     
     Second-dev & 2.84 & 2.09 & 24.19 & 11.52 & 54.27 \\
     
     Third-dev & 2.92 & 1.93 & 24.01 & 11.47 & 54.32  \\
     
     Fourth-dev & 2.97 & 1.97 & 24.03 & 11.51 & 54.28  \\
     
     Fifth-dev & 2.90 & 1.97 & 23.92 & 11.43 & 54.36 \\
     
     Sixth-dev & 2.90 & 2.12 & 24.02 & 11.49 & 54.30  \\
     
     Seventh-dev & 2.96 & 2.06 & 23.80 & 11.42 & 54.37  \\
     \hline
     First-std & 2.65 & 2.23 & 24.20 & 11.48 & 54.20 \\
     \hline
     
     Original-dev  & 57.01 & 37.51 & 83.54 & 65.79 & -\\
     Original-std  & 56.60 & 36.63 & 83.68 & 65.68 & - \\

     \hline
    \end{tabular}}
    \centering
    \captionsetup{justification=centering}
    \caption{MLB with Attention model evaluation results.}
    
\scalebox{0.53}{
    \begin{tabular}{ c | c c c c | c} 
     Task Type &    & \multicolumn{3}{c}{Open-Ended (BLEU-3)} &  \\ [0.5ex]
     \hline
     Method &    & \multicolumn{3}{c}{MUTAN with Attention} &  \\ [0.5ex]
     \hline
     Test Set&  \multicolumn{4}{c}{dev} & diff \\ [0.5ex]
     \hline
     Partition & Other & Num & Y/N & All & All \\ [0.5ex] 
     \hline
     First-dev & 2.01 & 2.18 & 24.36 & 11.20 & 54.78  \\ 
     
     Second-dev & 2.09 & 2.12 & 24.06 & 11.11 & 54.87 \\
     
     Third-dev & 2.08 & 2.15 & 24.25 & 11.19 & 54.79  \\
     
     Fourth-dev & 2.14 & 2.09 & 24.08 & 11.14 & 54.84  \\
     
     Fifth-dev & 2.05 & 2.00 & 24.10 & 11.09 & 54.89 \\
     
     Sixth-dev & 2.04 & 2.25 & 24.20 & 11.16 & 54.82  \\
     
     Seventh-dev & 2.06 & 2.26 & 23.87 & 11.03 & 54.95  \\
     \hline
     First-std & 2.06 & 2.15 & 24.13 & 11.16 & 54.61 \\
     \hline
     
     Original-dev  & 56.73 & 38.35 & 84.11 & 65.98 & -\\
     Original-std  & 56.29 & 37.47 & 84.04 & 65.77 & - \\

     \hline
    \end{tabular}}
    \centering
    \captionsetup{justification=centering}
    \caption{MUTAN with Attention model evaluation results.}
\end{subtable}%
    \hfil
\begin{subtable}[t]{0.3\linewidth}
        
\centering
\scalebox{0.53}{
    \begin{tabular}{ c | c c c c | c} 
     Task Type &    & \multicolumn{3}{c}{Open-Ended (BLEU-3)} &  \\ [0.5ex]
     \hline
     Method &    & \multicolumn{3}{c}{HieCoAtt (Alt,Resnet200)} &  \\ [0.5ex]
     \hline
     Test Set&  \multicolumn{4}{c}{dev} & diff \\ [0.5ex]
     \hline
     Partition & Other & Num & Y/N & All & All \\ [0.5ex] 
     \hline
     First-dev & 3.33 & 3.07 & 26.58 & 12.84 & 48.97  \\ 
     
     Second-dev & 3.25 & 3.04 & 26.09 & 12.64 & 49.17 \\
     
     Third-dev & 3.48 & 3.00 & 26.53 & 12.89 & 48.92  \\
     
     Fourth-dev & 3.43 & 2.99 & 26.40 & 12.81 & 49.00  \\
     
     Fifth-dev & 3.45 & 3.09 & 26.35 & 12.81 & 49.00 \\
     
     Sixth-dev & 3.41 & 2.99 & 26.62 & 12.89 & 48.92  \\
     
     Seventh-dev & 3.46 & 2.95 & 26.32 & 12.79 & 49.02  \\
     \hline
     First-std & 3.27 & 2.90 & 26.23 & 12.69 & 49.37\\
     \hline

     Original-dev  & 51.77 & 38.65 & 79.70 & 61.81 & -\\
     Original-std  & 51.95 & 38.22 & 79.95 & 62.06 & - \\

     \hline
    \end{tabular}}
    \centering
    \captionsetup{justification=centering}
    \caption{HieCoAtt (Alt,Resnet200) model evaluation results.}

\scalebox{0.53}{
    \begin{tabular}{ c | c c c c | c} 
     Task Type &    & \multicolumn{3}{c}{Open-Ended (BLEU-3)} &  \\ [0.5ex]
     \hline
     Method &    & \multicolumn{3}{c}{LSTM Q+I} &  \\ [0.5ex]
     \hline
     Test Set&  \multicolumn{4}{c}{dev} & diff \\ [0.5ex]
     \hline
     Partition & Other & Num & Y/N & All & All \\ [0.5ex] 
     \hline
     First-dev & 2.02 & 3.23 & 26.32 & 12.12 & 45.90  \\ 
     
     Second-dev & 2.08 & 3.14 & 26.01 & 12.02 & 46.00 \\
     
     Third-dev & 1.96 & 3.26 & 26.12 & 12.02 & 46.00  \\
     
     Fourth-dev & 2.05 & 3.28 & 25.95 & 11.99 & 46.03  \\
     
     Fifth-dev & 2.07 & 3.36 & 26.26 & 12.14 & 45.88 \\
     
     Sixth-dev & 2.10 & 3.29 & 25.93 & 12.01 & 46.01  \\
     
     Seventh-dev & 2.15 & 3.19 & 26.12 & 12.10 & 45.92  \\
     \hline
     First-std & 1.88 & 3.26 & 25.96 & 11.95 & 46.23 \\
     \hline
     
     Original-dev  & 43.40 & 36.46 & 80.87 & 58.02 & -\\
     Original-std  & 43.90 & 36.67 & 80.38 & 58.18 & - \\

     \hline
    \end{tabular}}
    \centering
    \captionsetup{justification=centering}
    \caption{LSTM Q+I model evaluation results.}
\end{subtable}
\caption{The table shows the six state-of-the-art pretrained VQA models evaluation results on the GBQD and VQA dataset. ``-'' indicates the results are not available, ``-std'' represents the accuracy of VQA model evaluated on the complete testing set of GBQD and VQA dataset and ``-dev'' indicates the accuracy of VQA model evaluated on the partial testing set of GBQD and VQA dataset. In addition, $diff = Original_{dev_{All}} - X_{dev_{All}}$, where $X$ is equal to the ``First'', ``Second'', etc.}
\label{table:table10}
\end{table*}

\begin{table*}
\renewcommand\arraystretch{1.28}
\setlength\tabcolsep{11pt}
    \centering
\begin{subtable}[t]{0.3\linewidth}
\centering
\scalebox{0.53}{
    \begin{tabular}{ c | c c c c | c} 
     Task Type &    & \multicolumn{3}{c}{Open-Ended (BLEU-4)} &  \\ [0.5ex]
     \hline
     Method &    & \multicolumn{3}{c}{MUTAN without Attention} &  \\ [0.5ex]
     \hline
     Test Set&  \multicolumn{4}{c}{dev} & diff \\ [0.5ex]
     \hline
     Partition & Other & Num & Y/N & All & All \\ [0.5ex] 
     \hline
     First-dev & 2.46 & 2.98 & 25.22 & 11.86 & 48.30  \\ 
     
     Second-dev & 2.47 & 3.05 & 25.23 & 11.88 & 48.28 \\
     
     Third-dev & 2.62 & 2.90 & 24.95 & 11.81 & 48.35  \\
     
     Fourth-dev & 2.71 & 2.96 & 24.87 & 11.83 & 48.33  \\
     
     Fifth-dev & 2.70 & 3.03 & 25.08 & 11.92 & 48.24 \\
     
     Sixth-dev & 2.65 & 2.84 & 25.30 & 11.97 & 48.19  \\
     
     Seventh-dev & 2.71 & 2.99 & 25.01 & 11.89 & 48.27  \\
     \hline
     First-std & 2.51 & 2.36 & 24.36 & 11.50 & 48.95 \\
     \hline
     
     Original-dev  & 47.16 & 37.32 & 81.45 & 60.16 & -\\
     Original-std  & 47.57 & 36.75 & 81.56 & 60.45 & - \\

     \hline
    \end{tabular}}
    \centering
    \captionsetup{justification=centering}
    \caption{MUTAN without Attention model evaluation results.}

\scalebox{0.53}{
    \begin{tabular}{ c | c c c c | c} 
     Task Type &    & \multicolumn{3}{c}{Open-Ended (BLEU-4)} &  \\ [0.5ex]
     \hline
     Method &    & \multicolumn{3}{c}{HieCoAtt (Alt,VGG19)} &  \\ [0.5ex]
     \hline
     Test Set&  \multicolumn{4}{c}{dev} & diff \\ [0.5ex]
     \hline
     Partition & Other & Num & Y/N & All & All \\ [0.5ex] 
     \hline
     First-dev & 2.80 & 3.17 & 26.55 & 12.59 & 47.89  \\ 
     
     Second-dev & 2.87 & 3.14 & 27.02 & 12.81 & 47.67 \\
     
     Third-dev & 3.02 & 2.93 & 26.60 & 12.69 & 47.79  \\
     
     Fourth-dev & 3.08 & 3.14 & 26.29 & 12.61 & 47.87  \\
     
     Fifth-dev & 3.09 & 3.28 & 26.52 & 12.73 & 47.75 \\
     
     Sixth-dev & 3.11 & 3.20 & 26.66 & 12.78 & 47.70  \\
     
     Seventh-dev & 3.03 & 3.26 & 26.71 & 12.77 & 47.71  \\
     \hline
     First-std & 2.73 & 2.46 & 25.81 & 12.21 & 48.11 \\
     \hline

     Original-dev  & 49.14 & 38.35 & 79.63 & 60.48 & -\\
     Original-std  & 49.15 & 36.52 & 79.45 & 60.32 & - \\

     \hline
    \end{tabular}}
    \centering
    \captionsetup{justification=centering}
    \caption{HieCoAtt (Alt,VGG19) model evaluation results.}
\end{subtable}%
    \hfil
\begin{subtable}[t]{0.3\linewidth}

\centering
\scalebox{0.53}{
    \begin{tabular}{ c | c c c c | c} 
     Task Type &    & \multicolumn{3}{c}{Open-Ended (BLEU-4)} &  \\ [0.5ex]
     \hline
     Method &    & \multicolumn{3}{c}{MLB with Attention} &  \\ [0.5ex]
     \hline
     Test Set&  \multicolumn{4}{c}{dev} & diff \\ [0.5ex]
     \hline
     Partition & Other & Num & Y/N & All & All \\ [0.5ex] 
     \hline
     First-dev & 2.65 & 2.41 & 24.63 & 11.64 & 54.15  \\ 
     
     Second-dev & 2.72 & 2.47 & 24.63 & 11.69 & 54.10 \\
     
     Third-dev & 2.83 & 2.40 & 24.62 & 11.73 & 54.06  \\
     
     Fourth-dev & 2.88 & 2.38 & 24.28 & 11.61 & 54.18  \\
     
     Fifth-dev & 2.79 & 2.31 & 24.40 & 11.61 & 54.18 \\
     
     Sixth-dev & 2.89 & 2.36 & 24.31 & 11.63 & 54.16  \\
     
     Seventh-dev & 2.80 & 2.51 & 24.52 & 11.68 & 54.11  \\
     \hline
     First-std & 2.58 & 1.85 & 23.54 & 11.14 & 54.54 \\
     \hline
     
     Original-dev  & 57.01 & 37.51 & 83.54 & 65.79 & -\\
     Original-std  & 56.60 & 36.63 & 83.68 & 65.68 & - \\

     \hline
    \end{tabular}}
    \centering
    \captionsetup{justification=centering}
    \caption{MLB with Attention model evaluation results.}
    
\scalebox{0.53}{
    \begin{tabular}{ c | c c c c | c} 
     Task Type &    & \multicolumn{3}{c}{Open-Ended (BLEU-4)} &  \\ [0.5ex]
     \hline
     Method &    & \multicolumn{3}{c}{MUTAN with Attention} &  \\ [0.5ex]
     \hline
     Test Set&  \multicolumn{4}{c}{dev} & diff \\ [0.5ex]
     \hline
     Partition & Other & Num & Y/N & All & All \\ [0.5ex] 
     \hline
     First-dev & 2.00 & 2.35 & 24.55 & 11.29 & 54.69  \\ 
     
     Second-dev & 2.05 & 2.21 & 24.30 & 11.20 & 54.78 \\
     
     Third-dev & 2.01 & 2.32 & 24.61 & 11.32 & 54.66  \\
     
     Fourth-dev & 2.11 & 2.39 & 24.18 & 11.20 & 54.78  \\
     
     Fifth-dev & 1.94 & 2.37 & 24.47 & 11.23 & 54.75 \\
     
     Sixth-dev & 2.08 & 2.43 & 24.39 & 11.27 & 54.71  \\
     
     Seventh-dev & 2.00 & 2.35 & 24.23 & 11.16 & 54.82  \\
     \hline
     First-std & 1.98 & 1.94 & 23.62 & 10.90 & 54.87 \\
     \hline
     
     Original-dev  & 56.73 & 38.35 & 84.11 & 65.98 & -\\
     Original-std  & 56.29 & 37.47 & 84.04 & 65.77 & - \\

     \hline
    \end{tabular}}
    \centering
    \captionsetup{justification=centering}
    \caption{MUTAN with Attention model evaluation results.}
\end{subtable}%
    \hfil
\begin{subtable}[t]{0.3\linewidth}
        
\centering
\scalebox{0.53}{
    \begin{tabular}{ c | c c c c | c} 
     Task Type &    & \multicolumn{3}{c}{Open-Ended (BLEU-4)} &  \\ [0.5ex]
     \hline
     Method &    & \multicolumn{3}{c}{HieCoAtt (Alt,Resnet200)} &  \\ [0.5ex]
     \hline
     Test Set&  \multicolumn{4}{c}{dev} & diff \\ [0.5ex]
     \hline
     Partition & Other & Num & Y/N & All & All \\ [0.5ex] 
     \hline
     First-dev & 3.01& 3.37 & 26.55 & 12.71 & 49.10  \\ 
     
     Second-dev & 3.08 & 3.34 & 26.84 & 12.86 & 48.95 \\
     
     Third-dev & 3.23 & 3.03 & 26.71 & 12.85 & 48.96  \\
     
     Fourth-dev & 3.24 & 3.16 & 26.31 & 12.70 & 49.11  \\
     
     Fifth-dev & 3.35 & 3.10 & 26.16 & 12.68 & 49.13 \\
     
     Sixth-dev & 3.34 & 3.25 & 26.66 & 12.90 & 48.91  \\
     
     Seventh-dev & 3.21 & 3.23 & 26.56 & 12.79 & 49.02  \\
     \hline
     First-std & 3.25 & 2.52 & 25.84 & 12.49 & 49.57\\
     \hline

     Original-dev  & 51.77 & 38.65 & 79.70 & 61.81 & -\\
     Original-std  & 51.95 & 38.22 & 79.95 & 62.06 & - \\

     \hline
    \end{tabular}}
    \centering
    \captionsetup{justification=centering}
    \caption{HieCoAtt (Alt,Resnet200) model evaluation results.}

\scalebox{0.53}{
    \begin{tabular}{ c | c c c c | c} 
     Task Type &    & \multicolumn{3}{c}{Open-Ended (BLEU-4)} &  \\ [0.5ex]
     \hline
     Method &    & \multicolumn{3}{c}{LSTM Q+I} &  \\ [0.5ex]
     \hline
     Test Set&  \multicolumn{4}{c}{dev} & diff \\ [0.5ex]
     \hline
     Partition & Other & Num & Y/N & All & All \\ [0.5ex] 
     \hline
     First-dev & 1.84 & 3.20 & 26.41 & 12.07 & 45.95  \\ 
     
     Second-dev & 1.87 & 3.22 & 26.38 & 12.08 & 45.94 \\
     
     Third-dev & 1.93 & 3.28 & 26.41 & 12.12 & 45.90  \\
     
     Fourth-dev & 1.85 & 3.24 & 26.16 & 11.98 & 46.04  \\
     
     Fifth-dev & 1.91 & 3.32 & 26.26 & 12.06 & 45.96 \\
     
     Sixth-dev & 1.90 & 3.27 & 26.16 & 12.00 & 46.02  \\
     
     Seventh-dev & 1.97 & 3.31 & 26.07 & 12.00 & 46.02  \\
     \hline
     First-std & 2.03 & 2.86 & 25.73 & 11.88 & 46.30 \\
     \hline
     
     Original-dev  & 43.40 & 36.46 & 80.87 & 58.02 & -\\
     Original-std  & 43.90 & 36.67 & 80.38 & 58.18 & - \\

     \hline
    \end{tabular}}
    \centering
    \captionsetup{justification=centering}
    \caption{LSTM Q+I model evaluation results.}
\end{subtable}
\caption{The table shows the six state-of-the-art pretrained VQA models evaluation results on the GBQD and VQA dataset. ``-'' indicates the results are not available, ``-std'' represents the accuracy of VQA model evaluated on the complete testing set of GBQD and VQA dataset and ``-dev'' indicates the accuracy of VQA model evaluated on the partial testing set of GBQD and VQA dataset. In addition, $diff = Original_{dev_{All}} - X_{dev_{All}}$, where $X$ is equal to the ``First'', ``Second'', etc.}
\label{table:table11}
\end{table*}

\begin{table*}
\renewcommand\arraystretch{1.28}
\setlength\tabcolsep{11pt}
    \centering
\begin{subtable}[t]{0.3\linewidth}
\centering
\scalebox{0.53}{
    \begin{tabular}{ c | c c c c | c} 
     Task Type &    & \multicolumn{3}{c}{Open-Ended (ROUGE)} &  \\ [0.5ex]
     \hline
     Method &    & \multicolumn{3}{c}{MUTAN without Attention} &  \\ [0.5ex]
     \hline
     Test Set&  \multicolumn{4}{c}{dev} & diff \\ [0.5ex]
     \hline
     Partition & Other & Num & Y/N & All & All \\ [0.5ex] 
     \hline
     First-dev & 2.68 & 2.66 & 26.41 & 12.42 & 47.74  \\ 
     
     Second-dev & 3.40 & 3.12 & 25.47 & 12.43 & 47.73 \\
     
     Third-dev & 3.38 & 2.60 & 21.83 & 10.87 & 49.29  \\
     
     Fourth-dev & 3.04 & 2.17 & 23.19 & 11.21 & 48.95  \\
     
     Fifth-dev & 2.93 & 2.77 & 26.22 & 12.47 & 47.69 \\
     
     Sixth-dev & 2.43 & 2.66 & 27.14 & 12.60 & 47.56  \\
     
     Seventh-dev & 1.66 & 2.73 & 26.90 & 12.13 & 48.03  \\
     \hline
     First-std & 2.69 & 2.57 & 26.20 & 12.36 & 48.09 \\
     \hline
     
     Original-dev  & 47.16 & 37.32 & 81.45 & 60.16 & -\\
     Original-std  & 47.57 & 36.75 & 81.56 & 60.45 & - \\

     \hline
    \end{tabular}}
    \centering
    \captionsetup{justification=centering}
    \caption{MUTAN without Attention model evaluation results.}

\scalebox{0.53}{
    \begin{tabular}{ c | c c c c | c} 
     Task Type &    & \multicolumn{3}{c}{Open-Ended (ROUGE)} &  \\ [0.5ex]
     \hline
     Method &    & \multicolumn{3}{c}{HieCoAtt (Alt,VGG19)} &  \\ [0.5ex]
     \hline
     Test Set&  \multicolumn{4}{c}{dev} & diff \\ [0.5ex]
     \hline
     Partition & Other & Num & Y/N & All & All \\ [0.5ex] 
     \hline
     First-dev & 2.88 & 3.73 & 24.78 & 11.96 & 48.52  \\ 
     
     Second-dev & 3.26 & 3.75 & 27.49 & 13.26 & 47.22 \\
     
     Third-dev & 3.11 & 3.41 & 27.73 & 13.25 & 47.23  \\
     
     Fourth-dev & 3.05 & 3.20 & 25.74 & 12.38 & 48.10  \\
     
     Fifth-dev & 3.13 & 3.56 & 28.27 & 13.49 & 46.99 \\
     
     Sixth-dev & 3.33 & 3.35 & 27.67 & 13.32 & 47.16  \\
     
     Seventh-dev & 2.78 & 3.58 & 28.09 & 13.25 & 47.23  \\
     \hline
     First-std & 2.73 & 3.41 & 24.01 & 11.57 & 48.75 \\
     \hline

     Original-dev  & 49.14 & 38.35 & 79.63 & 60.48 & -\\
     Original-std  & 49.15 & 36.52 & 79.45 & 60.32 & - \\

     \hline
    \end{tabular}}
    \centering
    \captionsetup{justification=centering}
    \caption{HieCoAtt (Alt,VGG19) model evaluation results.}
\end{subtable}%
    \hfil
\begin{subtable}[t]{0.3\linewidth}

\centering
\scalebox{0.53}{
    \begin{tabular}{ c | c c c c | c} 
     Task Type &    & \multicolumn{3}{c}{Open-Ended (ROUGE)} &  \\ [0.5ex]
     \hline
     Method &    & \multicolumn{3}{c}{MLB with Attention} &  \\ [0.5ex]
     \hline
     Test Set&  \multicolumn{4}{c}{dev} & diff \\ [0.5ex]
     \hline
     Partition & Other & Num & Y/N & All & All \\ [0.5ex] 
     \hline
     First-dev & 3.08 & 2.59 & 24.70 & 11.90 & 53.89  \\ 
     
     Second-dev & 3.20 & 2.88 & 24.81 & 12.03 & 53.76 \\
     
     Third-dev & 3.10 & 2.63 & 22.57 & 11.04 & 54.75  \\
     
     Fourth-dev & 3.23 & 2.60 & 22.82 & 11.20 & 54.59  \\
     
     Fifth-dev & 3.20 & 2.42 & 24.75 & 11.96 & 53.83 \\
     
     Sixth-dev & 2.92 & 2.61 & 24.49 & 11.74 & 54.05  \\
     
     Seventh-dev & 2.67 & 2.62 & 27.49 & 12.85 & 52.94  \\
     \hline
     First-std & 2.94 & 2.38 & 24.16 & 11.63 & 54.05 \\
     \hline
     
     Original-dev  & 57.01 & 37.51 & 83.54 & 65.79 & -\\
     Original-std  & 56.60 & 36.63 & 83.68 & 65.68 & - \\

     \hline
    \end{tabular}}
    \centering
    \captionsetup{justification=centering}
    \caption{MLB with Attention model evaluation results.}
    
\scalebox{0.53}{
    \begin{tabular}{ c | c c c c | c} 
     Task Type &    & \multicolumn{3}{c}{Open-Ended (ROUGE)} &  \\ [0.5ex]
     \hline
     Method &    & \multicolumn{3}{c}{MUTAN with Attention} &  \\ [0.5ex]
     \hline
     Test Set&  \multicolumn{4}{c}{dev} & diff \\ [0.5ex]
     \hline
     Partition & Other & Num & Y/N & All & All \\ [0.5ex] 
     \hline
     First-dev & 2.14 & 2.02 & 25.38 & 11.66 & 54.32  \\ 
     
     Second-dev & 2.12 & 2.01 & 22.84 & 10.70 & 55.28 \\
     
     Third-dev & 2.13 & 2.37 & 22.73 & 10.61 & 55.37  \\
     
     Fourth-dev & 2.04 & 2.26 & 22.70 & 10.55 & 55.43  \\
     
     Fifth-dev & 1.97 & 2.26 & 22.72 & 10.52 & 55.46 \\
     
     Sixth-dev & 2.25 & 2.63 & 23.91 & 11.18 & 54.80  \\
     
     Seventh-dev & 1.93 & 2.63 & 25.10 & 11.51 & 54.47  \\
     \hline
     First-std & 2.04 & 1.88 & 24.83 & 11.42 & 54.35 \\
     \hline
     
     Original-dev  & 56.73 & 38.35 & 84.11 & 65.98 & -\\
     Original-std  & 56.29 & 37.47 & 84.04 & 65.77 & - \\

     \hline
    \end{tabular}}
    \centering
    \captionsetup{justification=centering}
    \caption{MUTAN with Attention model evaluation results.}
\end{subtable}%
    \hfil
\begin{subtable}[t]{0.3\linewidth}
        
\centering
\scalebox{0.53}{
    \begin{tabular}{ c | c c c c | c} 
     Task Type &    & \multicolumn{3}{c}{Open-Ended (ROUGE)} &  \\ [0.5ex]
     \hline
     Method &    & \multicolumn{3}{c}{HieCoAtt (Alt,Resnet200)} &  \\ [0.5ex]
     \hline
     Test Set&  \multicolumn{4}{c}{dev} & diff \\ [0.5ex]
     \hline
     Partition & Other & Num & Y/N & All & All \\ [0.5ex] 
     \hline
     First-dev & 3.68 & 3.13 & 27.23 & 13.29 & 48.52  \\ 
     
     Second-dev & 3.83 & 3.76 & 27.45 & 13.52 & 48.29 \\
     
     Third-dev & 3.41 & 3.28 & 27.26 & 13.19 & 48.62  \\
     
     Fourth-dev & 3.25 & 3.37 & 25.69 & 12.47 & 49.34  \\
     
     Fifth-dev & 3.33 & 3.47 & 28.33 & 13.60 & 48.21 \\
     
     Sixth-dev & 3.56 & 2.99 & 27.83 & 13.46 & 48.35  \\
     
     Seventh-dev & 2.45 & 3.38 & 28.14 & 13.09 & 48.72  \\
     \hline
     First-std & 3.39 & 3.01 & 26.51 & 12.88 & 49.18\\
     \hline

     Original-dev  & 51.77 & 38.65 & 79.70 & 61.81 & -\\
     Original-std  & 51.95 & 38.22 & 79.95 & 62.06 & - \\

     \hline
    \end{tabular}}
    \centering
    \captionsetup{justification=centering}
    \caption{HieCoAtt (Alt,Resnet200) model evaluation results.}

\scalebox{0.53}{
    \begin{tabular}{ c | c c c c | c} 
     Task Type &    & \multicolumn{3}{c}{Open-Ended (ROUGE)} &  \\ [0.5ex]
     \hline
     Method &    & \multicolumn{3}{c}{LSTM Q+I} &  \\ [0.5ex]
     \hline
     Test Set&  \multicolumn{4}{c}{dev} & diff \\ [0.5ex]
     \hline
     Partition & Other & Num & Y/N & All & All \\ [0.5ex] 
     \hline
     First-dev & 1.71 & 3.56 & 26.51 & 12.09 & 45.93  \\ 
     
     Second-dev & 2.01 & 3.40 & 26.09 & 12.04 & 45.98 \\
     
     Third-dev & 1.91 & 2.92 & 23.70 & 10.96 & 47.06  \\
     
     Fourth-dev & 1.61 & 3.37 & 25.35 & 11.54 & 46.48  \\
     
     Fifth-dev & 1.57 & 3.32 & 25.92 & 11.75 & 46.27 \\
     
     Sixth-dev & 2.21 & 2.79 & 27.24 & 12.54 & 45.48  \\
     
     Seventh-dev & 1.58 & 2.99 & 27.26 & 12.27 & 45.75  \\
     \hline
     First-std & 1.79 & 3.42 & 26.42 & 12.11 & 46.07 \\
     \hline
     
     Original-dev  & 43.40 & 36.46 & 80.87 & 58.02 & -\\
     Original-std  & 43.90 & 36.67 & 80.38 & 58.18 & - \\

     \hline
    \end{tabular}}
    \centering
    \captionsetup{justification=centering}
    \caption{LSTM Q+I model evaluation results.}
\end{subtable}
\caption{The table shows the six state-of-the-art pretrained VQA models evaluation results on the GBQD and VQA dataset. ``-'' indicates the results are not available, ``-std'' represents the accuracy of VQA model evaluated on the complete testing set of GBQD and VQA dataset and ``-dev'' indicates the accuracy of VQA model evaluated on the partial testing set of GBQD and VQA dataset. In addition, $diff = Original_{dev_{All}} - X_{dev_{All}}$, where $X$ is equal to the ``First'', ``Second'', etc.}
\label{table:table12}
\end{table*}

\begin{table*}
\renewcommand\arraystretch{1.28}
\setlength\tabcolsep{11pt}
    \centering
\begin{subtable}[t]{0.3\linewidth}
\centering
\scalebox{0.53}{
    \begin{tabular}{ c | c c c c | c} 
     Task Type &    & \multicolumn{3}{c}{Open-Ended (CIDEr)} &  \\ [0.5ex]
     \hline
     Method &    & \multicolumn{3}{c}{MUTAN without Attention} &  \\ [0.5ex]
     \hline
     Test Set&  \multicolumn{4}{c}{dev} & diff \\ [0.5ex]
     \hline
     Partition & Other & Num & Y/N & All & All \\ [0.5ex] 
     \hline
     First-dev & 0.62 & 12.76 & 0.18 & 1.75 & 58.41  \\ 
     
     Second-dev & 2.47 & 2.89 & 21.03 & 10.13 & 50.03 \\
     
     Third-dev & 0.96 & 0.33 & 1.07 & 0.94 & 59.22  \\
     
     Fourth-dev & 1.44 & 1.79 & 12.32 & 5.94 & 54.22  \\
     
     Fifth-dev & 2.73 & 3.09 & 29.80 & 13.88 & 46.28 \\
     
     Sixth-dev & 1.56 & 2.26 & 31.64 & 11.98 & 48.18  \\
     
     Seventh-dev & 0.83 & 1.10 & 14.58 & 6.50 & 53.66  \\
     \hline
     First-std & 0.66 & 12.92 & 0.12 & 1.72 & 58.73 \\
     \hline
     
     Original-dev  & 47.16 & 37.32 & 81.45 & 60.16 & -\\
     Original-std  & 47.57 & 36.75 & 81.56 & 60.45 & - \\

     \hline
    \end{tabular}}
    \centering
    \captionsetup{justification=centering}
    \caption{MUTAN without Attention model evaluation results.}

\scalebox{0.53}{
    \begin{tabular}{ c | c c c c | c} 
     Task Type &    & \multicolumn{3}{c}{Open-Ended (CIDEr)} &  \\ [0.5ex]
     \hline
     Method &    & \multicolumn{3}{c}{HieCoAtt (Alt,VGG19)} &  \\ [0.5ex]
     \hline
     Test Set&  \multicolumn{4}{c}{dev} & diff \\ [0.5ex]
     \hline
     Partition & Other & Num & Y/N & All & All \\ [0.5ex] 
     \hline
     First-dev & 1.60 & 4.80 & 0.09 & 1.33 & 59.15  \\ 
     
     Second-dev & 2.51 & 2.00 & 20.72 & 9.93 & 50.55 \\
     
     Third-dev & 1.21 & 0.35 & 0.06 & 0.65 & 59.83  \\
     
     Fourth-dev & 5.53 & 2.03 & 6.98 & 5.75 & 54.73  \\
     
     Fifth-dev & 2.34 & 2.78 & 25.13 & 11.74 & 48.74 \\
     
     Sixth-dev & 2.43 & 3.61 & 29.75 & 13.77 & 46.71  \\
     
     Seventh-dev & 1.66 & 1.89 & 15.72 & 7.45 & 53.03  \\
     \hline
     First-std & 1.51 & 5.25 & 0.08 & 12.38 & 59.01 \\
     \hline

     Original-dev  & 49.14 & 38.35 & 79.63 & 60.48 & -\\
     Original-std  & 49.15 & 36.52 & 79.45 & 60.32 & - \\

     \hline
    \end{tabular}}
    \centering
    \captionsetup{justification=centering}
    \caption{HieCoAtt (Alt,VGG19) model evaluation results.}
\end{subtable}%
    \hfil
\begin{subtable}[t]{0.3\linewidth}

\centering
\scalebox{0.53}{
    \begin{tabular}{ c | c c c c | c} 
     Task Type &    & \multicolumn{3}{c}{Open-Ended (CIDEr)} &  \\ [0.5ex]
     \hline
     Method &    & \multicolumn{3}{c}{MLB with Attention} &  \\ [0.5ex]
     \hline
     Test Set&  \multicolumn{4}{c}{dev} & diff \\ [0.5ex]
     \hline
     Partition & Other & Num & Y/N & All & All \\ [0.5ex] 
     \hline
     First-dev & 1.57 & 21.95 & 1.61 & 3.80 & 61.99  \\ 
     
     Second-dev & 2.48 & 2.90 & 21.91 & 10.50 & 55.29 \\
     
     Third-dev & 2.09 & 2.26 & 22.52 & 10.50 & 55.29  \\
     
     Fourth-dev & 3.20 & 2.91 & 25.53 & 12.33 & 53.46  \\
     
     Fifth-dev & 2.18 & 3.37 & 27.05 & 12.51 & 53.28 \\
     
     Sixth-dev & 2.48 & 2.19 & 32.04 & 14.58 & 51.21  \\
     
     Seventh-dev & 1.68 & 2.07 & 23.26 & 10.58 & 55.21  \\
     \hline
     First-std & 1.60 & 22.19 & 1.44 & 3.68 & 62.00 \\
     \hline
     
     Original-dev  & 57.01 & 37.51 & 83.54 & 65.79 & -\\
     Original-std  & 56.60 & 36.63 & 83.68 & 65.68 & - \\

     \hline
    \end{tabular}}
    \centering
    \captionsetup{justification=centering}
    \caption{MLB with Attention model evaluation results.}
    
\scalebox{0.53}{
    \begin{tabular}{ c | c c c c | c} 
     Task Type &    & \multicolumn{3}{c}{Open-Ended (CIDEr)} &  \\ [0.5ex]
     \hline
     Method &    & \multicolumn{3}{c}{MUTAN with Attention} &  \\ [0.5ex]
     \hline
     Test Set&  \multicolumn{4}{c}{dev} & diff \\ [0.5ex]
     \hline
     Partition & Other & Num & Y/N & All & All \\ [0.5ex] 
     \hline
     First-dev & 0.81 & 15.35 & 1.44 & 2.64 & 63.34  \\ 
     
     Second-dev & 1.20 & 2.79 & 22.16 & 9.97 & 56.01 \\
     
     Third-dev & 1.73 & 3.46 & 26.13 & 11.93 & 54.05  \\
     
     Fourth-dev & 2.01 & 2.63 & 23.70 & 10.98 & 55.00  \\
     
     Fifth-dev & 1.27 & 2.86 & 27.83 & 12.34 & 53.64 \\
     
     Sixth-dev & 1.46 & 1.92 & 30.38 & 13.38 & 52.60  \\
     
     Seventh-dev & 1.42 & 2.33 & 23.47 & 10.57 & 55.41  \\
     \hline
     First-std & 0.82 & 15.92 & 1.33 & 2.60 & 63.17 \\
     \hline
     
     Original-dev  & 56.73 & 38.35 & 84.11 & 65.98 & -\\
     Original-std  & 56.29 & 37.47 & 84.04 & 65.77 & - \\

     \hline
    \end{tabular}}
    \centering
    \captionsetup{justification=centering}
    \caption{MUTAN with Attention model evaluation results.}
\end{subtable}%
    \hfil
\begin{subtable}[t]{0.3\linewidth}
        
\centering
\scalebox{0.53}{
    \begin{tabular}{ c | c c c c | c} 
     Task Type &    & \multicolumn{3}{c}{Open-Ended (CIDEr)} &  \\ [0.5ex]
     \hline
     Method &    & \multicolumn{3}{c}{HieCoAtt (Alt,Resnet200)} &  \\ [0.5ex]
     \hline
     Test Set&  \multicolumn{4}{c}{dev} & diff \\ [0.5ex]
     \hline
     Partition & Other & Num & Y/N & All & All \\ [0.5ex] 
     \hline
     First-dev & 1.36 & 17.21 & 0.74 & 2.82 & 58.99  \\ 
     
     Second-dev & 2.57 & 3.49 & 22.36 & 10.79 & 51.02 \\
     
     Third-dev & 1.36 & 2.74 & 7.53 & 4.04 & 57.77  \\
     
     Fourth-dev & 7.02 & 3.16 & 6.16 & 6.25 & 55.56  \\
     
     Fifth-dev & 2.62 & 4.43 & 24.71 & 11.88 & 49.93 \\
     
     Sixth-dev & 2.96 & 3.70 & 31.80 & 14.87 & 46.94  \\
     
     Seventh-dev & 2.20 & 3.13 & 18.06 & 8.81 & 53.00  \\
     \hline
     First-std & 1.44 & 17.21 & 0.78 & 2.81 & 59.25\\
     \hline

     Original-dev  & 51.77 & 38.65 & 79.70 & 61.81 & -\\
     Original-std  & 51.95 & 38.22 & 79.95 & 62.06 & - \\

     \hline
    \end{tabular}}
    \centering
    \captionsetup{justification=centering}
    \caption{HieCoAtt (Alt,Resnet200) model evaluation results.}

\scalebox{0.53}{
    \begin{tabular}{ c | c c c c | c} 
     Task Type &    & \multicolumn{3}{c}{Open-Ended (CIDEr)} &  \\ [0.5ex]
     \hline
     Method &    & \multicolumn{3}{c}{LSTM Q+I} &  \\ [0.5ex]
     \hline
     Test Set&  \multicolumn{4}{c}{dev} & diff \\ [0.5ex]
     \hline
     Partition & Other & Num & Y/N & All & All \\ [0.5ex] 
     \hline
     First-dev & 0.81 & 2.63 & 21.86 & 9.65 & 48.37  \\ 
     
     Second-dev & 1.88 & 2.61 & 26.16 & 11.92 & 46.10 \\
     
     Third-dev & 1.24 & 0.78 & 9.38 & 4.53 & 53.49  \\
     
     Fourth-dev & 1.65 & 1.67 & 17.29 & 8.07 & 49.95  \\
     
     Fifth-dev & 2.31 & 2.66 & 23.24 & 10.94 & 47.08 \\
     
     Sixth-dev & 1.22 & 2.28 & 30.87 & 13.50 & 44.52  \\
     
     Seventh-dev & 1.12 & 1.60 & 20.68 & 9.20 & 48.82  \\
     \hline
     First-std & 0.86 & 2.61 & 21.88 & 9.70 & 48.48 \\
     \hline
     
     Original-dev  & 43.40 & 36.46 & 80.87 & 58.02 & -\\
     Original-std  & 43.90 & 36.67 & 80.38 & 58.18 & - \\

     \hline
    \end{tabular}}
    \centering
    \captionsetup{justification=centering}
    \caption{LSTM Q+I model evaluation results.}
\end{subtable}
\caption{The table shows the six state-of-the-art pretrained VQA models evaluation results on the GBQD and VQA dataset. ``-'' indicates the results are not available, ``-std'' represents the accuracy of VQA model evaluated on the complete testing set of GBQD and VQA dataset and ``-dev'' indicates the accuracy of VQA model evaluated on the partial testing set of GBQD and VQA dataset. In addition, $diff = Original_{dev_{All}} - X_{dev_{All}}$, where $X$ is equal to the ``First'', ``Second'', etc.}
\label{table:table13}
\end{table*}

\begin{table*}
\renewcommand\arraystretch{1.28}
\setlength\tabcolsep{11pt}
    \centering
\begin{subtable}[t]{0.3\linewidth}
\centering
\scalebox{0.53}{
    \begin{tabular}{ c | c c c c | c} 
     Task Type &    & \multicolumn{3}{c}{Open-Ended (METEOR)} &  \\ [0.5ex]
     \hline
     Method &    & \multicolumn{3}{c}{MUTAN without Attention} &  \\ [0.5ex]
     \hline
     Test Set&  \multicolumn{4}{c}{dev} & diff \\ [0.5ex]
     \hline
     Partition & Other & Num & Y/N & All & All \\ [0.5ex] 
     \hline
     First-dev & 2.46 & 2.64 & 20.43 & 9.86 & 50.30  \\ 
     
     Second-dev & 2.55 & 2.56 & 18.70 & 9.18 & 50.98 \\
     
     Third-dev & 2.39 & 2.56 & 18.90 & 9.18 & 50.98  \\
     
     Fourth-dev & 2.18 & 2.43 & 20.38 & 9.68 & 50.48  \\
     
     Fifth-dev & 2.19 & 2.54 & 21.77 & 10.26 & 49.90 \\
     
     Sixth-dev & 2.05 & 2.68 & 22.66 & 10.58 & 49.58  \\
     
     Seventh-dev & 1.99 & 2.76 & 23.18 & 10.77 & 49.39  \\
     \hline
     First-std & 2.37 & 2.48 & 20.11 & 9.69 & 50.76 \\
     \hline
     
     Original-dev  & 47.16 & 37.32 & 81.45 & 60.16 & -\\
     Original-std  & 47.57 & 36.75 & 81.56 & 60.45 & - \\

     \hline
    \end{tabular}}
    \centering
    \captionsetup{justification=centering}
    \caption{MUTAN without Attention model evaluation results.}

\scalebox{0.53}{
    \begin{tabular}{ c | c c c c | c} 
     Task Type &    & \multicolumn{3}{c}{Open-Ended (METEOR)} &  \\ [0.5ex]
     \hline
     Method &    & \multicolumn{3}{c}{HieCoAtt (Alt,VGG19)} &  \\ [0.5ex]
     \hline
     Test Set&  \multicolumn{4}{c}{dev} & diff \\ [0.5ex]
     \hline
     Partition & Other & Num & Y/N & All & All \\ [0.5ex] 
     \hline
     First-dev & 2.67 & 3.09 & 20.98 & 10.23 & 50.25  \\ 
     
     Second-dev & 2.91 & 2.88 & 19.80 & 9.84 & 50.64 \\
     
     Third-dev & 2.83 & 3.16 & 20.14 & 9.97 & 50.51  \\
     
     Fourth-dev & 2.68 & 3.31 & 21.24 & 10.37 & 50.11  \\
     
     Fifth-dev & 2.60 & 3.13 & 22.05 & 10.64 & 49.84 \\
     
     Sixth-dev & 2.54 & 3.23 & 22.88 & 10.96 & 49.52  \\
     
     Seventh-dev & 2.56 & 3.23 & 22.94 & 11.00 & 49.48  \\
     \hline
     First-std & 2.65 & 2.89 & 20.92 & 10.21 & 50.11 \\
     \hline

     Original-dev  & 49.14 & 38.35 & 79.63 & 60.48 & -\\
     Original-std  & 49.15 & 36.52 & 79.45 & 60.32 & - \\

     \hline
    \end{tabular}}
    \centering
    \captionsetup{justification=centering}
    \caption{HieCoAtt (Alt,VGG19) model evaluation results.}
\end{subtable}%
    \hfil
\begin{subtable}[t]{0.3\linewidth}

\centering
\scalebox{0.53}{
    \begin{tabular}{ c | c c c c | c} 
     Task Type &    & \multicolumn{3}{c}{Open-Ended (METEOR)} &  \\ [0.5ex]
     \hline
     Method &    & \multicolumn{3}{c}{MLB with Attention} &  \\ [0.5ex]
     \hline
     Test Set&  \multicolumn{4}{c}{dev} & diff \\ [0.5ex]
     \hline
     Partition & Other & Num & Y/N & All & All \\ [0.5ex] 
     \hline
     First-dev & 2.33 & 2.55 & 22.96 & 10.82 & 54.97  \\ 
     
     Second-dev & 2.54 & 2.37 & 22.60 & 10.76 & 55.03 \\
     
     Third-dev & 2.51 & 2.03 & 22.17 & 10.53 & 55.26  \\
     
     Fourth-dev & 2.33 & 2.29 & 22.86 & 10.75 & 55.04  \\
     
     Fifth-dev & 2.33 & 2.24 & 23.27 & 10.91 & 54.88 \\
     
     Sixth-dev & 2.36 & 2.15 & 24.65 & 11.48 & 54.31  \\
     
     Seventh-dev & 2.29 & 2.12 & 24.33 & 11.32 & 54.47  \\
     \hline
     First-std & 2.29 & 2.17 & 23.14 & 10.87 & 54.81 \\
     \hline
     
     Original-dev  & 57.01 & 37.51 & 83.54 & 65.79 & -\\
     Original-std  & 56.60 & 36.63 & 83.68 & 65.68 & - \\

     \hline
    \end{tabular}}
    \centering
    \captionsetup{justification=centering}
    \caption{MLB with Attention model evaluation results.}
    
\scalebox{0.53}{
    \begin{tabular}{ c | c c c c | c} 
     Task Type &    & \multicolumn{3}{c}{Open-Ended (METEOR)} &  \\ [0.5ex]
     \hline
     Method &    & \multicolumn{3}{c}{MUTAN with Attention} &  \\ [0.5ex]
     \hline
     Test Set&  \multicolumn{4}{c}{dev} & diff \\ [0.5ex]
     \hline
     Partition & Other & Num & Y/N & All & All \\ [0.5ex] 
     \hline
     First-dev & 1.87 & 2.19 & 24.18 & 11.06 & 54.92  \\ 
     
     Second-dev & 1.89 & 2.18 & 23.81 & 10.92 & 55.06 \\
     
     Third-dev & 1.88 & 2.11 & 23.48 & 10.77 & 55.21  \\
     
     Fourth-dev & 1.82 & 2.20 & 23.94 & 10.94 & 55.04  \\
     
     Fifth-dev & 1.67 & 2.28 & 24.18 & 10.98 & 55.00 \\
     
     Sixth-dev & 1.74 & 2.20 & 24.77 & 11.24 & 54.74  \\
     
     Seventh-dev & 1.69 & 2.30 & 25.05 & 11.34 & 54.64  \\
     \hline
     First-std & 1.83 & 2.02 & 24.28 & 11.10 & 54.67 \\
     \hline
     
     Original-dev  & 56.73 & 38.35 & 84.11 & 65.98 & -\\
     Original-std  & 56.29 & 37.47 & 84.04 & 65.77 & - \\

     \hline
    \end{tabular}}
    \centering
    \captionsetup{justification=centering}
    \caption{MUTAN with Attention model evaluation results.}
\end{subtable}%
    \hfil
\begin{subtable}[t]{0.3\linewidth}
        
\centering
\scalebox{0.53}{
    \begin{tabular}{ c | c c c c | c} 
     Task Type &    & \multicolumn{3}{c}{Open-Ended (METEOR)} &  \\ [0.5ex]
     \hline
     Method &    & \multicolumn{3}{c}{HieCoAtt (Alt,Resnet200)} &  \\ [0.5ex]
     \hline
     Test Set&  \multicolumn{4}{c}{dev} & diff \\ [0.5ex]
     \hline
     Partition & Other & Num & Y/N & All & All \\ [0.5ex] 
     \hline
     First-dev & 3.34 & 3.38 & 21.92 & 10.97 & 50.84  \\ 
     
     Second-dev & 3.21 & 3.17 & 21.55 & 10.73 & 51.08 \\
     
     Third-dev & 3.29 & 3.51 & 21.44 & 10.76 & 51.05  \\
     
     Fourth-dev & 3.20 & 3.41 & 22.21 & 11.02 & 50.79  \\
     
     Fifth-dev & 3.24 & 3.36 & 22.64 & 11.21 & 50.60 \\
     
     Sixth-dev & 3.05 & 3.34 & 23.26 & 11.38 & 50.43  \\
     
     Seventh-dev & 3.12 & 3.53 & 23.52 & 11.54 & 50.27  \\
     \hline
     First-std & 3.32 & 3.18 & 21.78 & 10.92 & 51.14\\
     \hline

     Original-dev  & 51.77 & 38.65 & 79.70 & 61.81 & -\\
     Original-std  & 51.95 & 38.22 & 79.95 & 62.06 & - \\

     \hline
    \end{tabular}}
    \centering
    \captionsetup{justification=centering}
    \caption{HieCoAtt (Alt,Resnet200) model evaluation results.}

\scalebox{0.53}{
    \begin{tabular}{ c | c c c c | c} 
     Task Type &    & \multicolumn{3}{c}{Open-Ended (METEOR)} &  \\ [0.5ex]
     \hline
     Method &    & \multicolumn{3}{c}{LSTM Q+I} &  \\ [0.5ex]
     \hline
     Test Set&  \multicolumn{4}{c}{dev} & diff \\ [0.5ex]
     \hline
     Partition & Other & Num & Y/N & All & All \\ [0.5ex] 
     \hline
     First-dev & 2.15 & 3.26 & 21.91 & 10.38 & 47.64  \\ 
     
     Second-dev & 2.04 & 3.24 & 21.91 & 10.32 & 47.70 \\
     
     Third-dev & 2.08 & 2.68 & 21.67 & 10.19 & 47.83  \\
     
     Fourth-dev & 2.12 & 2.62 & 22.57 & 10.57 & 47.45  \\
     
     Fifth-dev & 1.94 & 2.86 & 23.67 & 10.96 & 47.06 \\
     
     Sixth-dev & 1.96 & 2.69 & 24.49 & 11.28 & 46.74  \\
     
     Seventh-dev & 1.94 & 2.61 & 24.12 & 11.12 & 46.90  \\
     \hline
     First-std & 2.00 & 2.76 & 22.00 & 10.32 & 47.86 \\
     \hline
     
     Original-dev  & 43.40 & 36.46 & 80.87 & 58.02 & -\\
     Original-std  & 43.90 & 36.67 & 80.38 & 58.18 & - \\

     \hline
    \end{tabular}}
    \centering
    \captionsetup{justification=centering}
    \caption{LSTM Q+I model evaluation results.}
\end{subtable}
\caption{The table shows the six state-of-the-art pretrained VQA models evaluation results on the GBQD and VQA dataset. ``-'' indicates the results are not available, ``-std'' represents the accuracy of VQA model evaluated on the complete testing set of GBQD and VQA dataset and ``-dev'' indicates the accuracy of VQA model evaluated on the partial testing set of GBQD and VQA dataset. In addition, $diff = Original_{dev_{All}} - X_{dev_{All}}$, where $X$ is equal to the ``First'', ``Second'', etc.}
\label{table:table14}
\end{table*}

\begin{table*}
\renewcommand\arraystretch{1.28}
\setlength\tabcolsep{11pt}
    \centering
\begin{subtable}[t]{0.3\linewidth}
\centering
\scalebox{0.53}{
    \begin{tabular}{ c | c c c c | c} 
     Task Type &    & \multicolumn{3}{c}{Open-Ended (BLEU-1)} &  \\ [0.5ex]
     \hline
     Method &    & \multicolumn{3}{c}{MUTAN without Attention} &  \\ [0.5ex]
     \hline
     Test Set&  \multicolumn{4}{c}{dev} & diff \\ [0.5ex]
     \hline
     Partition & Other & Num & Y/N & All & All \\ [0.5ex] 
     \hline
     First-dev & 1.37 & 2.63 & 33.31 & 14.62 & 45.54  \\ 
     
     Second-dev & 1.61 & 2.71 & 28.93 & 12.94 & 47.22 \\
     
     Third-dev & 1.56 & 2.91 & 29.32 & 13.10 & 47.06  \\
     
     Fourth-dev & 1.51 & 2.76 & 29.03 & 12.94 & 47.22  \\
     
     Fifth-dev & 1.62 & 2.85 & 29.11 & 13.04 & 47.12 \\
     
     Sixth-dev & 1.63 & 2.69 & 29.16 & 13.04 & 47.12  \\
     
     Seventh-dev & 1.57 & 2.81 & 28.93 & 12.93 & 47.23  \\
     \hline
     First-std & 1.48 & 2.60 & 33.10 & 14.63 & 45.82 \\
     \hline
     
     Original-dev  & 47.16 & 37.32 & 81.45 & 60.16 & -\\
     Original-std  & 47.57 & 36.75 & 81.56 & 60.45 & - \\

     \hline
    \end{tabular}}
    \centering
    \captionsetup{justification=centering}
    \caption{MUTAN without Attention model evaluation results.}

\scalebox{0.53}{
    \begin{tabular}{ c | c c c c | c} 
     Task Type &    & \multicolumn{3}{c}{Open-Ended (BLEU-1)} &  \\ [0.5ex]
     \hline
     Method &    & \multicolumn{3}{c}{HieCoAtt (Alt,VGG19)} &  \\ [0.5ex]
     \hline
     Test Set&  \multicolumn{4}{c}{dev} & diff \\ [0.5ex]
     \hline
     Partition & Other & Num & Y/N & All & All \\ [0.5ex] 
     \hline
     First-dev & 2.28 & 2.62 & 27.61 & 12.71 & 47.77  \\ 
     
     Second-dev & 2.33 & 2.68 & 28.23 & 13.00 & 47.48 \\
     
     Third-dev & 2.25 & 2.76 & 28.43 & 13.05 & 47.43  \\
     
     Fourth-dev & 2.23 & 2.84 & 28.46 & 13.06 & 47.42  \\
     
     Fifth-dev & 2.22 & 2.47 & 28.48 & 13.03 & 47.45 \\
     
     Sixth-dev & 2.23 & 2.65 & 28.37 & 13.00 & 47.48  \\
     
     Seventh-dev & 2.24 & 2.68 & 28.34 & 13.00 & 47.48  \\
     \hline
     First-std & 2.35 & 2.64 & 27.33 & 12.68 & 47.64 \\
     \hline

     Original-dev  & 49.14 & 38.35 & 79.63 & 60.48 & -\\
     Original-std  & 49.15 & 36.52 & 79.45 & 60.32 & - \\

     \hline
    \end{tabular}}
    \centering
    \captionsetup{justification=centering}
    \caption{HieCoAtt (Alt,VGG19) model evaluation results.}
\end{subtable}%
    \hfil
\begin{subtable}[t]{0.3\linewidth}

\centering
\scalebox{0.53}{
    \begin{tabular}{ c | c c c c | c} 
     Task Type &    & \multicolumn{3}{c}{Open-Ended (BLEU-1)} &  \\ [0.5ex]
     \hline
     Method &    & \multicolumn{3}{c}{MLB with Attention} &  \\ [0.5ex]
     \hline
     Test Set&  \multicolumn{4}{c}{dev} & diff \\ [0.5ex]
     \hline
     Partition & Other & Num & Y/N & All & All \\ [0.5ex] 
     \hline
     First-dev & 1.99 & 2.38 & 28.08 & 12.74 & 53.05  \\ 
     
     Second-dev & 1.90 & 2.22 & 28.31 & 12.77 & 53.02 \\
     
     Third-dev & 2.00 & 2.48 & 28.26 & 12.83 & 52.96  \\
     
     Fourth-dev & 1.97 & 2.53 & 28.00 & 12.72 & 53.07  \\
     
     Fifth-dev & 1.89 & 2.23 & 28.10 & 12.68 & 53.11 \\
     
     Sixth-dev & 1.90 & 2.24 & 28.18 & 12.72 & 53.07  \\
     
     Seventh-dev & 1.81 & 2.37 & 28.13 & 12.67 & 53.12  \\
     \hline
     First-std & 2.13 & 2.43 & 28.05 & 12.84 & 52.84 \\
     \hline
     
     Original-dev  & 57.01 & 37.51 & 83.54 & 65.79 & -\\
     Original-std  & 56.60 & 36.63 & 83.68 & 65.68 & - \\

     \hline
    \end{tabular}}
    \centering
    \captionsetup{justification=centering}
    \caption{MLB with Attention model evaluation results.}
    
\scalebox{0.53}{
    \begin{tabular}{ c | c c c c | c} 
     Task Type &    & \multicolumn{3}{c}{Open-Ended (BLEU-1)} &  \\ [0.5ex]
     \hline
     Method &    & \multicolumn{3}{c}{MUTAN with Attention} &  \\ [0.5ex]
     \hline
     Test Set&  \multicolumn{4}{c}{dev} & diff \\ [0.5ex]
     \hline
     Partition & Other & Num & Y/N & All & All \\ [0.5ex] 
     \hline
     First-dev & 1.52 & 2.32 & 29.67 & 13.16 & 52.82  \\ 
     
     Second-dev & 1.38 & 1.98 & 28.68 & 12.65 & 53.33 \\
     
     Third-dev & 1.48 & 2.03 & 28.50 & 12.63 & 53.35  \\
     
     Fourth-dev & 1.54 & 2.19 & 28.78 & 12.79 & 53.19  \\
     
     Fifth-dev & 1.49 & 2.27 & 28.52 & 12.67 & 53.31 \\
     
     Sixth-dev & 1.48 & 2.24 & 28.27 & 12.56 & 53.42  \\
     
     Seventh-dev & 1.45 & 2.04 & 28.64 & 12.67 & 53.31  \\
     \hline
     First-std & 1.62 & 2.34 & 29.50 & 13.19 & 52.58 \\
     \hline
     
     Original-dev  & 56.73 & 38.35 & 84.11 & 65.98 & -\\
     Original-std  & 56.29 & 37.47 & 84.04 & 65.77 & - \\

     \hline
    \end{tabular}}
    \centering
    \captionsetup{justification=centering}
    \caption{MUTAN with Attention model evaluation results.}
\end{subtable}%
    \hfil
\begin{subtable}[t]{0.3\linewidth}
        
\centering
\scalebox{0.53}{
    \begin{tabular}{ c | c c c c | c} 
     Task Type &    & \multicolumn{3}{c}{Open-Ended (BLEU-1)} &  \\ [0.5ex]
     \hline
     Method &    & \multicolumn{3}{c}{HieCoAtt (Alt,Resnet200)} &  \\ [0.5ex]
     \hline
     Test Set&  \multicolumn{4}{c}{dev} & diff \\ [0.5ex]
     \hline
     Partition & Other & Num & Y/N & All & All \\ [0.5ex] 
     \hline
     First-dev & 2.57 & 2.93 & 28.04 & 13.06 & 48.75  \\ 
     
     Second-dev & 2.60 & 2.79 & 28.16 & 13.11 & 48.70 \\
     
     Third-dev & 2.62 & 2.83 & 28.18 & 13.13 & 48.68  \\
     
     Fourth-dev & 2.68 & 2.97 & 28.34 & 13.24 & 48.57  \\
     
     Fifth-dev & 2.72 & 2.97 & 28.19 & 13.20 & 48.61 \\
     
     Sixth-dev & 2.71 & 3.07 & 28.03 & 13.14 & 48.67  \\
     
     Seventh-dev & 2.60 & 2.92 & 27.85 & 13.00 & 48.81  \\
     \hline
     First-std & 2.82 & 3.04 & 27.74 & 13.11 & 48.95\\
     \hline

     Original-dev  & 51.77 & 38.65 & 79.70 & 61.81 & -\\
     Original-std  & 51.95 & 38.22 & 79.95 & 62.06 & - \\

     \hline
    \end{tabular}}
    \centering
    \captionsetup{justification=centering}
    \caption{HieCoAtt (Alt,Resnet200) model evaluation results.}

\scalebox{0.53}{
    \begin{tabular}{ c | c c c c | c} 
     Task Type &    & \multicolumn{3}{c}{Open-Ended (BLEU-1)} &  \\ [0.5ex]
     \hline
     Method &    & \multicolumn{3}{c}{LSTM Q+I} &  \\ [0.5ex]
     \hline
     Test Set&  \multicolumn{4}{c}{dev} & diff \\ [0.5ex]
     \hline
     Partition & Other & Num & Y/N & All & All \\ [0.5ex] 
     \hline
     First-dev & 1.51 & 2.29 & 29.27 & 12.99 & 45.03  \\ 
     
     Second-dev & 1.57 & 2.29 & 29.61 & 13.16 & 44.86 \\
     
     Third-dev & 1.63 & 2.43 & 29.74 & 13.25 & 44.77  \\
     
     Fourth-dev & 1.61 & 2.39 & 29.65 & 13.20 & 44.82  \\
     
     Fifth-dev & 1.61 & 2.22 & 29.78 & 13.23 & 44.79 \\
     
     Sixth-dev & 1.53 & 2.53 & 29.80 & 13.24 & 44.78  \\
     
     Seventh-dev & 1.53 & 2.43 & 29.63 & 13.16 & 44.86  \\
     \hline
     First-std & 1.56 & 2.44 & 28.76 & 12.86 & 45.32 \\
     \hline
     
     Original-dev  & 43.40 & 36.46 & 80.87 & 58.02 & -\\
     Original-std  & 43.90 & 36.67 & 80.38 & 58.18 & - \\

     \hline
    \end{tabular}}
    \centering
    \captionsetup{justification=centering}
    \caption{LSTM Q+I model evaluation results.}
\end{subtable}
\caption{The table shows the six state-of-the-art pretrained VQA models evaluation results on the YNBQD and VQA dataset. ``-'' indicates the results are not available, ``-std'' represents the accuracy of VQA model evaluated on the complete testing set of YNBQD and VQA dataset and ``-dev'' indicates the accuracy of VQA model evaluated on the partial testing set of YNBQD and VQA dataset. In addition, $diff = Original_{dev_{All}} - X_{dev_{All}}$, where $X$ is equal to the ``First'', ``Second'', etc.}
\label{table:table15}
\end{table*}

\begin{table*}
\renewcommand\arraystretch{1.28}
\setlength\tabcolsep{11pt}
    \centering
\begin{subtable}[t]{0.3\linewidth}
\centering
\scalebox{0.53}{
    \begin{tabular}{ c | c c c c | c} 
     Task Type &    & \multicolumn{3}{c}{Open-Ended (BLEU-2)} &  \\ [0.5ex]
     \hline
     Method &    & \multicolumn{3}{c}{MUTAN without Attention} &  \\ [0.5ex]
     \hline
     Test Set&  \multicolumn{4}{c}{dev} & diff \\ [0.5ex]
     \hline
     Partition & Other & Num & Y/N & All & All \\ [0.5ex] 
     \hline
     First-dev & 1.41 & 2.89 & 33.51 & 14.75 & 45.41  \\ 
     
     Second-dev & 1.66 & 2.86 & 29.20 & 13.10 & 47.06 \\
     
     Third-dev & 1.56 & 2.91 & 29.32 & 13.10 & 47.06  \\
     
     Fourth-dev & 1.51 & 2.76 & 29.03 & 12.94 & 47.22  \\
     
     Fifth-dev & 1.61 & 2.85 & 29.34 & 13.13 & 47.03 \\
     
     Sixth-dev & 1.63 & 2.86 & 29.20 & 13.08 & 47.08  \\
     
     Seventh-dev & 1.50 & 2.97 & 28.87 & 12.89 & 47.27  \\
     \hline
     First-std & 1.35 & 2.86 & 32.94 & 14.52 & 45.93 \\
     \hline
     
     Original-dev  & 47.16 & 37.32 & 81.45 & 60.16 & -\\
     Original-std  & 47.57 & 36.75 & 81.56 & 60.45 & - \\

     \hline
    \end{tabular}}
    \centering
    \captionsetup{justification=centering}
    \caption{MUTAN without Attention model evaluation results.}

\scalebox{0.53}{
    \begin{tabular}{ c | c c c c | c} 
     Task Type &    & \multicolumn{3}{c}{Open-Ended (BLEU-2)} &  \\ [0.5ex]
     \hline
     Method &    & \multicolumn{3}{c}{HieCoAtt (Alt,VGG19)} &  \\ [0.5ex]
     \hline
     Test Set&  \multicolumn{4}{c}{dev} & diff \\ [0.5ex]
     \hline
     Partition & Other & Num & Y/N & All & All \\ [0.5ex] 
     \hline
     First-dev & 2.23 & 3.01 & 27.25 & 12.58 & 47.90  \\ 
     
     Second-dev & 2.62 & 3.50 & 28.01 & 13.14 & 47.34 \\
     
     Third-dev & 2.27 & 3.10 & 28.33 & 13.06 & 47.42  \\
     
     Fourth-dev & 2.25 & 3.11 & 28.33 & 13.05 & 47.43  \\
     
     Fifth-dev & 2.24 & 3.05 & 28.34 & 13.04 & 47.44 \\
     
     Sixth-dev & 2.29 & 3.34 & 28.36 & 13.11 & 47.37  \\
     
     Seventh-dev & 2.24 & 3.05 & 28.58 & 13.14 & 47.34  \\
     \hline
     First-std & 2.36 & 2.29 & 27.51 & 12.78 & 47.54 \\
     \hline

     Original-dev  & 49.14 & 38.35 & 79.63 & 60.48 & -\\
     Original-std  & 49.15 & 36.52 & 79.45 & 60.32 & - \\

     \hline
    \end{tabular}}
    \centering
    \captionsetup{justification=centering}
    \caption{HieCoAtt (Alt,VGG19) model evaluation results.}
\end{subtable}%
    \hfil
\begin{subtable}[t]{0.3\linewidth}

\centering
\scalebox{0.53}{
    \begin{tabular}{ c | c c c c | c} 
     Task Type &    & \multicolumn{3}{c}{Open-Ended (BLEU-2)} &  \\ [0.5ex]
     \hline
     Method &    & \multicolumn{3}{c}{MLB with Attention} &  \\ [0.5ex]
     \hline
     Test Set&  \multicolumn{4}{c}{dev} & diff \\ [0.5ex]
     \hline
     Partition & Other & Num & Y/N & All & All \\ [0.5ex] 
     \hline
     First-dev & 1.98 & 2.64 & 28.38 & 12.89 & 52.90  \\ 
     
     Second-dev & 2.08 & 2.45 & 28.05 & 12.78 & 53.01 \\
     
     Third-dev & 2.00 & 2.48 & 28.26 & 12.83 & 52.96  \\
     
     Fourth-dev & 1.97 & 2.53 & 28.00 & 12.72 & 53.07  \\
     
     Fifth-dev & 1.92 & 2.34 & 28.64 & 12.93 & 52.86 \\
     
     Sixth-dev & 1.95 & 2.37 & 28.66 & 12.96 & 52.83  \\
     
     Seventh-dev & 1.90 & 2.44 & 28.34 & 12.81 & 52.98  \\
     \hline
     First-std & 2.06 & 2.70 & 28.50 & 13.02 & 52.66 \\
     \hline
     
     Original-dev  & 57.01 & 37.51 & 83.54 & 65.79 & -\\
     Original-std  & 56.60 & 36.63 & 83.68 & 65.68 & - \\

     \hline
    \end{tabular}}
    \centering
    \captionsetup{justification=centering}
    \caption{MLB with Attention model evaluation results.}
    
\scalebox{0.53}{
    \begin{tabular}{ c | c c c c | c} 
     Task Type &    & \multicolumn{3}{c}{Open-Ended (BLEU-2)} &  \\ [0.5ex]
     \hline
     Method &    & \multicolumn{3}{c}{MUTAN with Attention} &  \\ [0.5ex]
     \hline
     Test Set&  \multicolumn{4}{c}{dev} & diff \\ [0.5ex]
     \hline
     Partition & Other & Num & Y/N & All & All \\ [0.5ex] 
     \hline
     First-dev & 1.45 & 2.24 & 29.63 & 13.10 & 52.88  \\ 
     
     Second-dev & 1.38 & 2.22 & 28.41 & 12.57 & 53.41 \\
     
     Third-dev & 1.48 & 2.03 & 28.50 & 12.63 & 53.35  \\
     
     Fourth-dev & 1.54 & 2.19 & 28.78 & 12.79 & 53.19  \\
     
     Fifth-dev & 1.39 & 2.39 & 28.71 & 12.71 & 53.27 \\
     
     Sixth-dev & 1.47 & 2.08 & 28.65 & 12.69 & 53.29  \\
     
     Seventh-dev & 1.45 & 2.16 & 28.50 & 12.63 & 53.35  \\
     \hline
     First-std & 1.58 & 2.38 & 29.15 & 13.02 & 52.75 \\
     \hline
     
     Original-dev  & 56.73 & 38.35 & 84.11 & 65.98 & -\\
     Original-std  & 56.29 & 37.47 & 84.04 & 65.77 & - \\

     \hline
    \end{tabular}}
    \centering
    \captionsetup{justification=centering}
    \caption{MUTAN with Attention model evaluation results.}
\end{subtable}%
    \hfil
\begin{subtable}[t]{0.3\linewidth}
        
\centering
\scalebox{0.53}{
    \begin{tabular}{ c | c c c c | c} 
     Task Type &    & \multicolumn{3}{c}{Open-Ended (BLEU-2)} &  \\ [0.5ex]
     \hline
     Method &    & \multicolumn{3}{c}{HieCoAtt (Alt,Resnet200)} &  \\ [0.5ex]
     \hline
     Test Set&  \multicolumn{4}{c}{dev} & diff \\ [0.5ex]
     \hline
     Partition & Other & Num & Y/N & All & All \\ [0.5ex] 
     \hline
     First-dev & 2.62 & 3.50 & 28.01 & 13.14 & 48.67  \\ 
     
     Second-dev & 2.72 & 3.48 & 28.12 & 13.22 & 48.59 \\
     
     Third-dev & 2.79 & 3.39 & 28.17 & 13.27 & 48.54  \\
     
     Fourth-dev & 2.76 & 3.44 & 28.32 & 13.33 & 48.48  \\
     
     Fifth-dev & 2.74 & 3.48 & 28.24 & 13.28 & 48.53 \\
     
     Sixth-dev & 2.68 & 3.38 & 28.30 & 13.27 & 48.54  \\
     
     Seventh-dev & 2.79 & 3.30 & 28.46 & 13.38 & 48.43  \\
     \hline
     First-std & 2.61 & 3.20 & 28.00 & 13.13 & 48.93\\
     \hline

     Original-dev  & 51.77 & 38.65 & 79.70 & 61.81 & -\\
     Original-std  & 51.95 & 38.22 & 79.95 & 62.06 & - \\

     \hline
    \end{tabular}}
    \centering
    \captionsetup{justification=centering}
    \caption{HieCoAtt (Alt,Resnet200) model evaluation results.}

\scalebox{0.53}{
    \begin{tabular}{ c | c c c c | c} 
     Task Type &    & \multicolumn{3}{c}{Open-Ended (BLEU-2)} &  \\ [0.5ex]
     \hline
     Method &    & \multicolumn{3}{c}{LSTM Q+I} &  \\ [0.5ex]
     \hline
     Test Set&  \multicolumn{4}{c}{dev} & diff \\ [0.5ex]
     \hline
     Partition & Other & Num & Y/N & All & All \\ [0.5ex] 
     \hline
     First-dev & 1.60 & 2.65 & 28.81 & 12.88 & 45.14  \\ 
     
     Second-dev & 1.65 & 2.33 & 29.48 & 13.15 & 44.87 \\
     
     Third-dev & 1.63 & 2.54 & 29.66 & 13.23 & 44.79  \\
     
     Fourth-dev & 1.60 & 2.63 & 29.26 & 13.07 & 44.95  \\
     
     Fifth-dev & 1.61 & 2.55 & 29.96 & 13.34 & 44.68 \\
     
     Sixth-dev & 1.68 & 2.60 & 29.39 & 13.15 & 44.87  \\
     
     Seventh-dev & 1.53 & 2.44 & 29.75 & 13.21 & 44.81  \\
     \hline
     First-std & 1.55 & 2.59 & 29.22 & 13.06 & 45.12 \\
     \hline
     
     Original-dev  & 43.40 & 36.46 & 80.87 & 58.02 & -\\
     Original-std  & 43.90 & 36.67 & 80.38 & 58.18 & - \\

     \hline
    \end{tabular}}
    \centering
    \captionsetup{justification=centering}
    \caption{LSTM Q+I model evaluation results.}
\end{subtable}
\caption{The table shows the six state-of-the-art pretrained VQA models evaluation results on the YNBQD and VQA dataset. ``-'' indicates the results are not available, ``-std'' represents the accuracy of VQA model evaluated on the complete testing set of YNBQD and VQA dataset and ``-dev'' indicates the accuracy of VQA model evaluated on the partial testing set of YNBQD and VQA dataset. In addition, $diff = Original_{dev_{All}} - X_{dev_{All}}$, where $X$ is equal to the ``First'', ``Second'', etc.}
\label{table:table16}
\end{table*}

\begin{table*}
\renewcommand\arraystretch{1.28}
\setlength\tabcolsep{11pt}
    \centering
\begin{subtable}[t]{0.3\linewidth}
\centering
\scalebox{0.53}{
    \begin{tabular}{ c | c c c c | c} 
     Task Type &    & \multicolumn{3}{c}{Open-Ended (BLEU-3)} &  \\ [0.5ex]
     \hline
     Method &    & \multicolumn{3}{c}{MUTAN without Attention} &  \\ [0.5ex]
     \hline
     Test Set&  \multicolumn{4}{c}{dev} & diff \\ [0.5ex]
     \hline
     Partition & Other & Num & Y/N & All & All \\ [0.5ex] 
     \hline
     First-dev & 1.33 & 2.69 & 32.75 & 14.37 & 45.79  \\ 
     
     Second-dev & 1.58 & 2.67 & 28.73 & 12.84 & 47.32 \\
     
     Third-dev & 1.56 & 2.82 & 28.35 & 12.69 & 47.47  \\
     
     Fourth-dev & 1.58 & 2.83 & 28.47 & 12.75 & 47.41  \\
     
     Fifth-dev & 1.57 & 2.62 & 28.57 & 12.77 & 47.39 \\
     
     Sixth-dev & 1.53 & 2.51 & 28.61 & 12.75 & 47.41  \\
     
     Seventh-dev & 1.57 & 2.68 & 28.03 & 12.55 & 47.61  \\
     \hline
     First-std & 1.22 & 2.81 & 32.85 & 14.42 & 46.03 \\
     \hline
     
     Original-dev  & 47.16 & 37.32 & 81.45 & 60.16 & -\\
     Original-std  & 47.57 & 36.75 & 81.56 & 60.45 & - \\

     \hline
    \end{tabular}}
    \centering
    \captionsetup{justification=centering}
    \caption{MUTAN without Attention model evaluation results.}

\scalebox{0.53}{
    \begin{tabular}{ c | c c c c | c} 
     Task Type &    & \multicolumn{3}{c}{Open-Ended (BLEU-3)} &  \\ [0.5ex]
     \hline
     Method &    & \multicolumn{3}{c}{HieCoAtt (Alt,VGG19)} &  \\ [0.5ex]
     \hline
     Test Set&  \multicolumn{4}{c}{dev} & diff \\ [0.5ex]
     \hline
     Partition & Other & Num & Y/N & All & All \\ [0.5ex] 
     \hline
     First-dev & 2.45 & 3.03 & 27.34 & 12.73 & 47.75  \\ 
     
     Second-dev & 2.29 & 3.06 & 27.91 & 12.89 & 47.59 \\
     
     Third-dev & 2.33 & 2.90 & 27.97 & 12.91 & 47.57  \\
     
     Fourth-dev & 2.30 & 2.97 & 28.38 & 13.08 & 47.40  \\
     
     Fifth-dev & 2.34 & 2.79 & 27.81 & 12.84 & 47.64 \\
     
     Sixth-dev & 2.27 & 3.01 & 28.13 & 12.96 & 47.52  \\
     
     Seventh-dev & 2.24 & 2.90 & 27.84 & 12.82 & 47.66  \\
     \hline
     First-std & 2.31 & 2.97 & 26.98 & 12.55 & 47.77 \\
     \hline

     Original-dev  & 49.14 & 38.35 & 79.63 & 60.48 & -\\
     Original-std  & 49.15 & 36.52 & 79.45 & 60.32 & - \\

     \hline
    \end{tabular}}
    \centering
    \captionsetup{justification=centering}
    \caption{HieCoAtt (Alt,VGG19) model evaluation results.}
\end{subtable}%
    \hfil
\begin{subtable}[t]{0.3\linewidth}

\centering
\scalebox{0.53}{
    \begin{tabular}{ c | c c c c | c} 
     Task Type &    & \multicolumn{3}{c}{Open-Ended (BLEU-3)} &  \\ [0.5ex]
     \hline
     Method &    & \multicolumn{3}{c}{MLB with Attention} &  \\ [0.5ex]
     \hline
     Test Set&  \multicolumn{4}{c}{dev} & diff \\ [0.5ex]
     \hline
     Partition & Other & Num & Y/N & All & All \\ [0.5ex] 
     \hline
     First-dev & 2.05 & 2.59 & 27.65 & 12.62 & 53.17  \\ 
     
     Second-dev & 2.02 & 2.52 & 27.77 & 12.64 & 53.15 \\
     
     Third-dev & 1.92 & 2.37 & 28.01 & 12.67 & 53.12  \\
     
     Fourth-dev & 1.94 & 2.58 & 27.70 & 12.58 & 53.21  \\
     
     Fifth-dev & 1.85 & 2.51 & 27.94 & 12.63 & 53.16 \\
     
     Sixth-dev & 1.89 & 2.54 & 28.14 & 12.74 & 53.05  \\
     
     Seventh-dev & 1.94 & 2.18 & 27.58 & 12.49 & 53.30  \\
     \hline
     First-std & 1.91 & 2.78 & 28.31 & 12.88 & 52.80 \\
     \hline
     
     Original-dev  & 57.01 & 37.51 & 83.54 & 65.79 & -\\
     Original-std  & 56.60 & 36.63 & 83.68 & 65.68 & - \\

     \hline
    \end{tabular}}
    \centering
    \captionsetup{justification=centering}
    \caption{MLB with Attention model evaluation results.}
    
\scalebox{0.53}{
    \begin{tabular}{ c | c c c c | c} 
     Task Type &    & \multicolumn{3}{c}{Open-Ended (BLEU-3)} &  \\ [0.5ex]
     \hline
     Method &    & \multicolumn{3}{c}{MUTAN with Attention} &  \\ [0.5ex]
     \hline
     Test Set&  \multicolumn{4}{c}{dev} & diff \\ [0.5ex]
     \hline
     Partition & Other & Num & Y/N & All & All \\ [0.5ex] 
     \hline
     First-dev & 1.50 & 2.31 & 28.94 & 12.85 & 53.13  \\ 
     
     Second-dev & 1.41 & 2.37 & 28.03 & 12.44 & 53.54 \\
     
     Third-dev & 1.47 & 2.26 & 27.96 & 12.43 & 53.55  \\
     
     Fourth-dev & 1.45 & 1.91 & 28.04 & 12.42 & 53.56  \\
     
     Fifth-dev & 1.46 & 2.33 & 28.45 & 12.63 & 53.35 \\
     
     Sixth-dev & 1.47 & 2.12 & 28.25 & 12.53 & 53.45  \\
     
     Seventh-dev & 1.41 & 1.95 & 27.83 & 12.31 & 53.67  \\
     \hline
     First-std & 1.47 & 2.44 & 29.26 & 13.02 & 52.75 \\
     \hline
     
     Original-dev  & 56.73 & 38.35 & 84.11 & 65.98 & -\\
     Original-std  & 56.29 & 37.47 & 84.04 & 65.77 & - \\

     \hline
    \end{tabular}}
    \centering
    \captionsetup{justification=centering}
    \caption{MUTAN with Attention model evaluation results.}
\end{subtable}%
    \hfil
\begin{subtable}[t]{0.3\linewidth}
        
\centering
\scalebox{0.53}{
    \begin{tabular}{ c | c c c c | c} 
     Task Type &    & \multicolumn{3}{c}{Open-Ended (BLEU-3)} &  \\ [0.5ex]
     \hline
     Method &    & \multicolumn{3}{c}{HieCoAtt (Alt,Resnet200)} &  \\ [0.5ex]
     \hline
     Test Set&  \multicolumn{4}{c}{dev} & diff \\ [0.5ex]
     \hline
     Partition & Other & Num & Y/N & All & All \\ [0.5ex] 
     \hline
     First-dev & 2.55 & 3.26 & 27.80 & 12.99 & 48.82  \\ 
     
     Second-dev & 2.56 & 3.20 & 28.05 & 13.09 & 48.72 \\
     
     Third-dev & 2.79 & 3.08 & 27.84 & 13.10 & 48.71  \\
     
     Fourth-dev & 2.71 & 3.24 & 28.31 & 13.27 & 48.54  \\
     
     Fifth-dev & 2.69 & 3.08 & 27.62 & 12.96 & 48.85 \\
     
     Sixth-dev & 2.83 & 3.12 & 28.02 & 13.20 & 48.61  \\
     
     Seventh-dev & 2.68 & 3.06 & 27.62 & 12.96 & 48.85  \\
     \hline
     First-std & 2.45 & 3.14 & 27.61 & 12.89 & 49.17\\
     \hline

     Original-dev  & 51.77 & 38.65 & 79.70 & 61.81 & -\\
     Original-std  & 51.95 & 38.22 & 79.95 & 62.06 & - \\

     \hline
    \end{tabular}}
    \centering
    \captionsetup{justification=centering}
    \caption{HieCoAtt (Alt,Resnet200) model evaluation results.}

\scalebox{0.53}{
    \begin{tabular}{ c | c c c c | c} 
     Task Type &    & \multicolumn{3}{c}{Open-Ended (BLEU-3)} &  \\ [0.5ex]
     \hline
     Method &    & \multicolumn{3}{c}{LSTM Q+I} &  \\ [0.5ex]
     \hline
     Test Set&  \multicolumn{4}{c}{dev} & diff \\ [0.5ex]
     \hline
     Partition & Other & Num & Y/N & All & All \\ [0.5ex] 
     \hline
     First-dev & 1.46 & 2.66 & 28.97 & 12.88 & 45.14  \\ 
     
     Second-dev & 1.54 & 2.72 & 29.35 & 13.08 & 44.94 \\
     
     Third-dev & 1.57 & 2.91 & 29.73 & 13.27 & 44.75  \\
     
     Fourth-dev & 1.54 & 2.68 & 29.34 & 13.07 & 44.95  \\
     
     Fifth-dev & 1.46 & 2.70 & 29.88 & 13.26 & 44.76 \\
     
     Sixth-dev & 1.48 & 2.67 & 29.58 & 13.14 & 44.88  \\
     
     Seventh-dev & 1.56 & 2.47 & 29.26 & 13.03 & 44.99  \\
     \hline
     First-std & 1.49 & 2.61 & 29.11 & 12.99 & 45.19 \\
     \hline
     
     Original-dev  & 43.40 & 36.46 & 80.87 & 58.02 & -\\
     Original-std  & 43.90 & 36.67 & 80.38 & 58.18 & - \\

     \hline
    \end{tabular}}
    \centering
    \captionsetup{justification=centering}
    \caption{LSTM Q+I model evaluation results.}
\end{subtable}
\caption{The table shows the six state-of-the-art pretrained VQA models evaluation results on the YNBQD and VQA dataset. ``-'' indicates the results are not available, ``-std'' represents the accuracy of VQA model evaluated on the complete testing set of YNBQD and VQA dataset and ``-dev'' indicates the accuracy of VQA model evaluated on the partial testing set of YNBQD and VQA dataset. In addition, $diff = Original_{dev_{All}} - X_{dev_{All}}$, where $X$ is equal to the ``First'', ``Second'', etc.}
\label{table:table17}
\end{table*}

\begin{table*}
\renewcommand\arraystretch{1.28}
\setlength\tabcolsep{11pt}
    \centering
\begin{subtable}[t]{0.3\linewidth}
\centering
\scalebox{0.53}{
    \begin{tabular}{ c | c c c c | c} 
     Task Type &    & \multicolumn{3}{c}{Open-Ended (BLEU-4)} &  \\ [0.5ex]
     \hline
     Method &    & \multicolumn{3}{c}{MUTAN without Attention} &  \\ [0.5ex]
     \hline
     Test Set&  \multicolumn{4}{c}{dev} & diff \\ [0.5ex]
     \hline
     Partition & Other & Num & Y/N & All & All \\ [0.5ex] 
     \hline
     First-dev & 1.31 & 2.63 & 33.27 & 14.57 & 45.59  \\ 
     
     Second-dev & 1.58 & 2.63 & 29.26 & 13.06 & 47.10 \\
     
     Third-dev & 1.53 & 2.68 & 28.93 & 12.90 & 47.26  \\
     
     Fourth-dev & 1.58 & 2.59 & 28.92 & 12.91 & 47.25  \\
     
     Fifth-dev & 1.51 & 2.69 & 29.28 & 13.03 & 47.13 \\
     
     Sixth-dev & 1.59 & 2.47 & 29.40 & 13.10 & 47.06  \\
     
     Seventh-dev & 1.54 & 2.53 & 28.56 & 12.74 & 47.42  \\
     \hline
     First-std & 1.41 & 2.67 & 33.06 & 14.58 & 45.87 \\
     \hline
     
     Original-dev  & 47.16 & 37.32 & 81.45 & 60.16 & -\\
     Original-std  & 47.57 & 36.75 & 81.56 & 60.45 & - \\

     \hline
    \end{tabular}}
    \centering
    \captionsetup{justification=centering}
    \caption{MUTAN without Attention model evaluation results.}

\scalebox{0.53}{
    \begin{tabular}{ c | c c c c | c} 
     Task Type &    & \multicolumn{3}{c}{Open-Ended (BLEU-4)} &  \\ [0.5ex]
     \hline
     Method &    & \multicolumn{3}{c}{HieCoAtt (Alt,VGG19)} &  \\ [0.5ex]
     \hline
     Test Set&  \multicolumn{4}{c}{dev} & diff \\ [0.5ex]
     \hline
     Partition & Other & Num & Y/N & All & All \\ [0.5ex] 
     \hline
     First-dev & 2.19 & 2.64 & 27.47 & 12.61 & 47.87  \\ 
     
     Second-dev & 2.17 & 2.78 & 28.07 & 12.86 & 47.62 \\
     
     Third-dev & 2.17 & 2.73 & 28.46 & 13.02 & 47.46  \\
     
     Fourth-dev & 2.17 & 2.79 & 28.29 & 12.95 & 47.53  \\
     
     Fifth-dev & 2.23 & 2.72 & 28.06 & 12.88 & 47.60 \\
     
     Sixth-dev & 2.24 & 3.03 & 28.44 & 13.07 & 47.41  \\
     
     Seventh-dev & 2.06 & 2.60 & 28.53 & 12.98 & 47.50  \\
     \hline
     First-std & 2.29 & 2.85 & 27.52 & 12.74 & 47.48 \\
     \hline

     Original-dev  & 49.14 & 38.35 & 79.63 & 60.48 & -\\
     Original-std  & 49.15 & 36.52 & 79.45 & 60.32 & - \\

     \hline
    \end{tabular}}
    \centering
    \captionsetup{justification=centering}
    \caption{HieCoAtt (Alt,VGG19) model evaluation results.}
\end{subtable}%
    \hfil
\begin{subtable}[t]{0.3\linewidth}

\centering
\scalebox{0.53}{
    \begin{tabular}{ c | c c c c | c} 
     Task Type &    & \multicolumn{3}{c}{Open-Ended (BLEU-4)} &  \\ [0.5ex]
     \hline
     Method &    & \multicolumn{3}{c}{MLB with Attention} &  \\ [0.5ex]
     \hline
     Test Set&  \multicolumn{4}{c}{dev} & diff \\ [0.5ex]
     \hline
     Partition & Other & Num & Y/N & All & All \\ [0.5ex] 
     \hline
     First-dev & 1.92 & 2.36 & 28.46 & 12.86 & 52.93  \\ 
     
     Second-dev & 1.88 & 2.02 & 28.06 & 12.64 & 53.15 \\
     
     Third-dev & 1.92 & 2.26 & 28.50 & 12.87 & 52.92  \\
     
     Fourth-dev & 1.84 & 2.29 & 27.93 & 12.60 & 53.19  \\
     
     Fifth-dev & 1.86 & 2.13 & 28.34 & 12.76 & 53.03 \\
     
     Sixth-dev & 1.86 & 2.29 & 28.49 & 12.84 & 52.95  \\
     
     Seventh-dev & 1.83 & 2.25 & 28.17 & 12.68 & 53.11  \\
     \hline
     First-std & 2.01 & 2.62 & 28.09 & 12.82 & 52.86 \\
     \hline
     
     Original-dev  & 57.01 & 37.51 & 83.54 & 65.79 & -\\
     Original-std  & 56.60 & 36.63 & 83.68 & 65.68 & - \\

     \hline
    \end{tabular}}
    \centering
    \captionsetup{justification=centering}
    \caption{MLB with Attention model evaluation results.}
    
\scalebox{0.53}{
    \begin{tabular}{ c | c c c c | c} 
     Task Type &    & \multicolumn{3}{c}{Open-Ended (BLEU-4)} &  \\ [0.5ex]
     \hline
     Method &    & \multicolumn{3}{c}{MUTAN with Attention} &  \\ [0.5ex]
     \hline
     Test Set&  \multicolumn{4}{c}{dev} & diff \\ [0.5ex]
     \hline
     Partition & Other & Num & Y/N & All & All \\ [0.5ex] 
     \hline
     First-dev & 1.42 & 1.96 & 29.38 & 12.95 & 53.03  \\ 
     
     Second-dev & 1.38 & 1.97 & 28.35 & 12.51 & 53.47 \\
     
     Third-dev & 1.35 & 1.66 & 28.78 & 12.64 & 53.34  \\
     
     Fourth-dev & 1.33 & 2.12 & 28.57 & 12.60 & 53.38  \\
     
     Fifth-dev & 1.32 & 1.90 & 28.72 & 12.63 & 53.35 \\
     
     Sixth-dev & 1.43 & 1.76 & 28.46 & 12.56 & 53.42  \\
     
     Seventh-dev & 1.38 & 1.87 & 28.54 & 12.58 & 53.40  \\
     \hline
     First-std & 1.53 & 2.26 & 29.29 & 13.05 & 52.72 \\
     \hline
     
     Original-dev  & 56.73 & 38.35 & 84.11 & 65.98 & -\\
     Original-std  & 56.29 & 37.47 & 84.04 & 65.77 & - \\

     \hline
    \end{tabular}}
    \centering
    \captionsetup{justification=centering}
    \caption{MUTAN with Attention model evaluation results.}
\end{subtable}%
    \hfil
\begin{subtable}[t]{0.3\linewidth}
        
\centering
\scalebox{0.53}{
    \begin{tabular}{ c | c c c c | c} 
     Task Type &    & \multicolumn{3}{c}{Open-Ended (BLEU-4)} &  \\ [0.5ex]
     \hline
     Method &    & \multicolumn{3}{c}{HieCoAtt (Alt,Resnet200)} &  \\ [0.5ex]
     \hline
     Test Set&  \multicolumn{4}{c}{dev} & diff \\ [0.5ex]
     \hline
     Partition & Other & Num & Y/N & All & All \\ [0.5ex] 
     \hline
     First-dev & 2.41 & 3.19 & 28.14 & 13.06 & 48.75  \\ 
     
     Second-dev & 2.46 & 2.92 & 28.18 & 13.06 & 48.75 \\
     
     Third-dev & 2.62 & 2.89 & 28.29 & 13.18 & 48.63  \\
     
     Fourth-dev & 2.58 & 3.15 & 28.36 & 13.23 & 48.58  \\
     
     Fifth-dev & 2.55 & 2.97 & 27.86 & 12.98 & 48.83 \\
     
     Sixth-dev & 2.60 & 3.12 & 28.14 & 13.14 & 48.67  \\
     
     Seventh-dev & 2.60 & 2.87 & 28.18 & 13.13 & 48.68  \\
     \hline
     First-std & 2.61 & 2.90 & 28.06 & 13.13 & 48.93\\
     \hline

     Original-dev  & 51.77 & 38.65 & 79.70 & 61.81 & -\\
     Original-std  & 51.95 & 38.22 & 79.95 & 62.06 & - \\

     \hline
    \end{tabular}}
    \centering
    \captionsetup{justification=centering}
    \caption{HieCoAtt (Alt,Resnet200) model evaluation results.}

\scalebox{0.53}{
    \begin{tabular}{ c | c c c c | c} 
     Task Type &    & \multicolumn{3}{c}{Open-Ended (BLEU-4)} &  \\ [0.5ex]
     \hline
     Method &    & \multicolumn{3}{c}{LSTM Q+I} &  \\ [0.5ex]
     \hline
     Test Set&  \multicolumn{4}{c}{dev} & diff \\ [0.5ex]
     \hline
     Partition & Other & Num & Y/N & All & All \\ [0.5ex] 
     \hline
     First-dev & 1.59 & 2.17 & 28.94 & 12.88 & 45.14  \\ 
     
     Second-dev & 1.57 & 2.49 & 29.33 & 13.06 & 44.96 \\
     
     Third-dev & 1.54 & 2.40 & 29.56 & 13.13 & 44.89  \\
     
     Fourth-dev & 1.56 & 2.43 & 29.20 & 13.00 & 45.02  \\
     
     Fifth-dev & 1.51 & 2.44 & 29.41 & 13.06 & 44.96 \\
     
     Sixth-dev & 1.44 & 2.30 & 29.42 & 13.02 & 45.00  \\
     
     Seventh-dev & 1.58 & 2.16 & 29.10 & 12.93 & 45.09  \\
     \hline
     First-std & 1.53 & 2.59 & 29.29 & 13.08 & 45.10 \\
     \hline
     
     Original-dev  & 43.40 & 36.46 & 80.87 & 58.02 & -\\
     Original-std  & 43.90 & 36.67 & 80.38 & 58.18 & - \\

     \hline
    \end{tabular}}
    \centering
    \captionsetup{justification=centering}
    \caption{LSTM Q+I model evaluation results.}
\end{subtable}
\caption{The table shows the six state-of-the-art pretrained VQA models evaluation results on the YNBQD and VQA dataset. ``-'' indicates the results are not available, ``-std'' represents the accuracy of VQA model evaluated on the complete testing set of YNBQD and VQA dataset and ``-dev'' indicates the accuracy of VQA model evaluated on the partial testing set of YNBQD and VQA dataset. In addition, $diff = Original_{dev_{All}} - X_{dev_{All}}$, where $X$ is equal to the ``First'', ``Second'', etc.}
\label{table:table18}
\end{table*}

\begin{table*}
\renewcommand\arraystretch{1.28}
\setlength\tabcolsep{11pt}
    \centering
\begin{subtable}[t]{0.3\linewidth}
\centering
\scalebox{0.53}{
    \begin{tabular}{ c | c c c c | c} 
     Task Type &    & \multicolumn{3}{c}{Open-Ended (ROUGE)} &  \\ [0.5ex]
     \hline
     Method &    & \multicolumn{3}{c}{MUTAN without Attention} &  \\ [0.5ex]
     \hline
     Test Set&  \multicolumn{4}{c}{dev} & diff \\ [0.5ex]
     \hline
     Partition & Other & Num & Y/N & All & All \\ [0.5ex] 
     \hline
     First-dev & 1.55 & 2.35 & 29.92 & 13.28 & 46.88  \\ 
     
     Second-dev & 1.47 & 2.67 & 28.79 & 12.81 & 47.35 \\
     
     Third-dev & 1.40 & 2.48 & 25.47 & 11.40 & 48.76  \\
     
     Fourth-dev & 1.67 & 2.59 & 26.52 & 11.97 & 48.19  \\
     
     Fifth-dev & 1.59 & 2.92 & 29.18 & 13.06 & 47.10 \\
     
     Sixth-dev & 1.97 & 2.62 & 29.74 & 13.44 & 46.72  \\
     
     Seventh-dev & 1.69 & 2.60 & 29.30 & 13.12 & 47.04  \\
     \hline
     First-std & 1.38 & 2.56 & 29.67 & 13.16 & 47.29 \\
     \hline
     
     Original-dev  & 47.16 & 37.32 & 81.45 & 60.16 & -\\
     Original-std  & 47.57 & 36.75 & 81.56 & 60.45 & - \\

     \hline
    \end{tabular}}
    \centering
    \captionsetup{justification=centering}
    \caption{MUTAN without Attention model evaluation results.}

\scalebox{0.53}{
    \begin{tabular}{ c | c c c c | c} 
     Task Type &    & \multicolumn{3}{c}{Open-Ended (ROUGE)} &  \\ [0.5ex]
     \hline
     Method &    & \multicolumn{3}{c}{HieCoAtt (Alt,VGG19)} &  \\ [0.5ex]
     \hline
     Test Set&  \multicolumn{4}{c}{dev} & diff \\ [0.5ex]
     \hline
     Partition & Other & Num & Y/N & All & All \\ [0.5ex] 
     \hline
     First-dev & 2.09 & 2.54 & 25.59 & 11.79 & 48.69  \\ 
     
     Second-dev & 2.19 & 3.17 & 28.79 & 13.21 & 47.27 \\
     
     Third-dev & 2.17 & 2.68 & 29.00 & 13.24 & 47.24  \\
     
     Fourth-dev & 2.16 & 2.69 & 26.97 & 12.40 & 48.08  \\
     
     Fifth-dev & 2.34 & 3.00 & 29.58 & 13.59 & 46.89 \\
     
     Sixth-dev & 2.39 & 2.91 & 28.80 & 13.28 & 47.20  \\
     
     Seventh-dev & 2.36 & 3.01 & 29.38 & 13.52 & 46.96  \\
     \hline
     First-std & 2.15 & 3.11 & 25.19 & 11.75 & 48.57 \\
     \hline

     Original-dev  & 49.14 & 38.35 & 79.63 & 60.48 & -\\
     Original-std  & 49.15 & 36.52 & 79.45 & 60.32 & - \\

     \hline
    \end{tabular}}
    \centering
    \captionsetup{justification=centering}
    \caption{HieCoAtt (Alt,VGG19) model evaluation results.}
\end{subtable}%
    \hfil
\begin{subtable}[t]{0.3\linewidth}

\centering
\scalebox{0.53}{
    \begin{tabular}{ c | c c c c | c} 
     Task Type &    & \multicolumn{3}{c}{Open-Ended (ROUGE)} &  \\ [0.5ex]
     \hline
     Method &    & \multicolumn{3}{c}{MLB with Attention} &  \\ [0.5ex]
     \hline
     Test Set&  \multicolumn{4}{c}{dev} & diff \\ [0.5ex]
     \hline
     Partition & Other & Num & Y/N & All & All \\ [0.5ex] 
     \hline
     First-dev & 1.89 & 2.33 & 28.44 & 12.83 & 52.96  \\ 
     
     Second-dev & 1.64 & 2.50 & 31.01 & 13.79 & 52.00 \\
     
     Third-dev & 1.83 & 2.19 & 26.32 & 11.92 & 53.87  \\
     
     Fourth-dev & 1.85 & 2.61 & 27.86 & 12.60 & 53.19  \\
     
     Fifth-dev & 1.92 & 2.37 & 29.53 & 13.30 & 52.49 \\
     
     Sixth-dev & 2.28 & 2.50 & 27.24 & 12.54 & 53.25  \\
     
     Seventh-dev & 2.01 & 2.43 & 29.77 & 13.45 & 52.34  \\
     \hline
     First-std & 1.71 & 2.37 & 28.41 & 12.78 & 52.90 \\
     \hline
     
     Original-dev  & 57.01 & 37.51 & 83.54 & 65.79 & -\\
     Original-std  & 56.60 & 36.63 & 83.68 & 65.68 & - \\

     \hline
    \end{tabular}}
    \centering
    \captionsetup{justification=centering}
    \caption{MLB with Attention model evaluation results.}
    
\scalebox{0.53}{
    \begin{tabular}{ c | c c c c | c} 
     Task Type &    & \multicolumn{3}{c}{Open-Ended (ROUGE)} &  \\ [0.5ex]
     \hline
     Method &    & \multicolumn{3}{c}{MUTAN with Attention} &  \\ [0.5ex]
     \hline
     Test Set&  \multicolumn{4}{c}{dev} & diff \\ [0.5ex]
     \hline
     Partition & Other & Num & Y/N & All & All \\ [0.5ex] 
     \hline
     First-dev & 1.65 & 1.97 & 27.33 & 12.22 & 53.76  \\ 
     
     Second-dev & 1.11 & 1.94 & 28.33 & 12.37 & 53.61 \\
     
     Third-dev & 1.23 & 1.90 & 27.34 & 12.02 & 53.96  \\
     
     Fourth-dev & 1.38 & 2.09 & 26.72 & 11.85 & 54.13  \\
     
     Fifth-dev & 1.46 & 2.14 & 25.69 & 11.48 & 54.50 \\
     
     Sixth-dev & 1.73 & 2.17 & 27.42 & 12.32 & 53.66  \\
     
     Seventh-dev & 1.47 & 1.97 & 27.27 & 12.11 & 53.87  \\
     \hline
     First-std & 1.45 & 2.17 & 27.04 & 12.07 & 53.70 \\
     \hline
     
     Original-dev  & 56.73 & 38.35 & 84.11 & 65.98 & -\\
     Original-std  & 56.29 & 37.47 & 84.04 & 65.77 & - \\

     \hline
    \end{tabular}}
    \centering
    \captionsetup{justification=centering}
    \caption{MUTAN with Attention model evaluation results.}
\end{subtable}%
    \hfil
\begin{subtable}[t]{0.3\linewidth}
        
\centering
\scalebox{0.53}{
    \begin{tabular}{ c | c c c c | c} 
     Task Type &    & \multicolumn{3}{c}{Open-Ended (ROUGE)} &  \\ [0.5ex]
     \hline
     Method &    & \multicolumn{3}{c}{HieCoAtt (Alt,Resnet200)} &  \\ [0.5ex]
     \hline
     Test Set&  \multicolumn{4}{c}{dev} & diff \\ [0.5ex]
     \hline
     Partition & Other & Num & Y/N & All & All \\ [0.5ex] 
     \hline
     First-dev & 2.72 & 2.76 & 27.45 & 12.87 & 48.94  \\ 
     
     Second-dev & 2.68 & 2.77 & 28.25 & 13.19 & 48.62 \\
     
     Third-dev & 2.74 & 2.98 & 27.84 & 13.07 & 48.74  \\
     
     Fourth-dev & 2.43 & 2.87 & 26.70 & 12.44 & 49.37  \\
     
     Fifth-dev & 2.71 & 2.83 & 29.40 & 13.68 & 48.13 \\
     
     Sixth-dev & 3.06 & 2.82 & 28.54 & 13.49 & 48.32  \\
     
     Seventh-dev & 2.69 & 3.08 & 29.06 & 13.55 & 48.26  \\
     \hline
     First-std & 2.54 & 3.10 & 27.66 & 12.95 & 49.11 \\
     \hline

     Original-dev  & 51.77 & 38.65 & 79.70 & 61.81 & -\\
     Original-std  & 51.95 & 38.22 & 79.95 & 62.06 & - \\

     \hline
    \end{tabular}}
    \centering
    \captionsetup{justification=centering}
    \caption{HieCoAtt (Alt,Resnet200) model evaluation results.}

\scalebox{0.53}{
    \begin{tabular}{ c | c c c c | c} 
     Task Type &    & \multicolumn{3}{c}{Open-Ended (ROUGE)} &  \\ [0.5ex]
     \hline
     Method &    & \multicolumn{3}{c}{LSTM Q+I} &  \\ [0.5ex]
     \hline
     Test Set&  \multicolumn{4}{c}{dev} & diff \\ [0.5ex]
     \hline
     Partition & Other & Num & Y/N & All & All \\ [0.5ex] 
     \hline
     First-dev & 1.60 & 2.35 & 28.65 & 12.78 & 45.24  \\ 
     
     Second-dev & 1.64 & 2.37 & 28.63 & 12.79 & 45.23 \\
     
     Third-dev & 1.64 & 2.51 & 26.05 & 11.75 & 46.27  \\
     
     Fourth-dev & 1.50 & 2.49 & 28.21 & 12.57 & 45.45  \\
     
     Fifth-dev & 1.41 & 2.37 & 28.74 & 12.73 & 45.29 \\
     
     Sixth-dev & 1.60 & 2.40 & 28.81 & 12.85 & 45.17  \\
     
     Seventh-dev & 1.53 & 2.57 & 28.94 & 12.89 & 45.13  \\
     \hline
     First-std & 1.51 & 2.59 & 28.37 & 12.69 & 45.49 \\
     \hline
     
     Original-dev  & 43.40 & 36.46 & 80.87 & 58.02 & -\\
     Original-std  & 43.90 & 36.67 & 80.38 & 58.18 & - \\

     \hline
    \end{tabular}}
    \centering
    \captionsetup{justification=centering}
    \caption{LSTM Q+I model evaluation results.}
\end{subtable}
\caption{The table shows the six state-of-the-art pretrained VQA models evaluation results on the YNBQD and VQA dataset. ``-'' indicates the results are not available, ``-std'' represents the accuracy of VQA model evaluated on the complete testing set of YNBQD and VQA dataset and ``-dev'' indicates the accuracy of VQA model evaluated on the partial testing set of YNBQD and VQA dataset. In addition, $diff = Original_{dev_{All}} - X_{dev_{All}}$, where $X$ is equal to the ``First'', ``Second'', etc.}
\label{table:table19}
\end{table*}

\begin{table*}
\renewcommand\arraystretch{1.28}
\setlength\tabcolsep{11pt}
    \centering
\begin{subtable}[t]{0.3\linewidth}
\centering
\scalebox{0.53}{
    \begin{tabular}{ c | c c c c | c} 
     Task Type &    & \multicolumn{3}{c}{Open-Ended (CIDEr)} &  \\ [0.5ex]
     \hline
     Method &    & \multicolumn{3}{c}{MUTAN without Attention} &  \\ [0.5ex]
     \hline
     Test Set&  \multicolumn{4}{c}{dev} & diff \\ [0.5ex]
     \hline
     Partition & Other & Num & Y/N & All & All \\ [0.5ex] 
     \hline
     First-dev & 0.98 & 1.97 & 24.24 & 10.63 & 49.53  \\ 
     
     Second-dev & 1.18 & 1.90 & 24.06 & 10.65 & 49.51 \\
     
     Third-dev & 1.43 & 2.37 & 33.75 & 14.79 & 45.37  \\
     
     Fourth-dev & 1.28 & 2.46 & 37.13 & 16.12 & 44.04  \\
     
     Fifth-dev & 1.27 & 2.02 & 22.73 & 10.16 & 50.00 \\
     
     Sixth-dev & 1.25 & 1.73 & 27.49 & 12.07 & 48.09  \\
     
     Seventh-dev & 1.38 & 2.33 & 38.10 & 16.55 & 43.61  \\
     \hline
     First-std & 1.07 & 2.24 & 23.83 & 10.57 & 49.88 \\
     \hline
     
     Original-dev  & 47.16 & 37.32 & 81.45 & 60.16 & -\\
     Original-std  & 47.57 & 36.75 & 81.56 & 60.45 & - \\

     \hline
    \end{tabular}}
    \centering
    \captionsetup{justification=centering}
    \caption{MUTAN without Attention model evaluation results.}

\scalebox{0.53}{
    \begin{tabular}{ c | c c c c | c} 
     Task Type &    & \multicolumn{3}{c}{Open-Ended (CIDEr)} &  \\ [0.5ex]
     \hline
     Method &    & \multicolumn{3}{c}{HieCoAtt (Alt,VGG19)} &  \\ [0.5ex]
     \hline
     Test Set&  \multicolumn{4}{c}{dev} & diff \\ [0.5ex]
     \hline
     Partition & Other & Num & Y/N & All & All \\ [0.5ex] 
     \hline
     First-dev & 1.77 & 1.71 & 22.52 & 10.28 & 50.20  \\ 
     
     Second-dev & 2.36 & 1.75 & 31.07 & 14.08 & 46.40 \\
     
     Third-dev & 2.28 & 2.78 & 37.65 & 16.85 & 43.63  \\
     
     Fourth-dev & 2.61 & 3.60 & 31.71 & 14.66 & 45.82  \\
     
     Fifth-dev & 1.95 & 1.96 & 25.75 & 11.72 & 48.76 \\
     
     Sixth-dev & 2.28 & 2.08 & 46.58 & 20.44 & 40.04  \\
     
     Seventh-dev & 2.20 & 3.12 & 29.98 & 13.70 & 46.78  \\
     \hline
     First-std & 1.93 & 1.64 & 22.07 & 10.20 & 50.12 \\
     \hline

     Original-dev  & 49.14 & 38.35 & 79.63 & 60.48 & -\\
     Original-std  & 49.15 & 36.52 & 79.45 & 60.32 & - \\

     \hline
    \end{tabular}}
    \centering
    \captionsetup{justification=centering}
    \caption{HieCoAtt (Alt,VGG19) model evaluation results.}
\end{subtable}%
    \hfil
\begin{subtable}[t]{0.3\linewidth}

\centering
\scalebox{0.53}{
    \begin{tabular}{ c | c c c c | c} 
     Task Type &    & \multicolumn{3}{c}{Open-Ended (CIDEr)} &  \\ [0.5ex]
     \hline
     Method &    & \multicolumn{3}{c}{MLB with Attention} &  \\ [0.5ex]
     \hline
     Test Set&  \multicolumn{4}{c}{dev} & diff \\ [0.5ex]
     \hline
     Partition & Other & Num & Y/N & All & All \\ [0.5ex] 
     \hline
     First-dev & 1.28 & 2.21 & 24.63 & 10.96 & 54.83  \\ 
     
     Second-dev & 1.88 & 2.04 & 29.59 & 13.27 & 52.52 \\
     
     Third-dev & 1.53 & 2.22 & 30.38 & 13.45 & 52.34  \\
     
     Fourth-dev & 2.24 & 2.31 & 26.90 & 12.37 & 53.42  \\
     
     Fifth-dev & 1.70 & 1.80 & 22.54 & 10.27 & 55.52 \\
     
     Sixth-dev & 1.97 & 2.23 & 26.66 & 12.13 & 53.66  \\
     
     Seventh-dev & 1.95 & 2.20 & 34.14 & 15.19 & 50.60  \\
     \hline
     First-std & 1.41 & 2.24 & 24.34 & 10.94 & 54.74 \\
     \hline
     
     Original-dev  & 57.01 & 37.51 & 83.54 & 65.79 & -\\
     Original-std  & 56.60 & 36.63 & 83.68 & 65.68 & - \\

     \hline
    \end{tabular}}
    \centering
    \captionsetup{justification=centering}
    \caption{MLB with Attention model evaluation results.}
    
\scalebox{0.53}{
    \begin{tabular}{ c | c c c c | c} 
     Task Type &    & \multicolumn{3}{c}{Open-Ended (CIDEr)} &  \\ [0.5ex]
     \hline
     Method &    & \multicolumn{3}{c}{MUTAN with Attention} &  \\ [0.5ex]
     \hline
     Test Set&  \multicolumn{4}{c}{dev} & diff \\ [0.5ex]
     \hline
     Partition & Other & Num & Y/N & All & All \\ [0.5ex] 
     \hline
     First-dev & 0.89 & 1.67 & 24.52 & 10.67 & 55.31  \\ 
     
     Second-dev & 1.34 & 1.52 & 27.07 & 11.92 & 54.06 \\
     
     Third-dev & 1.18 & 1.77 & 28.88 & 12.61 & 53.37  \\
     
     Fourth-dev & 1.79 & 2.25 & 33.61 & 14.90 & 51.08  \\
     
     Fifth-dev & 1.14 & 1.09 & 22.81 & 10.03 & 55.95 \\
     
     Sixth-dev & 1.69 & 1.42 & 27.40 & 12.21 & 53.77  \\
     
     Seventh-dev & 1.46 & 1.63 & 40.24 & 17.39 & 48.59  \\
     \hline
     First-std & 0.95 & 1.94 & 24.22 & 10.64 & 55.13 \\
     \hline
     
     Original-dev  & 56.73 & 38.35 & 84.11 & 65.98 & -\\
     Original-std  & 56.29 & 37.47 & 84.04 & 65.77 & - \\

     \hline
    \end{tabular}}
    \centering
    \captionsetup{justification=centering}
    \caption{MUTAN with Attention model evaluation results.}
\end{subtable}%
    \hfil
\begin{subtable}[t]{0.3\linewidth}
        
\centering
\scalebox{0.53}{
    \begin{tabular}{ c | c c c c | c} 
     Task Type &    & \multicolumn{3}{c}{Open-Ended (CIDEr)} &  \\ [0.5ex]
     \hline
     Method &    & \multicolumn{3}{c}{HieCoAtt (Alt,Resnet200)} &  \\ [0.5ex]
     \hline
     Test Set&  \multicolumn{4}{c}{dev} & diff \\ [0.5ex]
     \hline
     Partition & Other & Num & Y/N & All & All \\ [0.5ex] 
     \hline
     First-dev & 2.02 & 3.46 & 25.31 & 11.74 & 50.07  \\ 
     
     Second-dev & 2.57 & 2.99 & 28.60 & 13.30 & 48.51 \\
     
     Third-dev & 3.04 & 3.48 & 33.16 & 15.45 & 46.36  \\
     
     Fourth-dev & 2.95 & 3.47 & 31.42 & 14.69 & 47.12  \\
     
     Fifth-dev & 2.46 & 3.09 & 25.13 & 11.83 & 49.98 \\
     
     Sixth-dev & 3.50 & 3.08 & 35.11 & 16.43 & 45.38  \\
     
     Seventh-dev & 2.83 & 2.90 & 31.99 & 14.80 & 47.01  \\
     \hline
     First-std & 2.13 & 3.55 & 24.88 & 11.65 & 50.41 \\
     \hline

     Original-dev  & 51.77 & 38.65 & 79.70 & 61.81 & -\\
     Original-std  & 51.95 & 38.22 & 79.95 & 62.06 & - \\

     \hline
    \end{tabular}}
    \centering
    \captionsetup{justification=centering}
    \caption{HieCoAtt (Alt,Resnet200) model evaluation results.}

\scalebox{0.53}{
    \begin{tabular}{ c | c c c c | c} 
     Task Type &    & \multicolumn{3}{c}{Open-Ended (CIDEr)} &  \\ [0.5ex]
     \hline
     Method &    & \multicolumn{3}{c}{LSTM Q+I} &  \\ [0.5ex]
     \hline
     Test Set&  \multicolumn{4}{c}{dev} & diff \\ [0.5ex]
     \hline
     Partition & Other & Num & Y/N & All & All \\ [0.5ex] 
     \hline
     First-dev & 1.16 & 3.75 & 24.60 & 11.06 & 46.96  \\ 
     
     Second-dev & 1.17 & 1.60 & 25.30 & 11.12 & 46.90 \\
     
     Third-dev & 1.18 & 2.06 & 30.91 & 13.48 & 44.54  \\
     
     Fourth-dev & 1.69 & 2.15 & 32.15 & 14.24 & 43.78  \\
     
     Fifth-dev & 1.09 & 2.51 & 26.54 & 11.69 & 46.33 \\
     
     Sixth-dev & 1.43 & 0.93 & 36.37 & 15.72 & 42.30  \\
     
     Seventh-dev & 1.47 & 2.06 & 36.58 & 15.94 & 42.08  \\
     \hline
     First-std & 1.17 & 3.67 & 24.15 & 10.90 & 47.28 \\
     \hline
     
     Original-dev  & 43.40 & 36.46 & 80.87 & 58.02 & -\\
     Original-std  & 43.90 & 36.67 & 80.38 & 58.18 & - \\

     \hline
    \end{tabular}}
    \centering
    \captionsetup{justification=centering}
    \caption{LSTM Q+I model evaluation results.}
\end{subtable}
\caption{The table shows the six state-of-the-art pretrained VQA models evaluation results on the YNBQD and VQA dataset. ``-'' indicates the results are not available, ``-std'' represents the accuracy of VQA model evaluated on the complete testing set of YNBQD and VQA dataset and ``-dev'' indicates the accuracy of VQA model evaluated on the partial testing set of YNBQD and VQA dataset. In addition, $diff = Original_{dev_{All}} - X_{dev_{All}}$, where $X$ is equal to the ``First'', ``Second'', etc.}
\label{table:table20}
\end{table*}

\begin{table*}
\renewcommand\arraystretch{1.28}
\setlength\tabcolsep{11pt}
    \centering
\begin{subtable}[t]{0.3\linewidth}
\centering
\scalebox{0.53}{
    \begin{tabular}{ c | c c c c | c} 
     Task Type &    & \multicolumn{3}{c}{Open-Ended (METEOR)} &  \\ [0.5ex]
     \hline
     Method &    & \multicolumn{3}{c}{MUTAN without Attention} &  \\ [0.5ex]
     \hline
     Test Set&  \multicolumn{4}{c}{dev} & diff \\ [0.5ex]
     \hline
     Partition & Other & Num & Y/N & All & All \\ [0.5ex] 
     \hline
     First-dev & 1.58 & 2.71 & 26.49 & 11.93 & 48.23  \\ 
     
     Second-dev & 1.53 & 2.66 & 26.84 & 12.04 & 48.12 \\
     
     Third-dev & 1.56 & 2.60 & 27.43 & 12.29 & 47.87  \\
     
     Fourth-dev & 1.46 & 2.56 & 27.65 & 12.33 & 47.83  \\
     
     Fifth-dev & 1.50 & 2.67 & 27.50 & 12.30 & 47.86 \\
     
     Sixth-dev & 1.51 & 2.70 & 27.33 & 12.24 & 47.92  \\
     
     Seventh-dev & 1.55 & 2.50 & 27.58 & 12.34 & 47.82  \\
     \hline
     First-std & 1.68 & 3.03 & 26.93 & 12.23 & 48.22 \\
     \hline
     
     Original-dev  & 47.16 & 37.32 & 81.45 & 60.16 & -\\
     Original-std  & 47.57 & 36.75 & 81.56 & 60.45 & - \\

     \hline
    \end{tabular}}
    \centering
    \captionsetup{justification=centering}
    \caption{MUTAN without Attention model evaluation results.}

\scalebox{0.53}{
    \begin{tabular}{ c | c c c c | c} 
     Task Type &    & \multicolumn{3}{c}{Open-Ended (METEOR)} &  \\ [0.5ex]
     \hline
     Method &    & \multicolumn{3}{c}{HieCoAtt (Alt,VGG19)} &  \\ [0.5ex]
     \hline
     Test Set&  \multicolumn{4}{c}{dev} & diff \\ [0.5ex]
     \hline
     Partition & Other & Num & Y/N & All & All \\ [0.5ex] 
     \hline
     First-dev & 2.24 & 2.88 & 27.39 & 12.63 & 47.85  \\ 
     
     Second-dev & 2.21 & 3.06 & 27.66 & 12.75 & 47.73 \\
     
     Third-dev & 2.22 & 3.30 & 27.80 & 12.83 & 47.65  \\
     
     Fourth-dev & 2.21 & 2.89 & 27.85 & 12.80 & 47.68  \\
     
     Fifth-dev & 2.29 & 2.89 & 27.93 & 12.88 & 47.60 \\
     
     Sixth-dev & 2.17 & 2.79 & 28.02 & 12.85 & 47.63  \\
     
     Seventh-dev & 2.29 & 2.97 & 28.21 & 13.00 & 47.48  \\
     \hline
     First-std & 2.17 & 2.77 & 27.54 & 12.69 & 47.63 \\
     \hline

     Original-dev  & 49.14 & 38.35 & 79.63 & 60.48 & -\\
     Original-std  & 49.15 & 36.52 & 79.45 & 60.32 & - \\

     \hline
    \end{tabular}}
    \centering
    \captionsetup{justification=centering}
    \caption{HieCoAtt (Alt,VGG19) model evaluation results.}
\end{subtable}%
    \hfil
\begin{subtable}[t]{0.3\linewidth}

\centering
\scalebox{0.53}{
    \begin{tabular}{ c | c c c c | c} 
     Task Type &    & \multicolumn{3}{c}{Open-Ended (METEOR)} &  \\ [0.5ex]
     \hline
     Method &    & \multicolumn{3}{c}{MLB with Attention} &  \\ [0.5ex]
     \hline
     Test Set&  \multicolumn{4}{c}{dev} & diff \\ [0.5ex]
     \hline
     Partition & Other & Num & Y/N & All & All \\ [0.5ex] 
     \hline
     First-dev & 1.92 & 2.06 & 26.27 & 11.93 & 53.86  \\ 
     
     Second-dev & 1.82 & 2.48 & 26.84 & 12.16 & 53.63 \\
     
     Third-dev & 1.81 & 2.24 & 27.33 & 12.33 & 53.46  \\
     
     Fourth-dev & 1.74 & 2.31 & 27.89 & 12.53 & 53.26  \\
     
     Fifth-dev & 1.84 & 2.34 & 27.57 & 12.45 & 53.34 \\
     
     Sixth-dev & 1.84 & 2.26 & 27.30 & 12.33 & 53.46  \\
     
     Seventh-dev & 1.78 & 2.26 & 27.68 & 12.46 & 53.33  \\
     \hline
     First-std & 1.91 & 2.20 & 26.76 & 12.18 & 53.50 \\
     \hline
     
     Original-dev  & 57.01 & 37.51 & 83.54 & 65.79 & -\\
     Original-std  & 56.60 & 36.63 & 83.68 & 65.68 & - \\

     \hline
    \end{tabular}}
    \centering
    \captionsetup{justification=centering}
    \caption{MLB with Attention model evaluation results.}
    
\scalebox{0.53}{
    \begin{tabular}{ c | c c c c | c} 
     Task Type &    & \multicolumn{3}{c}{Open-Ended (METEOR)} &  \\ [0.5ex]
     \hline
     Method &    & \multicolumn{3}{c}{MUTAN with Attention} &  \\ [0.5ex]
     \hline
     Test Set&  \multicolumn{4}{c}{dev} & diff \\ [0.5ex]
     \hline
     Partition & Other & Num & Y/N & All & All \\ [0.5ex] 
     \hline
     First-dev & 1.55 & 2.35 & 26.75 & 11.98 & 54.00  \\ 
     
     Second-dev & 1.48 & 2.46 & 27.23 & 12.16 & 53.82 \\
     
     Third-dev & 1.42 & 2.25 & 27.63 & 12.27 & 53.71  \\
     
     Fourth-dev & 1.38 & 2.49 & 28.28 & 12.54 & 53.44  \\
     
     Fifth-dev & 1.43 & 2.30 & 27.91 & 12.39 & 53.59 \\
     
     Sixth-dev & 1.44 & 2.25 & 27.97 & 12.41 & 53.57  \\
     
     Seventh-dev & 1.42 & 2.08 & 27.69 & 12.27 & 53.71  \\
     \hline
     First-std & 1.57 & 2.31 & 27.41 & 12.30 & 53.47 \\
     \hline
     
     Original-dev  & 56.73 & 38.35 & 84.11 & 65.98 & -\\
     Original-std  & 56.29 & 37.47 & 84.04 & 65.77 & - \\

     \hline
    \end{tabular}}
    \centering
    \captionsetup{justification=centering}
    \caption{MUTAN with Attention model evaluation results.}
\end{subtable}%
    \hfil
\begin{subtable}[t]{0.3\linewidth}
        
\centering
\scalebox{0.53}{
    \begin{tabular}{ c | c c c c | c} 
     Task Type &    & \multicolumn{3}{c}{Open-Ended (METEOR)} &  \\ [0.5ex]
     \hline
     Method &    & \multicolumn{3}{c}{HieCoAtt (Alt,Resnet200)} &  \\ [0.5ex]
     \hline
     Test Set&  \multicolumn{4}{c}{dev} & diff \\ [0.5ex]
     \hline
     Partition & Other & Num & Y/N & All & All \\ [0.5ex] 
     \hline
     First-dev & 2.71 & 3.26 & 26.99 & 12.73 & 49.08  \\ 
     
     Second-dev & 2.81 & 3.34 & 27.43 & 12.97 & 48.84 \\
     
     Third-dev & 2.83 & 3.41 & 27.46 & 13.00 & 48.81  \\
     
     Fourth-dev & 2.78 & 3.12 & 27.22 & 12.85 & 48.96  \\
     
     Fifth-dev & 2.70 & 3.12 & 27.30 & 12.84 & 48.97 \\
     
     Sixth-dev & 2.77 & 2.97 & 27.37 & 12.89 & 48.92  \\
     
     Seventh-dev & 2.76 & 3.03 & 27.78 & 13.06 & 48.75  \\
     \hline
     First-std & 2.73 & 3.03 & 27.26 & 12.87 & 49.19 \\
     \hline

     Original-dev  & 51.77 & 38.65 & 79.70 & 61.81 & -\\
     Original-std  & 51.95 & 38.22 & 79.95 & 62.06 & - \\

     \hline
    \end{tabular}}
    \centering
    \captionsetup{justification=centering}
    \caption{HieCoAtt (Alt,Resnet200) model evaluation results.}

\scalebox{0.53}{
    \begin{tabular}{ c | c c c c | c} 
     Task Type &    & \multicolumn{3}{c}{Open-Ended (METEOR)} &  \\ [0.5ex]
     \hline
     Method &    & \multicolumn{3}{c}{LSTM Q+I} &  \\ [0.5ex]
     \hline
     Test Set&  \multicolumn{4}{c}{dev} & diff \\ [0.5ex]
     \hline
     Partition & Other & Num & Y/N & All & All \\ [0.5ex] 
     \hline
     First-dev & 1.64 & 2.78 & 27.95 & 12.56 & 45.46  \\ 
     
     Second-dev & 1.48 & 2.82 & 28.42 & 12.68 & 45.34 \\
     
     Third-dev & 1.63 & 2.43 & 28.44 & 12.72 & 45.30  \\
     
     Fourth-dev & 1.47 & 2.58 & 28.65 & 12.74 & 45.28  \\
     
     Fifth-dev & 1.57 & 2.47 & 29.04 & 12.94 & 45.08 \\
     
     Sixth-dev & 1.59 & 2.57 & 28.46 & 12.72 & 45.30  \\
     
     Seventh-dev & 1.52 & 2.40 & 28.84 & 12.83 & 45.19  \\
     \hline
     First-std & 1.53 & 2.75 & 28.19 & 12.64 & 45.54 \\
     \hline
     
     Original-dev  & 43.40 & 36.46 & 80.87 & 58.02 & -\\
     Original-std  & 43.90 & 36.67 & 80.38 & 58.18 & - \\

     \hline
    \end{tabular}}
    \centering
    \captionsetup{justification=centering}
    \caption{LSTM Q+I model evaluation results.}
\end{subtable}
\caption{The table shows the six state-of-the-art pretrained VQA models evaluation results on the YNBQD and VQA dataset. ``-'' indicates the results are not available, ``-std'' represents the accuracy of VQA model evaluated on the complete testing set of YNBQD and VQA dataset and ``-dev'' indicates the accuracy of VQA model evaluated on the partial testing set of YNBQD and VQA dataset. In addition, $diff = Original_{dev_{All}} - X_{dev_{All}}$, where $X$ is equal to the ``First'', ``Second'', etc.}
\label{table:table21}
\end{table*}

\bibliographystyle{spbasic}      

\bibliography{IJCV}   

\end{sloppypar}
\end{document}